\documentclass[9pt, a4paper]{article}
\usepackage[margin=1in]{geometry}
\usepackage[utf8]{inputenc}
\usepackage[T1]{fontenc}

\usepackage{algorithm}
\usepackage{algpseudocode}
\usepackage{subcaption}
\usepackage{amsmath}
\usepackage{amssymb}
\usepackage{orcidlink}
\usepackage[title]{appendix}
\usepackage{amsmath}
\usepackage{amssymb}
\usepackage{titlesec}
\usepackage{authblk}
\usepackage{cite}

\usepackage{comment}
\usepackage{soul}
\usepackage{xcolor}
\usepackage{tikz}

\newcommand\copyrighttext{%
  \footnotesize {\textbf{\color{red}(*) Preprint version. This paper has been published on Springer Transportation Journal (eISSN 1572-9435). 
  Please cite DOI: 10.1007/s11116-025-10593-x}}}
\newcommand\copyrightnotice{%
\begin{tikzpicture}[remember picture,overlay]
\node[anchor=south,yshift=10pt] at (current page.south) {\fbox{\parbox{\dimexpr\textwidth-\fboxsep-\fboxrule\relax}{\copyrighttext}}};
\end{tikzpicture}%
}

\newcommand{\avgRealTraffic}{y}
\newcommand{\simulator}{\hat{y}}
\newcommand{\trafficModel}{\mathcal{M}}
\newcommand{\route}{\phi}
\newcommand{\regionRoute}{\Phi}
\newcommand{\region}{\rho}
\newcommand{\regionSet}{\mathcal{R}}
\newcommand{\weightFun}{\lambda}
\newcommand{\timeInterval}{\delta}
\newcommand{\timeIntervalSet}{\Delta}
\newcommand{\edgeSet}{\mathcal{E}}
\newcommand{\pivotEdgeSet}{\mathcal{E}^{pivot}}
\newcommand{\errorSet}{Err}
\newcommand{\error}{\epsilon}
\newcommand{\convergenceRate}{q}
\newcommand{\maxRegionalRouteLen}{m}
\newcommand{\nearbyRegionOp}{\sim}

\newcommand{\todayDate}{November 13, 2024}

\newcommand{\orcidDavide}{\orcidlink{0000-0002-6865-1833}}
\newcommand{\orcidAle}{\orcidlink{0000-0003-0053-4902}}
\newcommand{\orcidGianluca}{\orcidlink{0000-0001-8621-316X}}
\newcommand{\orcidBruno}{\orcidlink{0000-0002-0688-8173}}

\title{Calibration of Vehicular Traffic Simulation Models by Local Optimization*}

\author[1,2,*]{Davide Andrea Guastella\orcidDavide}
\author[2]{Alejandro Morales-Hern\'{a}ndez\orcidAle}
\author[3]{Bruno Cornelis\orcidBruno}
\author[2]{Gianluca Bontempi\orcidGianluca}
\affil[1]{Aix Marseille University, CNRS, LIS, Marseille, France}
\affil[2]{Machine Learning Group, Universit\'{e} Libre de Bruxelles}
\affil[3]{Macq Mobility; Department of Electronics and Informatics, Vrije Universiteit Brussel}

\affil[*]{davide.guastella@lis-lab.fr}

\date{}
\begin{document}

\maketitle
\copyrightnotice
\thispagestyle{empty}

\begin{abstract}
Simulation is a valuable tool for traffic management experts to assist them in refining and improving transportation systems and anticipating the impact of possible changes in the infrastructure network before their actual implementation. Calibrating simulation models using traffic count data is challenging because of the complexity of the environment, the lack of data, and the uncertainties in traffic dynamics.

This paper introduces a novel stochastic simulation-based traffic calibration technique. The novelty of the proposed method is: (\textit{i}) it performs local traffic calibration, (\textit{ii}) it allows calibrating simulated traffic in large-scale environments, (\textit{iii}) it requires only the traffic count data. The local approach enables decentralizing the calibration task to reach near real-time performance, enabling the fostering of digital twins. Using only traffic count data makes the proposed method generic so that it can be applied in different traffic scenarios at various scales (from neighborhood to region). We assess the proposed technique on a model of Brussels, Belgium, using data from real traffic monitoring devices. The proposed method has been implemented using the open-source traffic simulator SUMO. Experimental results show that the traffic model calibrated using the proposed method is on average 16\% more accurate than those obtained by the state-of-the-art methods, using the same dataset. We also make available the output traffic model obtained from real data. 
\end{abstract}


\section*{Introduction}

City administrations aim at sustainable development, where reducing the human footprint is a priority. Tackling sustainable development in densely urbanized environments poses the challenge of driving disruptive changes in complex environments. A major challenge in the traffic domain is providing traffic management experts with smart methods to analyze and evaluate traffic policies and avoid congestion, a major cause of CO$_2$ emissions~\cite{doi:10.1080/19427867.2022.2065592}. On the one hand, the impact of the control strategies on road infrastructures is not observable until they are deployed in the real world~\cite {KUSIC2023101858}. On the other hand, testing control strategies in real-life settings are expensive, risky, and often unfeasible~\cite{argota_sanchez_vaquerizo_2021}. In this context, urban traffic simulation models have become an indispensable asset, providing an \textit{in-silico} environment where it is possible to design and assess alternative control strategies before the deployment. The most important requirement for a traffic model is its realism. If the dynamics of simulated traffic are close to reality, then traffic management experts can reliably assess the consequences of control strategies in the real world~\cite{siebke_what_2022}.

This paper addresses the problem of estimating traffic simulation models using traffic count data acquired from traffic monitoring devices. We propose a local optimization method that calibrates the simulated traffic to match the input traffic data as much as possible. First, we partition the city network into a set of non-overlapping regions, where a region delimits a part of the road network. We create simulated vehicles starting from each region. The number of vehicles is progressively calibrated to minimize the difference between the real and simulated traffic. Although our proposal can be implemented with any microscopic traffic simulator software, in this work we use the open-source traffic simulator SUMO.

The novelties of the proposed method are: (\textit{i}) it performs local traffic calibration, (\textit{ii}) it allows calibrating simulated traffic in large-scale environments, (\textit{iii}) it requires only traffic count data.

The remainder of this paper is organized as follows: first, we motivate the local approach for addressing traffic calibration. Then, we discuss different state-of-the-art methods to address traffic calibration. Following, we formalize the problem addressed in this paper. The following section introduces the proposed technique for calibrating traffic models using traffic count data. We then illustrate the experimental setup and compare the results obtained from the proposed technique and state-of-the-art methods for calibrating traffic. Finally, we summarize the main findings of our work and outline future research directions.

\section*{Motivation}\label{sec:motivations}

The intuition behind the local calibration is to constrain both in space and time the task of calibrating the traffic to the observed data. This is challenging, as the calibration problem is underdetermined~\cite{osorio_high-dimensional_2019}: the available data can be used to define a virtual representation of traffic that is faithful to input information but not to the real dynamics of traffic~\cite{nie_towards_2023}. Our objective is to estimate traffic that matches the input data, without necessarily ensuring the accuracy of vehicle trajectories considering the real traffic demand or addressing uncertainties deriving from traffic counts.

By isolating parts of the road network, it is possible to vary traffic parameters locally, analyze the local error, and evaluate the impact of local parameters on nearby regions. A further motivation to perform local traffic calibration is related to computational advantages, as it is possible to distribute the load of the traffic calibration over different computational modules, one for each region and time interval. This is effective when considering traffic count data streams in large-scale environments.

\section*{Related Work} \label{sec:relatedwork}

The implementation of realistic traffic simulation models has been addressed from different perspectives and has employed various strategies to enhance model accuracy, scalability, and applicability in diverse urban settings. Sha et al.~\cite{sha2020applying} propose a Bayesian optimization framework for the calibrating problem of traffic simulation models. Bayesian optimization is an approach to optimize expensive objective functions. Hence, the benefits of the suggested calibration method because traffic simulations are generally expensive in terms of simulation time and resource consumption. The key idea of Bayesian optimization is to choose more promising values for future evaluation based on previous results. This is achieved by constructing a surrogate model which is a probability model of the original objective function and is easier to optimize (i.e. it is computationally efficient). The proposed method calibrates simulation parameters such as the vehicles' maximum speed, acceleration, deceleration, and minimum gap after the leading vehicle for four types of vehicles, resulting in 16 parameters to be calibrated (one per vehicle type). The objective function to minimize is the sum of the root mean square percentage error (RMSPE) of traffic count and speed. The authors validate their proposal using a model of a section of the New Jersey Turnpike and compare their results to those obtained by the Simultaneous Perturbation Stochastic Approximation (SPSA), a stochastic approximation algorithm well suited for optimizing functions involving a high number of parameters. The minimum value of the objective function achieved using Bayesian optimization and SPSA is 6.87\% and 7.41\%, respectively. While Bayesian optimization guarantees good results when considering several variables in a large-scale scenario, the SPSA technique showed convergence issues ~\cite{kostic_techniques_2017}. In addition to the inherent limitations of Bayesian optimization-based methods (e.g., computational complexity is proportional to the number of observed points), the primary drawback of this study is in the actual problem design. Since the authors optimize several parameters for four different vehicle types, they do not consider the traffic dynamics observed throughout the day. Therefore, the calibrated model uses the same parameter values for each vehicle without considering that these may be different at different times. Furthermore, the authors do not consider the number of vehicles entering/leaving the simulation in the area of study. This can have an impact on the traffic counts seen throughout the simulation, ultimately lowering the performance indicator used as the optimization objective.

A more realistic approach to calibrate traffic simulation scenarios is to tune the traffic demand as the set of vehicle trips (or routes) from origin to destination that are produced on a road during a specific period. Gonzalez-Delicado et al.~\cite{9606704} propose a method to create simulated traffic from traffic measurements using Cadyts, a tool that iteratively adjusts the probability of selecting the routes to match the input traffic counts data~\cite{behrisch_comparison_2022}. The accuracy of the traffic calibrated using Cadyts depends on the input routes. The authors start from a set of random routes, then they clone some of the routes that are subsequently provided as input to Cadyts. This process of cloning routes is repeated until the difference between the real and simulated traffic is minimized. The authors validate their proposal on an Alicante-Murcia freeway (Spain) section. The difference in vehicles per hour between the input traffic counts and the simulated ones is about 132 using the proposed iterative method and 225 using Cadyts. The key benefit of the proposed calibration method is that it does not require pre-calibration to generate realistic traffic demands and can generate calibrated mixed traffic (light and heavy vehicles) using as input only the traffic flows measured by the road detectors or sensors. However, the process of cloning routes does not consider the effect that cloned routes may have on other traffic measurement points. This can lead to inaccurate results in complex, large-scale scenarios.

The method proposed by Gonzalez-Delicado et al.~\cite{9606704} is closely related to optimizing Origin-Destination (OD) matrices. However, rather than directly optimizing the traffic moving between two sites, the authors focus on cloning previous trips, with the notion that considering historical knowledge about these trips can better adapt the simulated traffic patterns. Some of the most recent methods for creating realistic traffic simulation models directly modify OD matrices~\cite{zhang_efficient_2017}. Each entry in the OD matrix defines the volume of traffic flowing between two points in a road network connected by at least one path. We refer the interested reader to Abrahamsson, T. \cite{abrahamsson1998estimation} and Rong, C., Ding, J., and Li, Y.\cite{rong2023interdisciplinary} for a detailed review of OD matrices estimation using traffic counts.

Zhu et al.~\cite{zhu_demand_2020} propose an Adaptive Fine-Tuning (AFT) algorithm for estimating OD matrices for an area of the city of Nanjing (China). The authors use the SUMO simulator to simulate traffic and obtain feedback from simulated traffic to infer OD pairs that minimize the error between the simulated traffic by SUMO and real-world Radio Frequency Identification (RFID) data. The input RFID data is aggregated at a 5-minute interval from 6:00 AM to 10:00 AM on a working day. When resampling the data with a frequency of 15 minutes, approximately 80\% of links demonstrate a mean absolute percentage error (MAPE) smaller than 20\%. The results demonstrated that the method is a feasible and efficient pathway for investigating urban road networks' spatiotemporal characteristics and how they are reflected during OD matrices estimation.
The main drawback of this method is that one OD matrix is defined for each modeled time interval, thus not considering the effect of traffic congestion between consecutive time intervals.

Pourmoradnasseri et al.~\cite{pourmoradnasseri_leveraging_2023} present an approach for dynamic trip-based route flow and OD matrix estimation. The authors start with a prior demand estimation from a normalized trip distribution. This input trip distribution is obtained from various sources such as travel surveys, population census, and mobile data. Then, they select the shortest paths between all OD pairs. After assigning the routes for all the OD pairs, they measure the discrepancy between real and simulated traffic counts obtained through simulation. A least square minimization procedure is used to calibrate the OD matrix using this difference. The authors validate this method on a model of the road network of the city of Tartu, Estonia, using traffic data collected by 33 IoT traffic sensors. The authors obtain a relative error of about 10\%, calculated as ${(\hat{y}-y)}/{|y|}$, where $\hat{y}$ and $y$ are the simulated and real traffic counts, respectively. Although the method is not vulnerable to the availability and accuracy of a prior OD demand, the main drawback is that it requires a prior probability distribution of trips between OD pairs based on data that are not always available. Moreover, the initial probability distribution of trips can affect the subsequent calibration step. A further drawback of the method concerns using the least square technique, which is highly sensitive to outliers and may not be suitable in contexts where the number of sensors providing traffic counts scales up significantly (hundreds).

Mehrabani et al.~\cite{mehrabani2023development} use a similar approach to calibrate a traffic model for the Belgian highway network. The authors define a probabilistic travel demand model using various data sources, including city populations, spatial distances between cities, yearly vehicle per kilometers traveled, and yearly truck trips. The accuracy of the traffic demand model is validated by comparing real travel times of heavy trucks between cities, achieving an accuracy level of $R^2 = 0.93$. The advantage of this approach is that it requires a limited amount of data to estimate intercity traffic demand. However, the validation is limited to the comparison of travel times of heavy trucks, which represent only a small part of the overall vehicular traffic.

Pamula et al.~\cite{PAMULA2023105550} propose an iterative method based on deep learning for estimating and predicting OD matrices. The authors start from a prior OD matrix and find a configuration of values that minimizes the deviation of the real traffic volumes in the road sections from the estimated values determined based on the assignment of the estimated OD matrix to the transport network. Then, the authors use LSTM layers and autoencoders to predict the values of the OD-pairs in a 15-minute time horizon. The LSTM layer learns the relationships between link volumes and the estimated OD pairs. The authors evaluate the proposed method using a road network model of Gliwice, Poland. The authors obtain OD matrices with an average error ranging from 6.97–7.31\% for a time horizon of 15 minutes. By using autoencoders instead, the average error ranges from 6.54\% to 7.18\%. As in previous methods, one of the main benefits of this approach is the estimation of OD matrices directly from the traffic flow data without a prior OD matrix. However, the proposed method does not consider the behavior of vehicles in a congested road network, and the distribution of the trips in the road network follows a proportional strategy, which can lead to unrealistic results.

Pamula et al.~\cite{PAMULA2023105550} propose an iterative method based on deep learning for estimating and predicting OD matrices. The authors start from a prior OD matrix and find a configuration of values that minimizes the deviation of the real traffic volumes in the road sections from the estimated values determined based on the assignment of the estimated OD matrix to the transport network. Then, the authors use LSTM layers and autoencoders to predict the values of the OD-pairs in a 15-minute time horizon. The LSTM layer learns the relationships between link volumes and the estimated OD pairs. The authors evaluate the proposed method using a road network model of Gliwice, Poland. The authors obtain OD matrices with an average error ranging from 6.97–7.31\% for a time horizon of 15 minutes. By using autoencoders instead, the average error ranges from 6.54\% to 7.18\%. This method exhibits the same limitations already discussed for Pamula et al.~\cite{PAMULA2023105550}'s approach.

This paper proposes an optimization method to estimate realistic simulated traffic models from traffic count data. The advantages of the proposed method, compared to the state-of-art methods, are as follows:

\begin{itemize}
    \item does not require pre-calibrated OD matrices. It calibrates traffic using only raw traffic count data;
    \item large-scale calibration of traffic simulation models in large temporal interval (24 hours);
    \item local traffic calibration: we regulate the amount of traffic locally in space and time, ensuring temporal coherence of traffic conditions.
\end{itemize}

What distinguishes the proposed method from state-of-the-art approaches is that the computation is partitioned locally both in space and time. Hence, reducing the computation complexity of the calibration. The computational load to calibrate traffic is specific to the size of the local part of the environment and the time interval. The possibility of calibrating traffic locally in space and time, ensuring temporal coherence of traffic conditions across consecutive time intervals, makes the proposed solution pertinent for calibrating traffic in nearly real-time. Furthermore, we make available the traffic model obtained using the calibration technique with real traffic counts.

\section*{Problem Statement}

Let $\regionSet = \{\rho_1,\rho_2, \dots ,\rho_k\}$ be a set of $k$ regions and $T=\{t_1,t_2, \dots, t_\ell\}$ the set of $\ell$ time instants (in seconds) when the traffic data is aggregated. We define the calibration problem as the minimization of the differences between the real and simulated amount of vehicles observed in each region and time interval:

\begin{equation}\label{opt:calibration_problem}
    \operatorname*{argmin}_\trafficModel \sum_{i=1}^{\ell-1} \sum_{j=1}^{k} \left|\avgRealTraffic(\region_j,\timeInterval_i) - \simulator(\region_j,\timeInterval_i;\trafficModel)\right|
\end{equation}

\noindent where $\timeInterval_i=(t_i,t_{i+1})$, $1 \leq i < \ell$ is a time interval, $\avgRealTraffic(\region_j, \timeInterval_i)>0$ is the average real traffic observed in the region $\region_j$, during the time interval $\timeInterval_i$, $\simulator(\region_j,\timeInterval_i;\trafficModel)>0$ is a function that takes in input the traffic model $\trafficModel$, the region $\region_j$ and the time interval $\timeInterval_i$, and returns the average amount of simulated vehicles observed in the region $\region_j$ during time interval $\timeInterval_i$.

We use Equation~\ref{opt:calibration_problem} to measure the global accuracy of the traffic model. During the calibration process, the difference $\left|\avgRealTraffic(\region_j,\timeInterval_i) - \simulator(\region_j,\timeInterval_i;\trafficModel)\right|$ is calculated for each region and time interval, and it is used to adjust the amount of local traffic in the simulation. Table~\ref{tab:symbols} summarizes the symbols associated with the main concepts used to introduce the proposed method.

\begin{table*}[!ht]
\centering
\caption{Symbols and notations}
\label{tab:symbols}
\resizebox{\textwidth}{!}{%
\begin{tabular}{|l|p{11cm}|}
\hline
\multicolumn{1}{|c|}{\textbf{Symbol}} & \multicolumn{1}{|c|}{\textbf{Description}} \\ \hline
\textbf{Route} $\route \subseteq \edgeSet$ & A route is an ordered set of edges from an origin to a destination. \\ \hline
\textbf{Region} $\region \in \regionSet$ & A region is a polygon that delimits a part of the road network.\\ \hline
\textbf{Regional route} $\regionRoute \subseteq \regionSet$& A regional path is an ordered set of regions connected by at least one route.\\ \hline
\textbf{Edge $e \in \edgeSet$} & An edge in SUMO models a directed connection between two junctions (or intersections).\\ \hline
\textbf{Traffic Model} $\trafficModel$ & A traffic model is a set of simulated vehicles. Each vehicle has information on the departure time and the route to go from the origin to the destination of its trip.\\ \hline
\textbf{Average Real Traffic} $\avgRealTraffic(\region, \timeInterval)$ & The average amount of real traffic observed in region $\region$ during the time interval $\timeInterval$.\\ \hline
\textbf{Average Simulated}\textbf{Traffic $\simulator(\region, \timeInterval; \trafficModel)$} & The average amount of vehicles observed in region $\region$ during the time interval $\timeInterval$, obtained by the traffic simulator $\simulator$ using the traffic model $\trafficModel$.\\ \hline
\textbf{Weighting function} $\weightFun: \edgeSet \rightarrow \mathbb{R}^{+}$ & A weighting function $\weightFun$ is used for discriminating between edges to include in vehicles' route.\\ \hline
\textbf{Time interval} $\timeInterval_i$ & A pair of time instants $(t_i,t_{i+1})$ that refers to a specific duration of time over which vehicle counts, collected by traffic monitoring devices, are summed. \\ \hline
\end{tabular}
}
\end{table*}

\section*{Proposed Method}\label{sec:proposal}

The proposed technique aims to estimate a traffic model $\trafficModel$ that minimizes the difference between real and simulated traffic counts. Herein, estimating a traffic model refers to constructing a set of routes that are associated each one to a vehicle and a starting time. The amount of traffic observed in the simulation should match the input data in the same points where real traffic is observed. To achieve this, the calibration process iteratively regulates the amount of vehicles and adjusts the routes to minimize the difference between real and simulated traffic counts. While our method does not estimate OD matrices (a computationally intensive task for high-dimensional problems) the resulting estimated routes can be used to infer origin-destination pairs.

We begin by partitioning the environment into a set $\regionSet$ of non-overlapping regions. Each region contains the edges that intersect the most with the polygon of the region. Mobility is strongly influenced by the morpho-spatial reconfiguration of cities, resulting in different mobility patterns. The goal of dividing the environment into regions is to capture mobility patterns across neighborhoods and calibrate traffic based on the physical characteristics of the urban landscape. The shape of regions is arbitrary. These can be modeled according to the administrative boundaries, or socio-economic indicators. The choice of shape can lead to the emergence of mobility patterns related to the properties of the urban environment.

After separating the environment in regions, we create simulated traffic in each region. First, we create a regional route for each vehicle by exploiting the local calibration error (average difference between real and simulated traffic counts). A regional route is defined as an ordered sequence of regions where a vehicle travels to reach its destination. A regional route $\regionRoute$ can be defined as~\cite{batista_regional_2019}:

\begin{equation}
\regionRoute=(\region_1, \dots,\region_k),~k \leq |\regionSet|
\end{equation}

\noindent where $\rho_1$ and $\rho_k$ are respectively the vehicle's origin and destination regions. Then, we allocate a vehicle to the regional route, and perform a route assignment procedure to evaluate the edges, in each region in $\regionRoute$, that allows the vehicle to reach its destination.

We iteratively adjust the traffic model $\trafficModel$ to match the real traffic count data. Because the impact of vehicles in a congested road network cannot be anticipated, we perform several simulations and regulate at each iteration the vehicles in the traffic model $\trafficModel$ so that the resulting traffic counts observed in the simulation match the real ones. We calculate the error between the real and the simulated traffic counts to evaluate our proposal. The traffic calibration is iterated for a fixed number of times until there is no improvement in the objective function (Equation~\ref{opt:calibration_problem}).

Following, we list the parameters required by the proposed calibration technique:

\begin{itemize}    
    \item \textbf{Initial traffic}: the traffic model $\trafficModel$ to calibrate.
    \item \textbf{Stopping criterion}: the stop condition for the calibration algorithm.
    \item \textbf{Vehicle starting time perturbation}: A random time value that can be added to or removed from the starting time of vehicles.
    \item \textbf{Re-routing probability}: The probability that a vehicle can change its route during the simulation.
    \item \textbf{Convergence rate}: it measures how quickly the algorithm should converge to an optimal or near-optimal assignment.
    \item \textbf{Maximum length of regional routes}: the maximum number of regions that can be included in the regional route of a vehicle.
    \item \textbf{Routing policy}: the criteria used to discriminate edges to include in vehicle's routes.
\end{itemize}

\subsection*{Traffic Model Initialization}\label{sec:traffic_init}

The first step of the proposed method is to initialize the traffic model $\trafficModel$ to calibrate. The traffic model can be initialized using a calibrated model from a different day, using a model created according to the number of vehicles in the input dataset, or randomly. In the following, we introduce a procedure to randomly initialize the traffic model $\trafficModel$. For the sake of simplicity, we do not consider regional routes to create the initial random traffic model. This is adjusted in the subsequent calibration phase.

At each instant $t$, we calculate the number of vehicles present in the simulation. Let $\gamma$ be the expected total number of vehicles that must be present in the simulation at each time instant. If the number of vehicles in the simulation is less than $\gamma$, then new vehicles are added to the simulation until the number of vehicles in the simulation is equal to $\gamma$. The number of vehicles in the initial traffic definition affects only the speed at which the proposed calibration method converges. Thus, we consider $\gamma$ a parameter required by the proposed calibration technique when the initial traffic model is created randomly.

To create simulated vehicles, we first choose randomly the origin and destination edges from the road network, then build a route using the Dijkstra algorithm~\cite{dijkstra}. The search graph corresponds to the road network, where the junctions are the vertex, and the roads are modeled as edges of the graph. The Dijkstra algorithm outputs a route, consisting of an ordered set of edges in the road network. In the Dijkstra algorithm, the edges are discriminated by their weight. We use the edge travel time to assign a weight to each edge~\cite{barbecho_bautista_evaluation_2020}. This is estimated by dividing the road length by the maximum allowed speed on the edge:

\begin{equation}\label{eq:weight_initial}
	\weightFun(e)=\frac{l(e)}{s_{max}(e)}
\end{equation}

\noindent where $l(e)$ is the length of the edge $e$, $s_{max}(e)$ is the maximum allowed speed on edge $e$. 

The output of this step is a traffic model $\trafficModel$ where each vehicle is associated with a unique identifier, a starting time, and a route. For the sake of simplicity, we do not consider the traffic light programs, the car-following model (describing the speed of cars as a function of their leading cars' position), the intersection model (determining the behavior of vehicles at the intersections), and the lane-changing model (determining the lane choice in a multi-lane road). Then, the traffic simulator outputs the number of vehicles observed in the same points where the traffic monitoring devices (and where the input traffic counts were obtained to calibrate traffic) are situated in reality.

\subsection*{Traffic Model Calibration}

We calibrate the traffic model $\trafficModel$ obtained from the previous step so that simulating the set of vehicles in $\trafficModel$ minimizes the difference between simulated and real traffic counts (Equation~\ref{opt:calibration_problem}). Algorithm~\ref{alg:calibration} lists the steps of the proposed calibration technique. This operates iteratively by calibrating the traffic model and evaluating it through simulation. This is because it is impossible to anticipate the effect of vehicles in a congested road network.

\begin{algorithm}
\caption{Traffic calibration}\label{alg:calibration}
\begin{algorithmic}[1]
\Require $\regionSet = \{\rho_1,\rho_2, \dots ,\rho_k\}$ : set of regions
\Require $\timeIntervalSet = \{\timeInterval_1,\timeInterval_2,\dots,\timeInterval_{\ell-1}\}$ : time intervals
\Require $\trafficModel \gets$ : traffic model
\Require $\weightFun : \edgeSet \rightarrow \mathbb{R}^{+}$ : edge weighting function
\Require $S$ : iterations
\Require $w$ : number of iterations without improving objective function

\State $\Gamma \leftarrow \left [ \; \right ]$ \Comment{calibration history}
\State $s \leftarrow 0$
\State $\text{continue\_calibrating} \leftarrow True$
\While{$\text{continue\_calibrating} \quad \wedge \quad s < S$}
    \For{$\timeInterval_i = (t_i,t_{i+1}) \in \timeIntervalSet$}\label{alg_a:update_model}
        \State $\errorSet = \{ \error_{\region} : \error_{\region} = \avgRealTraffic(\region,\timeInterval_i) - \simulator(\region, \timeInterval_i; \trafficModel), \forall \region \in \regionSet\}$ \label{alg_a:check_error}
        \State $d = \{d_{\region}: d_{\region} = \left|\error_{\region} \times \convergenceRate \right|, \forall \error_{\region} \in Err\}$ \label{alg_a:delta}
        
        \For{$\region \in \regionSet$}
            \If{ \Call{sign}{$\error_{\region}$} $< 0$  } \label{alg_a:check_overflow}	\Comment{\textit{Traffic overflow}}
                \For{$v \gets 1$~\textbf{to}~$d_{\region}$}
                    \State \Call{RemoveVehicle}{$\trafficModel, \region, \timeInterval_i$} \label{alg_a:remove_veh}
                \EndFor
            \EndIf
        \If{ \Call{sign}{$\error_{\region}$} $> 0$  }\label{alg_a:check_underflow} \Comment{\textit{Traffic underflow}}
            \For{$v \gets 1$~\textbf{to}~$d_{\region}$}\label{alg_a:loop_add_veh}
                \State \Call{AddVehicle}{$\regionSet, \trafficModel, \region, \weightFun$} \label{alg_a:add_veh} \Comment{\textit{Alg. \ref{alg:startAlgo}}}
            \EndFor \label{alg_a:loop_add_veh_end}
        \EndIf
        \State \Call{ScheduleStartingTimes}{$\trafficModel, \region, \timeInterval_i$} \label{alg_a:schedule_starting}
        \EndFor
    \EndFor \label{alg_a:update_model_end}
    
    \State $\gamma \leftarrow \sum_{i=1}^{\ell-1} \sum_{j=1}^{k} \left|\avgRealTraffic(\region_j,\timeInterval_i) - \simulator(\region_j,\timeInterval_i;\trafficModel)\right|$ \Comment{Simulation and objective evaluation (Eq. \ref{opt:calibration_problem})}\label{alg:objective}
    
    \State $G \leftarrow \left\{ g_i \right\}_{i=s-w}^{s-1}; \quad g_i \in \Gamma$ \Comment{Check objective improvement} 
    \State $\text{continue\_calibrating} \leftarrow \gamma < g; \quad \forall g \in G$
    \State $\Gamma \leftarrow \Gamma \cup \gamma$
    \State $s \leftarrow s + 1$
\EndWhile

\State \Return $\trafficModel$
\end{algorithmic}
\end{algorithm}

The algorithm takes as input a set $\regionSet$ of regions, the traffic model $\trafficModel$, a set of time intervals $\timeIntervalSet$, and an edge weighting function $\weightFun$. First, the algorithm computes the average traffic error $\error_{\region}$ for each region $\region \in \regionSet$ during the time interval \mbox{$\timeInterval_i = [t_{i},t_{i+1}]$} using the traffic model $\trafficModel$ (line~\ref{alg_a:check_error}). Then, we calculate the amount $d_{\region}$ of vehicles to add or remove from the region $\region \in \regionSet$, considering the error from the previous calibration iteration. This error is calculated as the difference between the average of the real and the simulated traffic counts in the region $\region$, multiplied by a convergence rate $\convergenceRate$ (line~\ref{alg_a:delta}). The value of $\convergenceRate$ affects the speed at which changes to the traffic model are made. A low value of $\convergenceRate$ yields limited changes to the traffic model, while using a high value, many vehicles are either added or removed from the model. 

The sign of $d_{\region}$ indicates how to calibrate the traffic: a negative sign indicates that the previous simulation had more vehicles than the expected amount passing through $\region$ (traffic overflow, line~\ref{alg_a:check_overflow}). A positive sign indicates that few vehicles passed through a region (traffic underflow, line~\ref{alg_a:check_underflow}). Based on the sign of $d_{\region}$ the algorithm removes or adds new vehicles to the traffic model $\trafficModel$. In case of traffic overflow in the region $\region$, we remove $d_{\region}$ vehicles starting their route from the same region $\region$ (line~\ref{alg_a:remove_veh}). This is not always possible since a traffic overflow may result from vehicles transiting through $\region$ to reach a destination region, but not starting their route from the same region. In this case, we remove a random vehicle passing through the region $\region$ during the time interval $\timeInterval_i$ with a probability $p=0.5$. In case of traffic underflow (line~\ref{alg_a:check_underflow}), we add $d_{\region}$ new vehicles into the traffic model $\trafficModel$ (lines~\ref{alg_a:loop_add_veh}-\ref{alg_a:loop_add_veh_end}) using Algorithm~\ref{alg:startAlgo}. 

Finally, we schedule the starting time of all vehicles in the region $\region$ and the time interval $\timeInterval_i$  (line~\ref{alg_a:schedule_starting}). The starting time of the vehicles is evenly distributed in the time interval $\timeInterval_i$, and perturbed by a value randomly chosen with a uniform distribution between 0 and 20 seconds. This is to avoid vehicles starting at fixed time intervals, which is unrealistic. Algorithm~\ref{alg:calibration} finishes when there are no improvements in the objective function after $w$ iterations or the budget of $S$ iterations has been completely exhausted.

Algorithm~\ref{alg:startAlgo} describes the steps to add a new vehicle in the traffic model $\trafficModel$. This algorithm starts by initializing a regional route (line~\ref{algo_add:region_empty}) containing the region $\rho_{start}$ where the new vehicle starts its trip, and the final route (line~\ref{algo_add:final_route}). The final route contains the ordered set of edges that allow the new vehicle to reach its destination. 

\begin{algorithm}
\caption{\textsc{AddVehicle} procedure}\label{alg:startAlgo}
\begin{algorithmic}[1]
\Require $\regionSet = \{\region_1,\region_2, \dots ,\region_k\}$ : set of regions
\Require $\trafficModel \gets$ : traffic model
\Require $\region_{start}$ : region where to add a new vehicle
\Require $\weightFun : \edgeSet \rightarrow \mathbb{R}^{+}$ : edge weighting function
\Require $m$ : regional route length
\Ensure  $\route \gets$ the route for the new vehicle
\State $\regionRoute \gets (\rho_{start})$	\Comment{\textit{regional route}}   \label{algo_add:region_empty}
\State $\route  \gets ()$	\Comment{\textit{final route}}  \label{algo_add:final_route}
\State $C  \gets \{ \region \in \regionSet : \region_{start} \nearbyRegionOp \region\}$    \label{algo_add:nearby_regions_0} \Comment{nearby regions}    
\While{$ |\regionRoute| < \maxRegionalRouteLen \wedge C \neq \emptyset$} \label{algo_add:loop_begin}
    \State $\regionRoute \gets \regionRoute \cup \Call{ChooseNearbyRegion}{C}$  \label{algo_add:add_rho}
    \State $C  \gets \{ \region \in \regionSet : \Phi_{j} \sim \region \wedge \region \notin \regionRoute \}$    \label{algo_add:nearby_regions} \Comment{nearby regions}    
\EndWhile \label{algo_add:loop_end}

\State $\pivotEdgeSet \gets \{e : e= \Call{ChoosePivot}{\region},~\forall \region \in \regionRoute\}$ \label{algo_add:choose_via} \Comment{\textit{pivot edges}}

\For{$j \gets 1~\text{\textbf{to}}~|\pivotEdgeSet|-1$} \label{algo_add:route_begin}   \Comment{\textit{build the final route}} 
    \State $e_j \gets \pivotEdgeSet(j)$
    \State $e_{j+1} \gets \pivotEdgeSet(j+1)$
    \State $\route \leftarrow \route\,\cup$ \label{algo_add:find_path}  \Call{FindRoute}{$e_j, e_{j+1}, \weightFun$}
\EndFor \label{algo_add:route_end}

\State $\trafficModel\gets \trafficModel~\cup $ \Call{Vehicle}{$\route$} \label{algo_add:add_veh}   \Comment{\textit{Add new vehicle to $\trafficModel$}} 
\end{algorithmic}
\end{algorithm}

At line~\ref{algo_add:nearby_regions_0} we evaluate the nearby regions of $\region_{start}$. We use the boolean operator $\nearbyRegionOp$ to determine if two regions are adjacent. At lines~\ref{algo_add:loop_begin}-\ref{algo_add:loop_end}, the algorithm builds a regional route by iteratively choosing regions until the number of regions in $\regionRoute$ is $\maxRegionalRouteLen$, or there is at least one region that can be added in $\regionRoute$ (that is until the set $C$ contains at least one element). We evaluate the value of $\maxRegionalRouteLen$ empirically. 

We use a standard exploration-exploitation trade-off to select iteratively the regions to insert in $\regionRoute$. At line~\ref{algo_add:add_rho}, the algorithm evaluates the nearby region $\region \in C$ with the largest average difference between real and simulated traffic count (obtained from the previous simulation) with a small probability (we set this to 0.01); otherwise, the algorithm selects a random region $\region \in C$. The region $\region$ is then added to the regional route $\regionRoute$. The regional route cannot contain the same region twice (cycles are not allowed). Then, we evaluate the set $C$ of regions near the last one added in $\regionRoute$ that are not already in $\regionRoute$ (line~\ref{algo_add:nearby_regions}). We indicate by $\regionRoute_{j}$ the last region in $\regionRoute$.

The next step is to build a route that passes through all the regions in $\regionRoute$. To do this, the algorithm selects one pivot edge for each region $\region \in \regionRoute$ (line~\ref{algo_add:choose_via}). A pivot edge denotes a specific edge within the road network where the new vehicle must flow. By using pivot edges, we impose that the route of the new vehicle passes exactly through the regions in $\regionRoute$. Herein, a pivot edge for a region $\region$ is one of the edges in the same region associated with a traffic monitoring device. To discriminate between edges that can be selected as pivots in the same region, the function $\Call{ChoosePivot}{}$ uses an exploration-exploitation trade-off to prioritize the pivot edges with the highest difference between simulated and real traffic counts. To do this, we choose random edges with a small probability (we set this to $0.01$), otherwise, we choose the edge that maximizes the difference between real and simulated traffic counts. At lines~\ref{algo_add:route_begin}-\ref{algo_add:route_end}, the algorithm calculates a route for each pair of consecutive pivot edges ($e_{j}, e_{j+1}$) in $\pivotEdgeSet$. The function \Call{FindRoute}{} finds a route for each pair of via edges in the set $\pivotEdgeSet$ using the Dijkstra algorithm (line~\ref{algo_add:find_path}). We use the following function $\weightFun$ to weigh the edges between two pivots, to include in the vehicle route:

\begin{equation}\label{eq:weight_occ}
	\weightFun(e)=\tau^{\timeInterval_i}(e)
\end{equation}

\noindent where $\tau^\timeInterval_i(e)$ is the average travel time measured in the edge $e$ during the $i$-th time interval $\timeInterval_i$ of the last simulation. Time travel estimates the time for a vehicle to pass an edge. Differently from Equation~\ref{eq:weight_initial}, here the travel time is estimated according to the average speed observed in the simulation. Finally, at line~\ref{algo_add:add_veh} the algorithm adds the new vehicle associated with the route $\route$ in the traffic model $\trafficModel$.

\subsection*{Complexity Analysis}

Analyzing the complexity of Algorithm \ref{alg:startAlgo}, line 12 is executed $\beta*m$ times, with a complexity of $O(\beta*m)$ following the Big O notation for the worst case, where $\beta$ corresponds to the computational complexity of Dijkstra’s algorithm and $m$ the regional route length. Following the complexity analysis in the worst case, the complexity of this code section dominates the complexity of the whole algorithm (w.r.t. the complexity of line \ref{algo_add:choose_via} $O(m)$ and block of lines \ref{algo_add:loop_begin}-\ref{algo_add:loop_end} $O(m)$). Hence, we can assume that the computational complexity of adding a vehicle is directly related to the length of the regional routes and the computational complexity of Dijkstra’s algorithm. As a result, the complexity of addressing traffic underflow in Algorithm \ref{alg:calibration} (lines \ref{alg_a:loop_add_veh}-\ref{alg_a:loop_add_veh_end}) is $O(\beta*m*log(underflow_p))$, where $underflow_p$ corresponds to the underflow error (the number of vehicles we must add) for region $p$ and it is assumed to decrease over time (hence the logarithmic scale). Given that addressing traffic overflow only means removing vehicles from the traffic model, its computational complexity $O(log(overflow_p))$ is much less than the complexity of treating traffic underflow. Consequently, the computational complexity of the block of lines \ref{alg_a:update_model}-\ref{alg_a:update_model_end} in Algorithm \ref{alg:calibration} is $O(l*k* \beta*m*log(underflow_p))$ which represents addressing traffic underflow on each region ($k$ in total) and on each time interval ($l$ in total) in the worst case. Lastly, the computational complexity of the complete calibration process can be represented as $S*O(\max\{ l*k*\beta*m*log(underflow_p), \; \mathbb{SIM}_\mathcal{M}\})$, where $S$ is the number of iterations and $\mathbb{SIM}_\mathcal{M}$ corresponds to the time needed by the algorithm to perform a traffic simulation with a given model $\mathcal{M}$.

In expensive black-box settings, such as traffic calibration where the simulation is typically assumed to be a black box, the computational complexity of an optimization algorithm can be evaluated through the \textit{worst-case expected running time} \cite{doerr2020complexity}. Then, the running time (or \emph{optimization time}) of an algorithm for a given function $f$ can be measured by the \emph{number of function evaluations} that the algorithm performs until (and including) the evaluation of an optimal solution for $f$. Considering this in the context of traffic calibration algorithms, we can assume that performing a traffic simulation will be always the most complex and demanding task during the calibration. Therefore, the overall complexity of the calibration can be summarized as $S*O(\mathbb{SIM}_\mathcal{M})$ and it is directly related to the number of traffic simulations performed during the calibration.

A similar analysis can be performed for the space complexity of the algorithm. The space complexity of Algorithm \ref{alg:startAlgo} is dominated by the space complexity of the Dijkstra algorithm, and the space needed to allocate the route of the new vehicle added to the model. The former can be assumed constant since the traffic network does not change during the calibration. However, the latter changes depending on the regions and pivot edges selected to make the regional route. The generated route is added to the model by keeping a record of all the routes obtained for each of the vehicles in the model. For algorithm \ref{alg:calibration}, the space complexity is mainly dominated by the objective evaluation at line \ref{alg:objective} which requires loading the real values and the simulation results, hence a space complexity of $2*l*k^2$, where $l$ is the number of time intervals and $k$ the number of regions. Adding or removing vehicles from the traffic model depends only on the traffic underflow or overflow respectively, and the result is reflected in the XML route files used as input for the simulation. This property demonstrates the advantage of our proposal given that it is a well-known challenge dealing with large and sparse OD matrices to perform the traffic calibration.

\section*{Experimental Design}\label{sec:results}

This section presents the results obtained by the proposed traffic calibration technique. First, we describe the scenarios and the error metrics used to evaluate our proposal. Then, we provide details on the results of the calibration technique and compare them to those obtained by one technique available with the SUMO simulator, and by a calibration technique based on a standard optimization method.

All experiments have been carried out on a server machine equipped with Intel Xeon Gold 6242 processors, and Linux operating system (kernel version 4.15). We do not consider any particular optimization technique to perform parallel computation. The proposed method calibrates traffic sequentially in each time slot. Contrarily, in the SPSA-based method the traffic in each time slot is calculated independently from the other. However, in the case of the latter method, the temporal correlation of traffic between consecutive time slots is not guaranteed.

\subsection*{Scenario Description}
\label{subsec:scenario}

\begin{figure}[!ht]
    \centering
    \begin{subfigure}[]{.45\textwidth}
        \centering
        \includegraphics[width=\textwidth]{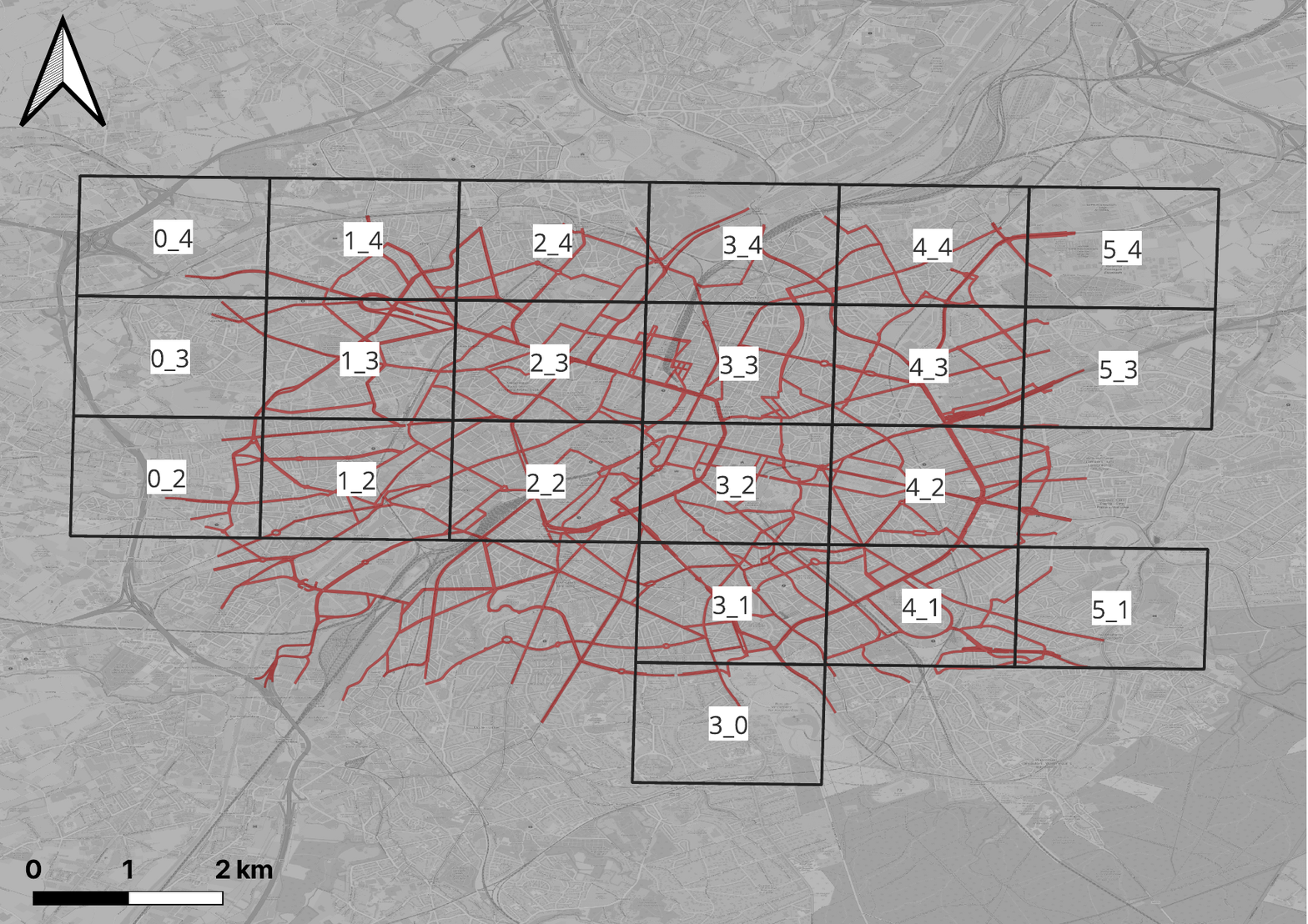}
        \caption{}
        \label{fig:region_2000}
    \end{subfigure}
    \hfill
    \begin{subfigure}[]{.45 \textwidth}  
        \centering 
        \includegraphics[width=\textwidth]{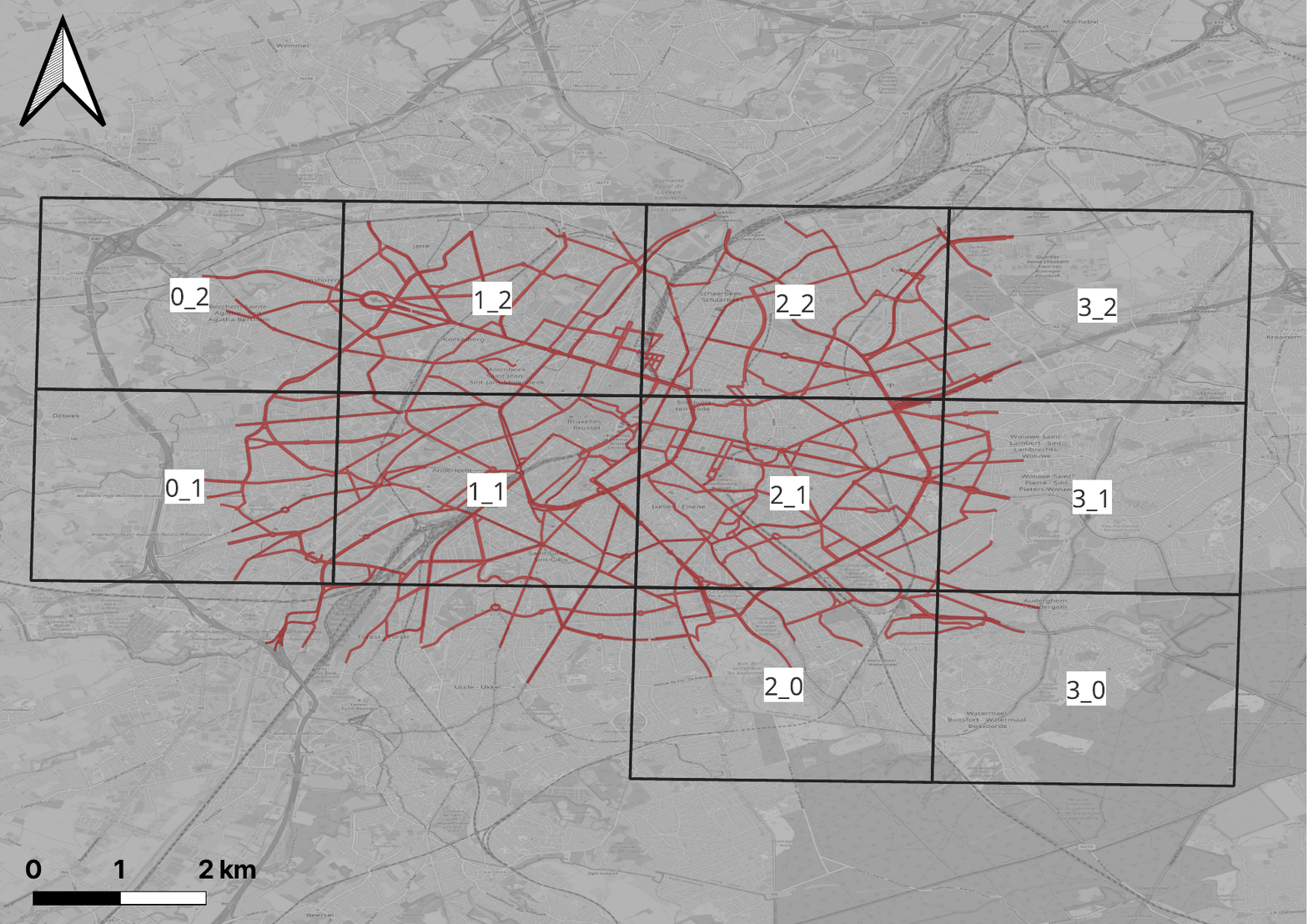}
        \caption{}
        \label{fig:region_3500}
    \end{subfigure}
    \vskip\baselineskip
    \begin{subfigure}[]{.45 \textwidth}  
        \centering 
        \includegraphics[width=\textwidth]{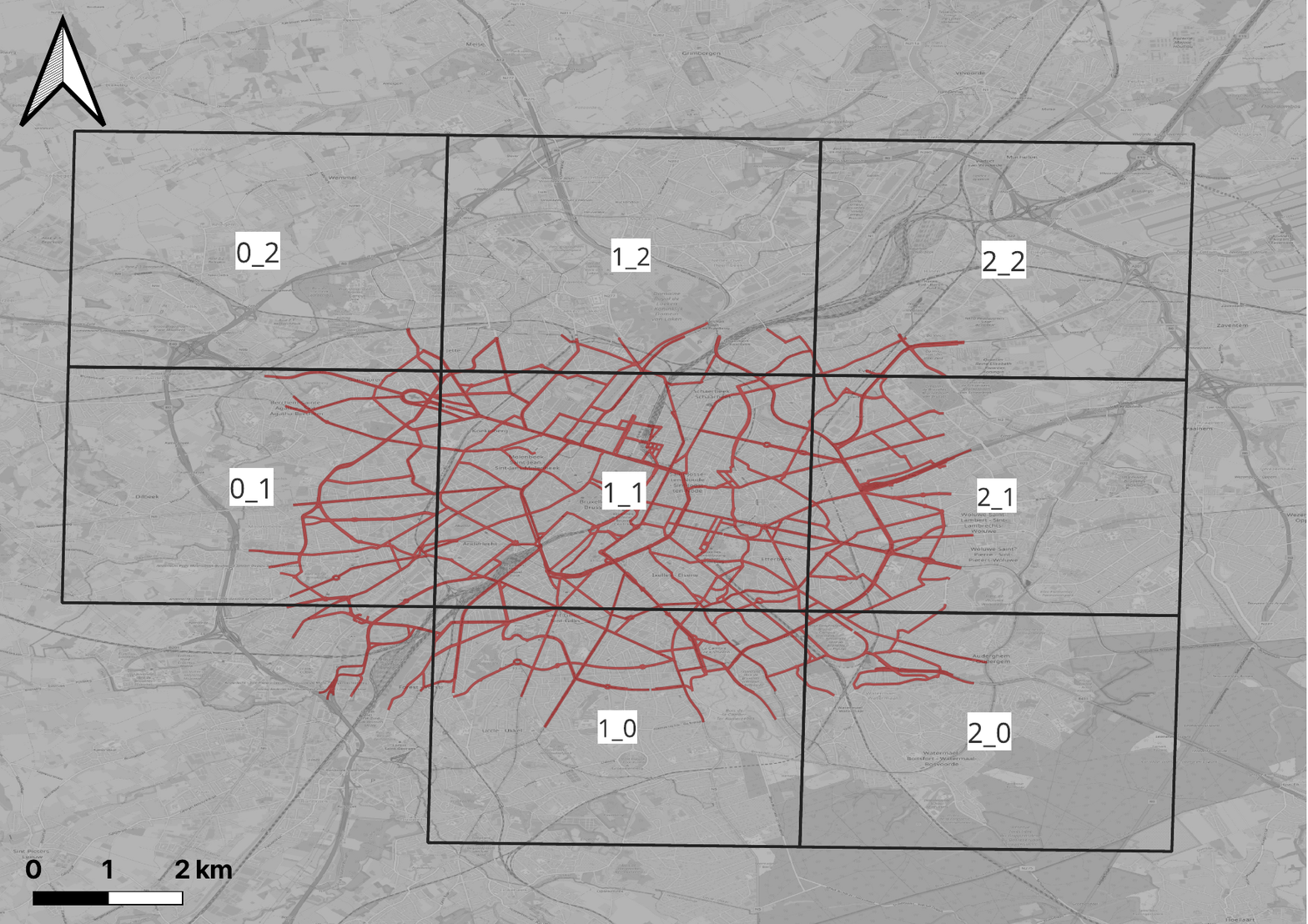}
        \caption{}
        \label{fig:region_5000}
    \end{subfigure}
    \caption{Modeled scenario in the city of Brussels, Belgium. We partition the environment using square regions of 2000m$^2$ (Figure~\ref{fig:region_2000}), 3500m$^2$ (Figure~\ref{fig:region_3500}) and 5000m$^2$ (Figure~\ref{fig:region_5000}).}
    \label{fig:bruxelles}
\end{figure}

We evaluate the proposed technique on a model of the city of Brussels (Belgium). The modeled area includes mostly primary roads (including the inner ring\footnote{\url{https://en.wikipedia.org/wiki/Small_Ring,_Brussels}. Last visited: \todayDate}), secondary and tertiary roads. For creating the road network model in SUMO, we first extract the area of interest using OpenStreetMap (OSM) and then, we use the tools available with SUMO to convert the OSM map into a convenient format that the simulator can use. Figure~\ref{fig:bruxelles} shows the considered road network for the Brussels scenario, considering square regions of 2000m$^2$, 3500m$^2$ and 5000m$^2$. Each road in the modeled network is assigned to a single region. A road is assigned to the first region whose polygon intersects with the polygon of the road.

We use traffic counts data collected from November 30, 2023 to December 3, 2023. Each dataset contains 24-hour data aggregated at 15-minute intervals. We evaluate our technique using regions of size 2000m$^2$, 3500m$^2$ and 5000m$^2$. We do not consider problems related to input data such as missing data or anomalies.

Given the large area to be modeled, we filter the data in the OSM file and extract only the roads where vehicles can circulate (removing railways, bike lanes, and pedestrian paths). The SUMO simulator provides the tool \Call{NetConvert}{} for converting OSM maps into a format that can be used by SUMO. This process relies on heuristics that approximate the original OSM attributes into connectivity features, particularly regarding the layout of intersections, rights of access, and the positioning and coordination of traffic light settings~\cite{argota_sanchez_vaquerizo_2021}.

\subsection*{Simulation Tool}

We use the open-source traffic simulator SUMO. SUMO can be used for microscopic simulation, where each vehicle and its dynamics are modeled individually, and mesoscopic simulation, where the movements of vehicles are modeled with queues and the traffic at intersections is modeled using a coarse model. We configure SUMO with a simplified microscopic model. This is done by simulating partially the behavior of vehicles in the intersections. When simulating traffic without considering the intersections, vehicles are still subject to right-of-way rules (waiting at traffic lights and minor roads) but they will appear instantly on the other side of the intersection after passing the stop line. The vehicles cannot block the intersection, wait within the intersection for left turns nor collide on the intersection\footnote{\url{https://sumo.dlr.de/docs/Simulation/Intersections.html\#internal_links}. Last visited: \todayDate}. In this work, we model junctions according to a microscopic model if the target queue of vehicles is jammed (that is, the occupancy of the edge is higher than 40\%). By using this configuration, the time required to perform a 24-hour simulation did not exceed 11 minutes, considering the Brussels scenario.

\subsection*{Evaluation Metrics}

We evaluate the accuracy of the proposed method using a 3-fold cross-validation: we divide the set of traffic monitoring devices into training (70\%) and testing (30\%) sets. For each fold, we calibrate the traffic model using the data from the sensors in the training set. Then, we evaluate the performance of the model considering the data collected from the devices in the test set.

To assess the performance of the simulated traffic, we evaluate the following metrics to compare the simulated traffic counts to the ground truth (the input data):

\begin{itemize}
    \item \textbf{Traffic volume}: the sum of the number of vehicles in the simulation and the ground truth. This measure is calculated in every region and time interval.
    \item{\textbf{Mean Absolute Error (MAE)}: the absolute difference between the number of vehicles in the simulation and the real data. This metric indicates the number of vehicles that are exceeding in the simulation compared to the reality. The MAE is calculated using the following formula:
    \begin{equation}
        \textnormal{MAE}(\hat{y},y) = \frac{1}{n} \sum_{i=1}^{n} \left| y_i-\hat{y_i} \right|
    \end{equation}
    \noindent where $n$ is the number of traffic monitoring devices, and $\simulator$ and $\avgRealTraffic$ the simulated and the real traffic volume respectively.
    }
    \item{\textbf{Root Mean Square Error (RMSE):} The square root of the sum of the squared differences between the predicted and observed traffic counts, divided by the number of observations:

    \begin{equation}
        \textnormal{RMSE}(\hat{y},y) = \sqrt{\frac{1}{n}\sum_{i=1}^{n}(y_i - \hat{y}_i)^2}
    \end{equation}
    
    \noindent where $n$ is the number of traffic monitoring devices, and $\simulator$ and $\avgRealTraffic$ the simulated and the real traffic volume respectively.
    }
    \item{\textbf{Geoffrey E. Harvers (GEH) statistic}: the GEH is used to compare two sets of traffic volumes and can be determined by using the following formula~\cite{dlr82548}:

    \begin{equation}
        \textnormal{GEH}(\hat{y},y) = \sqrt{\frac{2(\hat{y}-y)^2}{\hat{y}+y} }
    \end{equation}
    \noindent where $\simulator$ and $\avgRealTraffic$ are respectively the simulated and the real traffic volume.}
\end{itemize}

Because of the stochastic nature of the proposed method, we perform 50 experiments for each day (a day is associated with a specific data set). The values of the parameters required by the calibration technique are as follows:

\begin{itemize}    
    \item \textbf{Initial traffic}: we generate a random set of vehicles.
    \item \textbf{Stopping criterion}: the calibration technique stops if at least one of the following conditions is verified: (\textit{i}) the number of iterations is over 20 ($S=20$), or (\textit{ii}) there is no improvement in the objective function over the last 3 calibration iterations ($w=3$).
    \item \textbf{Vehicle starting time perturbation}: the starting time of the vehicles (that is, the time instant when they are inserted into the simulation) is randomly perturbed by a maximum of 20 seconds.
    \item \textbf{Re-routing probability}: each vehicle can change its route during the simulation with a probability $p$. Then, a new path is calculated using the average travel time of the edges during the last 100 seconds of the simulation. The parameter was varied in the discrete set $p=[0.2, 0.8]$.
    \item \textbf{Convergence rate}: it measures how quickly the algorithm should converge to an optimal or near-optimal assignment. The parameter was varied in the discrete set $q=[0.05, 0.15]$. 
    \item \textbf{Maximum length of regional routes}: the maximum number of regions that can be included in a regional route. We set this value to 5 for all vehicles.
    \item \textbf{Routing policy}: the criteria for discriminating edges that make up routes. We use the attribute travel time to discriminate edges and build routes for vehicles in calibration steps.
\end{itemize}


\subsection*{Baseline Techniques}

We compare the results obtained by the proposed technique to those obtained by RouteSampler\footnote{\url{https://sumo.dlr.de/docs/Tools/Turns.html}. Last visited: \todayDate}, a tool available with SUMO, and to those obtained by a calibration method based on the Simultaneous Perturbation Stochastic Approximation (SPSA) optimization technique.

RouteSampler takes as input a random traffic definition and filters the routes to match the real traffic density values. The traffic density is used to indicate the usual amount of traffic that should pass through the edges of the road network. Based on this information, the tool selects the routes from an initial set of trips in such a way that the input traffic density values are matched to the output routes set~\cite{emode23}. We used the default parameters for RouteSampler.

SPSA is an optimization technique widely employed in scenarios where traditional gradient-based methods are unsuitable because of the high number of variables. SPSA estimates gradients through stochastic parameter perturbations, assessing the objective function at these perturbed positions. By leveraging finite-difference approximations, SPSA explores the parameter space to converge towards the optimal solution. The idea behind SPSA is as follows: let us consider an initial vector for the parameters $\theta_0$, and set other hyper-parameters such as the step size $\Delta$, the number of iterations $k$, and the perturbation scaling factor $a_k$. At each iteration $k$, SPSA generates a random perturbation vector $\Delta_k$. The objective function is then evaluated at two points: $J(\theta_k + c_k \Delta_k)$ and $J(\theta_k - c_k \Delta_k)$, where $c_k$ is a scaling factor. The next step is to approximate the gradient of the objective function using the perturbation:

\begin{equation}
    \nabla J(\theta_k) \approx \frac{J(\theta_k + c_k \Delta_k) - J(\theta_k - c_k \Delta_k)}{2 c_k \Delta_k}
\end{equation}

Then, the parameters are updated using the gradient approximation as follows:

\begin{equation}
    \theta_{k+1} = \theta_k - \alpha_k \nabla J(\theta_k)
\end{equation}

\noindent where $\alpha_k$ is the step size. The algorithm is repeated until the maximum number of iterations (100) is reached.

In our experiments, we first generate random traffic. Based on the simulated vehicles, we evaluate all the origin-destination pairs. Each variable of the parameter vector $\theta$ in SPSA represents one OD pair, where the origin and the destination of vehicles are modeled as regions. By considering 30 regions (each of size 2000m$^2$), the number of parameters for $\theta$ is $\binom{30}{2}=435$. Here, we assume that all region pairs have a route connecting them. We choose the initial values for $\theta$ as the average amount of real traffic in each time interval to calibrate, divided by the number of regions. We also set the value of the step size as $\alpha_k =30$, evaluated empirically. 

\subsection*{Experimental Results}

Herein, we report the results obtained by the proposed method on the Brussels data acquired the November 30, 2023, using regions of size 2000$m^2$, a re-routing factor of $0.2$, and a convergence rate value of $0.15$. We calculate the results as the average of the 3-fold validation errors. The results for the remaining days and regions' sizes are available in Appendix~\ref{appendix:results_dec1}, Appendix~\ref{appendix:results_dec2}, and Appendix~\ref{appendix:results_dec3}.

\begin{figure}[!ht]
    \centering
    \includegraphics[width=.5\textwidth]{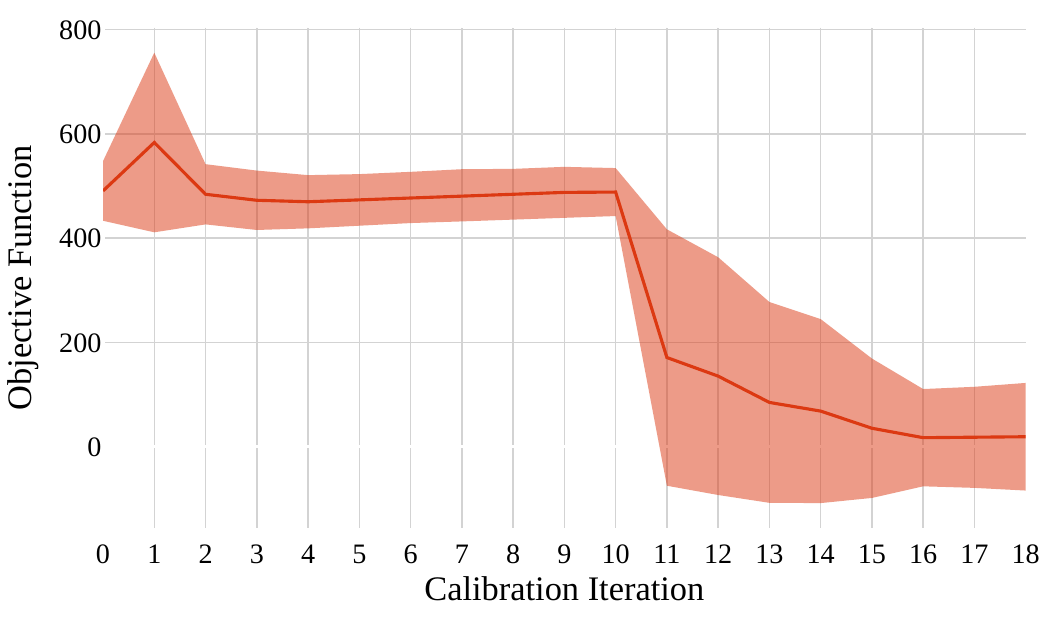}
    \caption{Average and standard deviation of the objective function (Equation~\ref{opt:calibration_problem}) obtained by the proposed technique.}
    \label{fig:obj_fun}
\end{figure}

Figure~\ref{fig:obj_fun} shows the trend of the objective function over the number of calibration iterations. This is calculated as the average of the value of the objective function over the 50 experiments performed. The objective function tends to converge after a few iterations (about 10), in all the experiments. This indicates that by using the proposed technique, it is possible to compute realistic traffic simulation models in a limited time (minutes). However, few calibration iterations can lead to local optima. By allowing the objective function to increase (therefore, introducing errors) it is possible to escape local optima at the cost of a higher computational time required by the technique.

\begin{figure}[!ht]
    \centering
    \includegraphics[width=.5\textwidth]{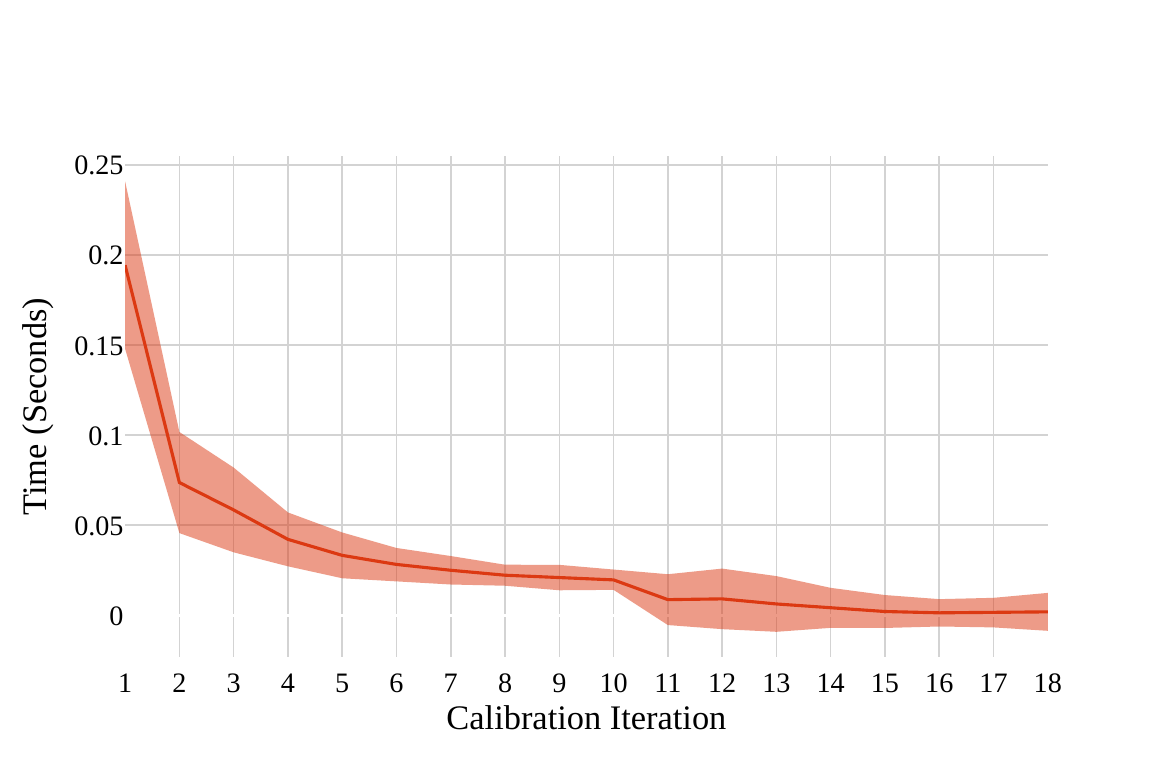}
    \caption{Computational time (seconds) required to calibrate traffic using the proposed method.}
    \label{fig:calib_time_3011}
\end{figure}

Figure~\ref{fig:calib_time_3011} shows the average and the standard deviation of the time required to calibrate simulated traffic in each region and iteration using the proposed method. We calculate the average considering the time required to calibrate traffic in all the experiments and folds. The average time required to calibrate a single region using the proposal region is below 1 second. The capability to regulate simulated traffic in near real-time enables the proposed technique to be used to define digital twin environments for traffic management. Data streams from traffic monitoring sources can be injected into the system to create a reproduction of urban traffic. This approach allows for the evaluation of changes in urban infrastructure within a synthetic environment before the implementation in the real world.

\begin{figure}[!ht]
    \centering
    \includegraphics[width=.5\textwidth]{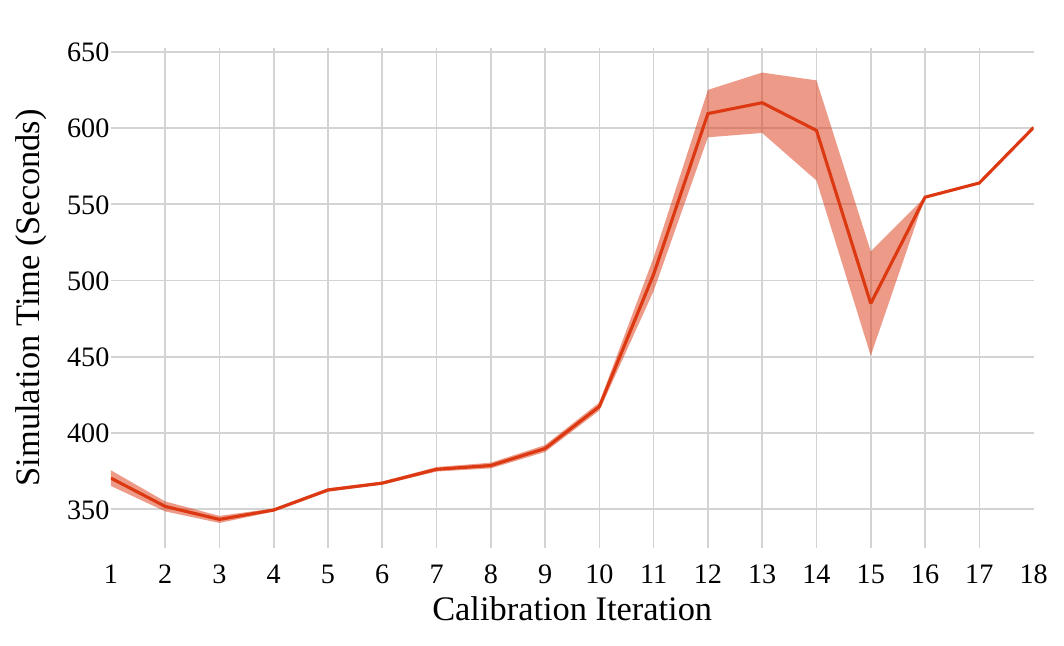}
    \caption{Computational time (seconds) required to simulate traffic using the proposed method.}
    \label{fig:sim_time}
\end{figure}

For each experiment, the number of iterations may vary as the proposed method is stochastic. Consequently, the uncertainty around the simulation time is reduced when the number of experiments reaching an iteration increases. In Figure~\ref{fig:sim_time}, we show the time required to simulate traffic at each iteration of the proposed method. Here, the uncertainty around the $i$-th iteration is shown as $y_i \pm \frac{\text{std}}{\sqrt{n_i}}$, where $y_i$ is the average time required to simulate traffic and $n_i$ is the number of experiments at each iteration $i$. Most algorithms finished the calibration at iteration 10 (thus the small time variability up to this iteration), which corresponds with the moment when the objective appears to reach the optimum value (Figure \ref{fig:obj_fun}).

\begin{figure}
    \centering
    \begin{subfigure}[b]{.49\textwidth}
        \centering
        \includegraphics[width=\textwidth]{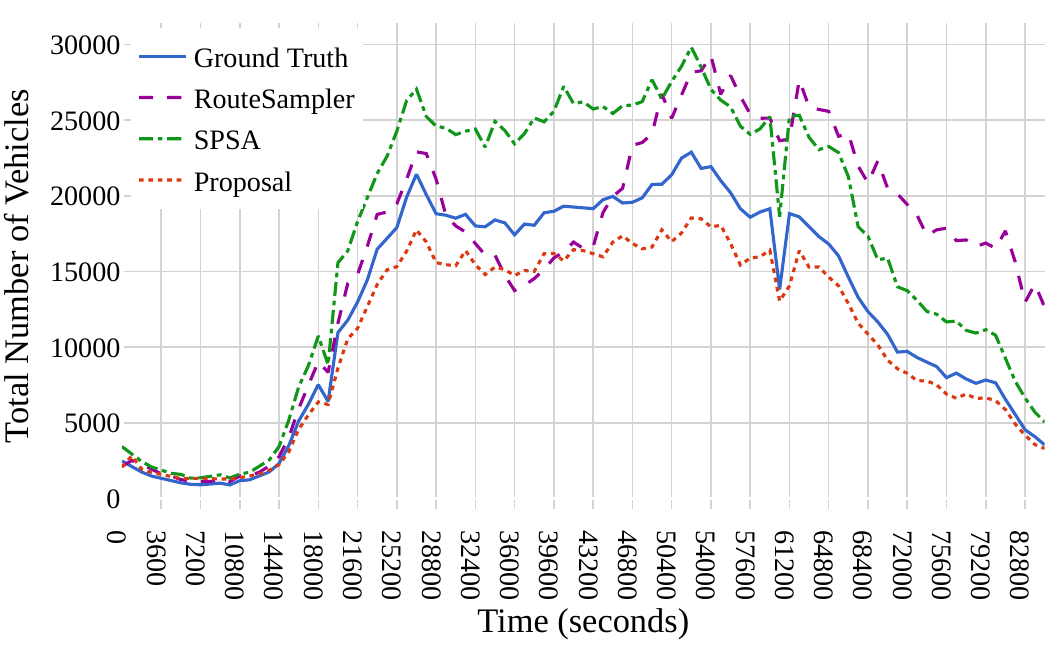}
        \caption{}
        \label{fig:traffic_vol_train}
    \end{subfigure}
    \begin{subfigure}[b]{.49\textwidth}
        \centering
        \includegraphics[width=\textwidth]{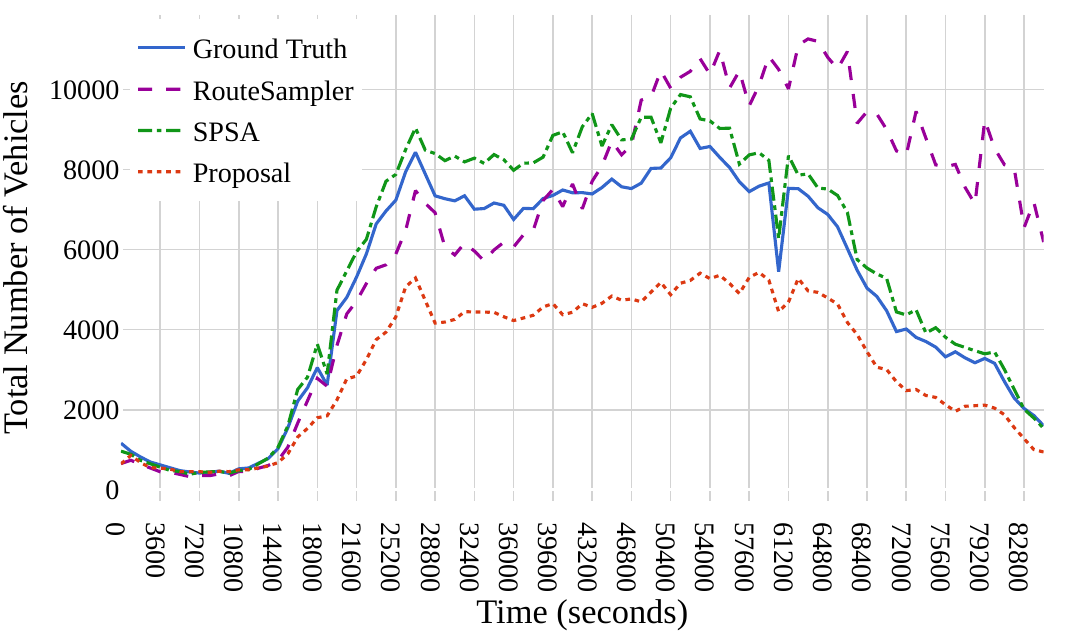}
        \caption{}
        \label{fig:traffic_vol_test}
    \end{subfigure}
    \caption{Total number of vehicles observed in all the regions and every time interval, in both the ground truth and the simulation. Figure~\ref{fig:traffic_vol_train} and Figure~\ref{fig:traffic_vol_test} show the total number of vehicles obtained on the training and testing sensors respectively.}
    \label{fig:traffic_vol}
\end{figure}

\begin{figure}
    \centering
    \begin{subfigure}[b]{.48\textwidth}
        \centering
        \includegraphics[width=\textwidth]{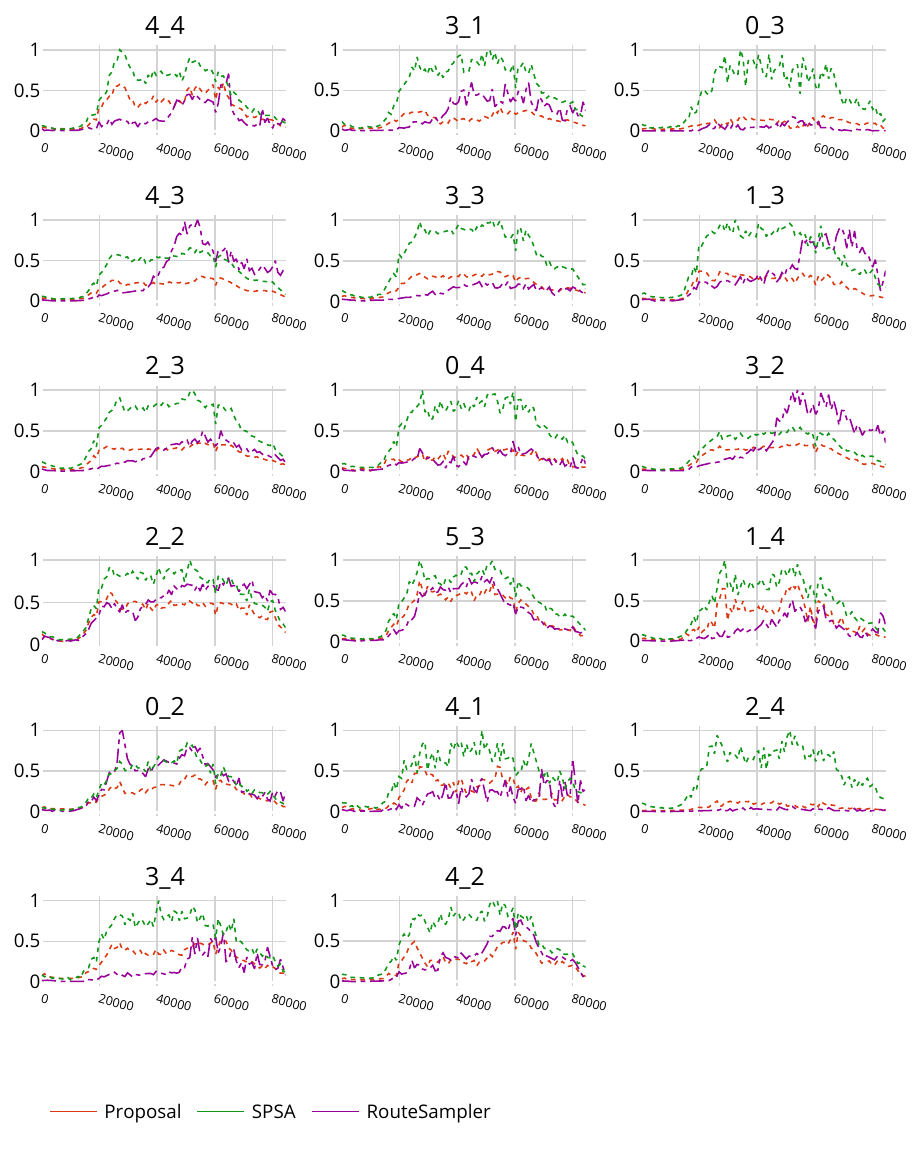}
        \caption{}
        \label{fig:region_vol_train}
    \end{subfigure}
    \hfill
    \begin{subfigure}[b]{.48\textwidth}
        \centering
        \includegraphics[width=\textwidth]{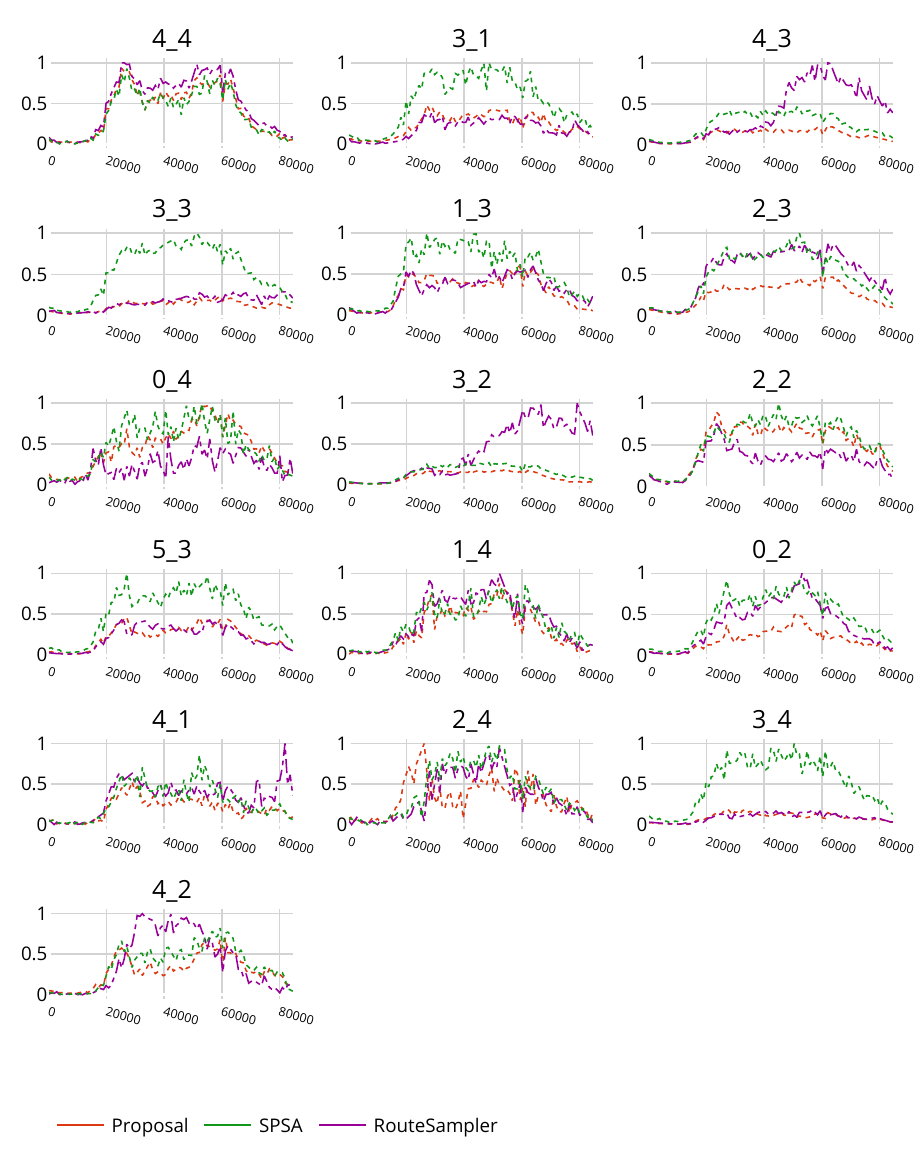}
        \caption{}
        \label{fig:region_vol_test}
    \end{subfigure}
    \caption{Local normalized average absolute traffic counts difference (MAE), calculated on the training set (Figure~\ref{fig:region_vol_train}) and test set (Figure~\ref{fig:region_vol_test}). The title of each figure indicates a region, as shown in Figure~\ref{fig:bruxelles}.}
    \label{fig:region_vol}
\end{figure}

Figure~\ref{fig:traffic_vol} compares the volume of traffic per second obtained by our proposal and the baseline methods. This is obtained by calculating, for each time interval, the sum of the vehicles in the simulation and the reality. The RouteSampler method underestimates the amount of traffic compared to the actual one. While the SPSA-based method outputs a good global fit of the traffic on the training sensors set, the local error (Figure~\ref{fig:region_vol}) is in most regions higher compared to that obtained by the proposed method. The proposed method regulates the number of vehicles proportionally to the local error, while in SPSA we fix the increment value for each variable. This results in a higher number of calibration iterations compared to the proposed technique, and consequently a slower convergence. Finally, the accuracy of SPSA is significantly affected by the initial solution and the increment value for the variables.

Figure~\ref{fig:region_vol} shows the absolute traffic counts difference (MAE), normalized with respect to the maximum error value, obtained after calibrating the traffic model using the proposed and the baseline techniques. The proposed method obtains a lower local error compared to SPSA. Since our proposal considers the effect of traffic during different time intervals, the dynamics of the generated simulated traffic are closer to the real one, compared to that of SPSA.

\begin{figure}[!ht]
    \centering
    \begin{subfigure}[b]{.49\textwidth}
        \centering
        \includegraphics[width=\textwidth]{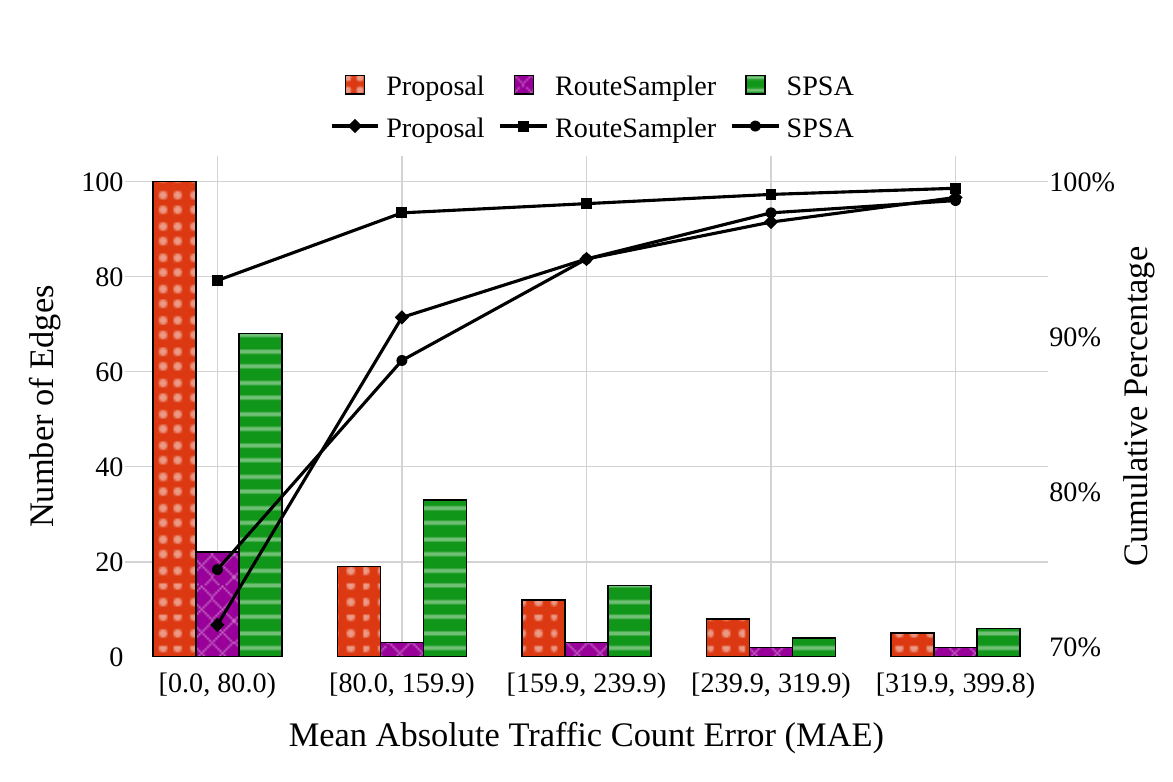}
        \caption{}
        \label{fig:pareto_train}
    \end{subfigure}
    \hfill
    \begin{subfigure}[b]{.49\textwidth}
        \centering
        \includegraphics[width=\textwidth]{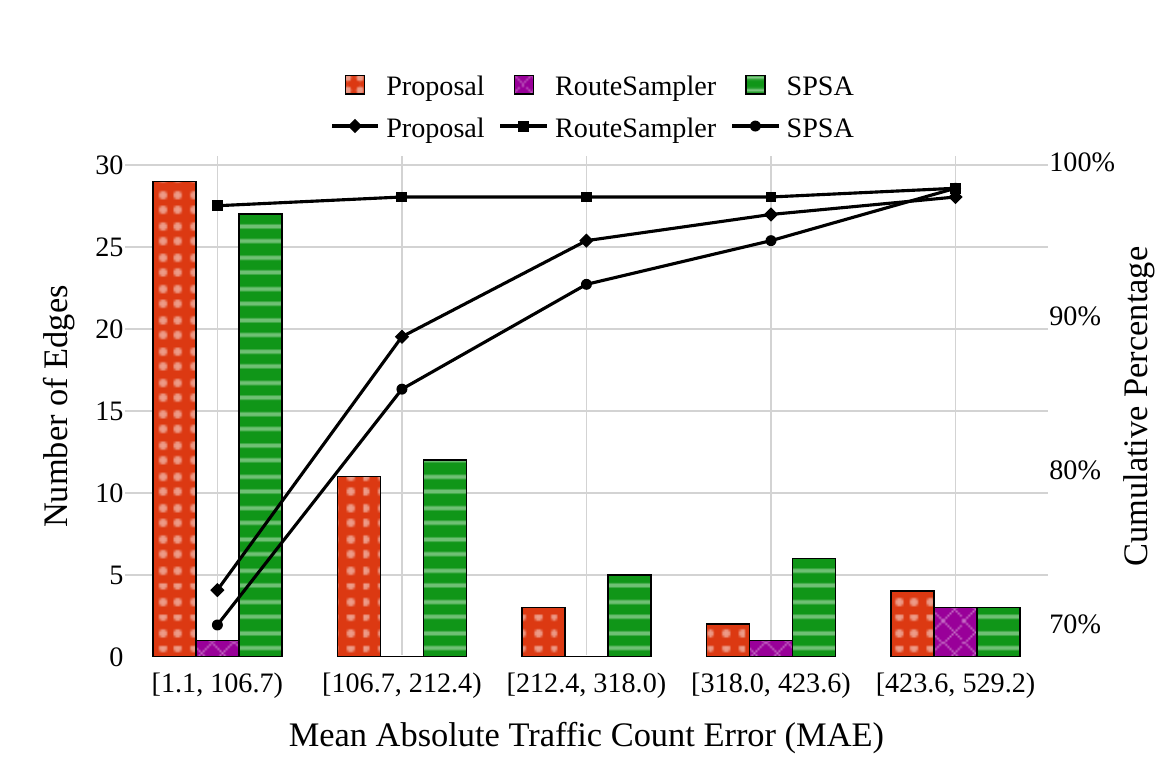}
        \caption{}
        \label{fig:pareto_test}
    \end{subfigure}
    \caption{Pareto chart grouping the training sensors set (Figure~\ref{fig:pareto_train}) and the test sensors set (Figure~\ref{fig:pareto_test}) according to average absolute traffic counts error (MAE).}
    \label{fig:pareto}
\end{figure}

Figure~\ref{fig:pareto} shows the Pareto chart~\cite{wilkinson_revising_2006} clustering the sensors according to the average absolute difference between real and simulated traffic counts (MAE). Each bin contains the number of edges in the road network equipped with sensors whose error falls into a specific range. The line represents the cumulative percentage of the total number of occurrences for each calibration technique. In both cases, the curve of cumulative percentage for the proposed method decreased more slowly than SPSA and RouteSampler. The average MAE over the 3 folds, calculated as the absolute difference between the number of vehicles in the ground truth and the simulation is 7.37 using our proposal, 8.54 using SPSA, and 8.58 using RouteSampler. Using the test set, the average MAE calculated over the 3 folds is 12.9 using our proposal, 3.4 using SPSA, and 11.2 using RouteSampler. This error is calculated by averaging the MAE in all time intervals and regions.

\begin{figure}[!ht]
    \centering
    \begin{subfigure}[]{.49\textwidth}
        \centering
        \includegraphics[width=\textwidth]{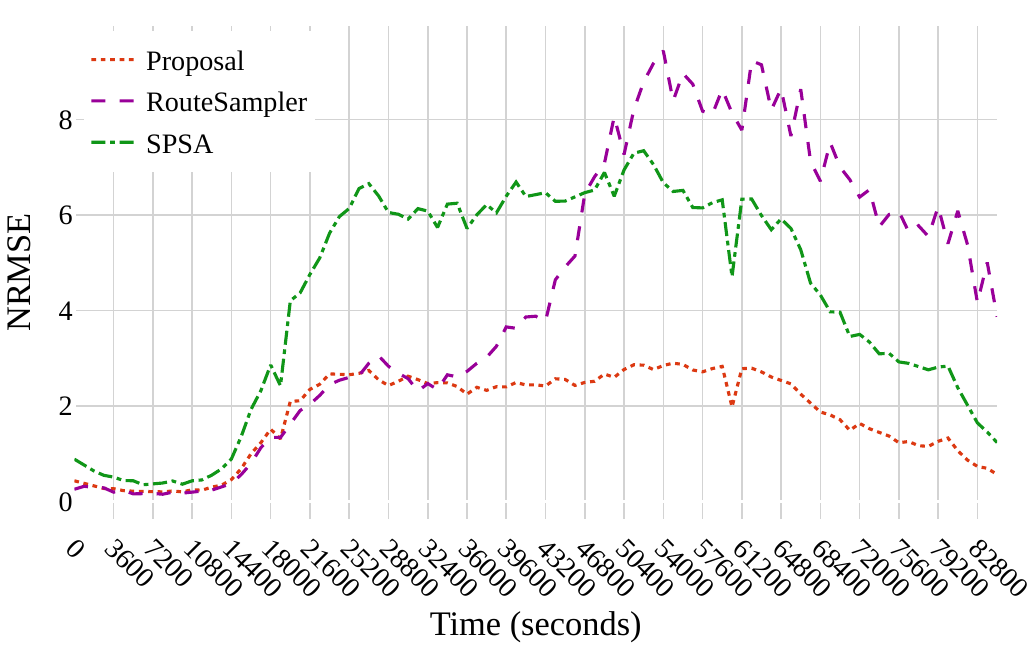}
        \caption{}
        \label{fig:nrmse_time_ts_train}
    \end{subfigure}
    \begin{subfigure}[]{.49\textwidth}  
        \centering 
        \includegraphics[width=\textwidth]{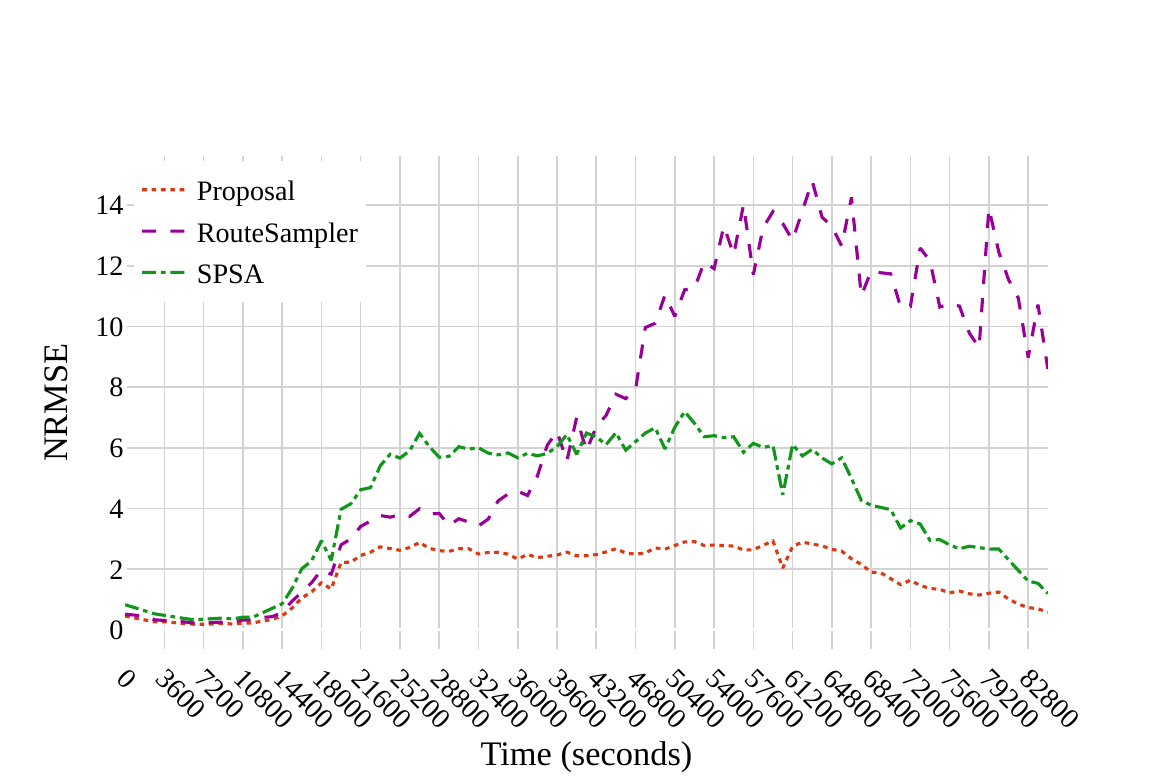}
        \caption{}
        \label{fig:nrmse_time_ts_test}
    \end{subfigure}
    \caption{Normalized RMSE, calculated using the training set (Figure~\ref{fig:nrmse_time_ts_train}) and the test set (Figure~\ref{fig:nrmse_time_ts_test}).}
    \label{fig:nrmse_time_ts}
\end{figure}

In Figure~\ref{fig:nrmse_time_ts}, we remark that SPSA outputs the highest error using the normalized RMSE metric. This is because of the hyperparameters configuration of the SPSA technique. To tackle this issue, it is necessary a trade-off between step size and the convergence rate. However, hyperparameter optimization in the presence of a high number of variables can be computationally expensive. 

Figure~\ref{fig:rmse} compares the normalized RMSE by time and sensors. In the former case (Figure~\ref{fig:rmse_time_train}-\ref{fig:rmse_time_test}), we calculate the RMSE by aggregating the traffic count differences by sensors, and normalizing the results by the average of traffic count differences for each sensor. 
In the latter case (Figure~\ref{fig:rmse_sensor_train}-\ref{fig:rmse_sensor_test}), we aggregate the traffic count differences by time interval, and normalize the results by the average of traffic count differences over time. Because RMSE gives more weight to larger errors due to the squaring operation, these results show that our technique is less sensitive (smaller RMSE values) to outliers compared to the baseline. This result is confirmed by calculating the normalized RMSE over time, shown in Figure~\ref{fig:nrmse_time_ts}. This is obtained by considering the RMSE calculated over time, normalized by the standard deviation of the real traffic counts.

\newcommand{\boxplotsize}{.3}
\begin{figure}[!ht]
    \centering
    \begin{subfigure}[b]{\boxplotsize\textwidth}
        \centering
        \includegraphics[width=\textwidth]{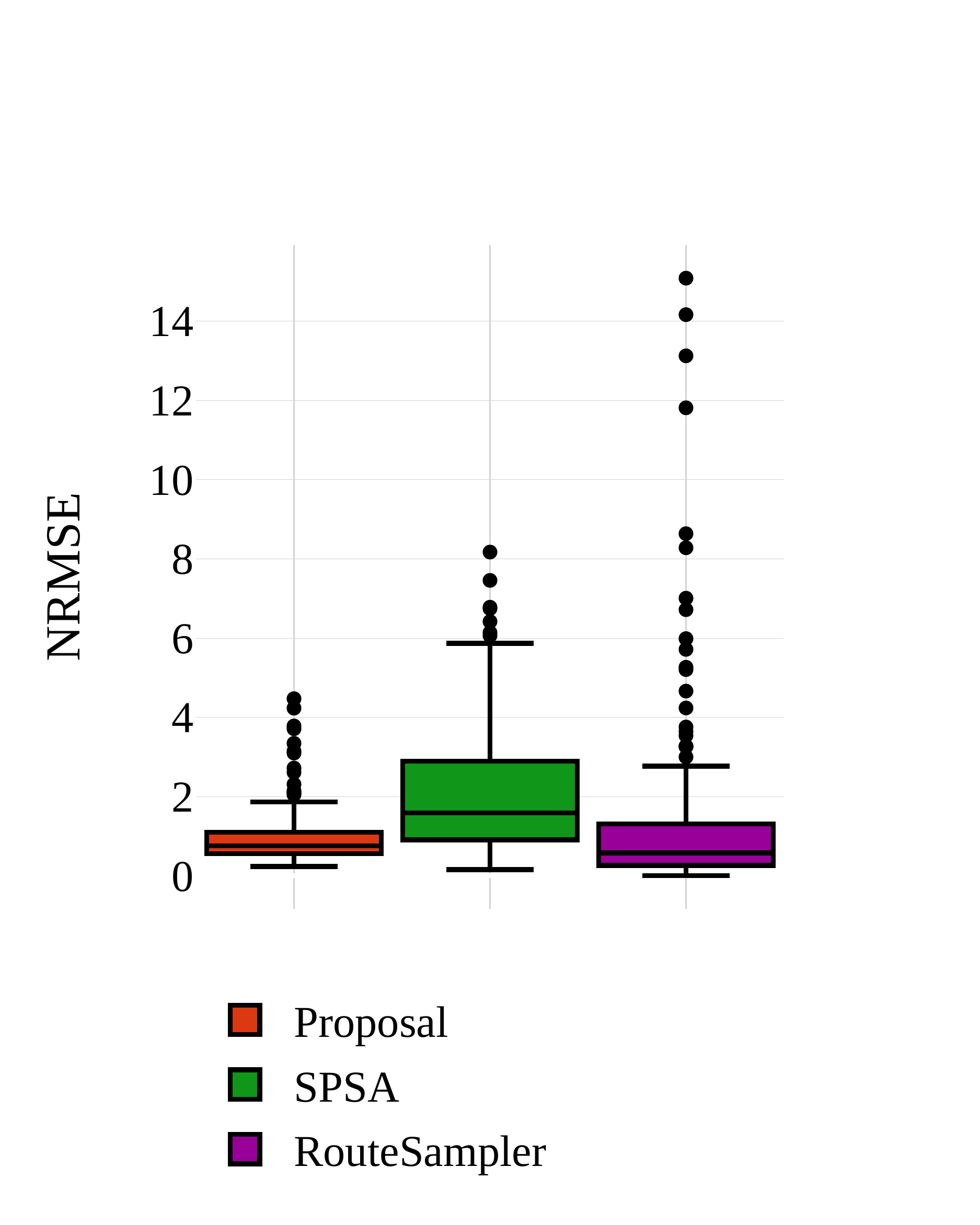}
        \caption{}
        \label{fig:rmse_sensor_train}
    \end{subfigure}
    \begin{subfigure}[b]{\boxplotsize\textwidth}
        \centering
        \includegraphics[width=\textwidth]{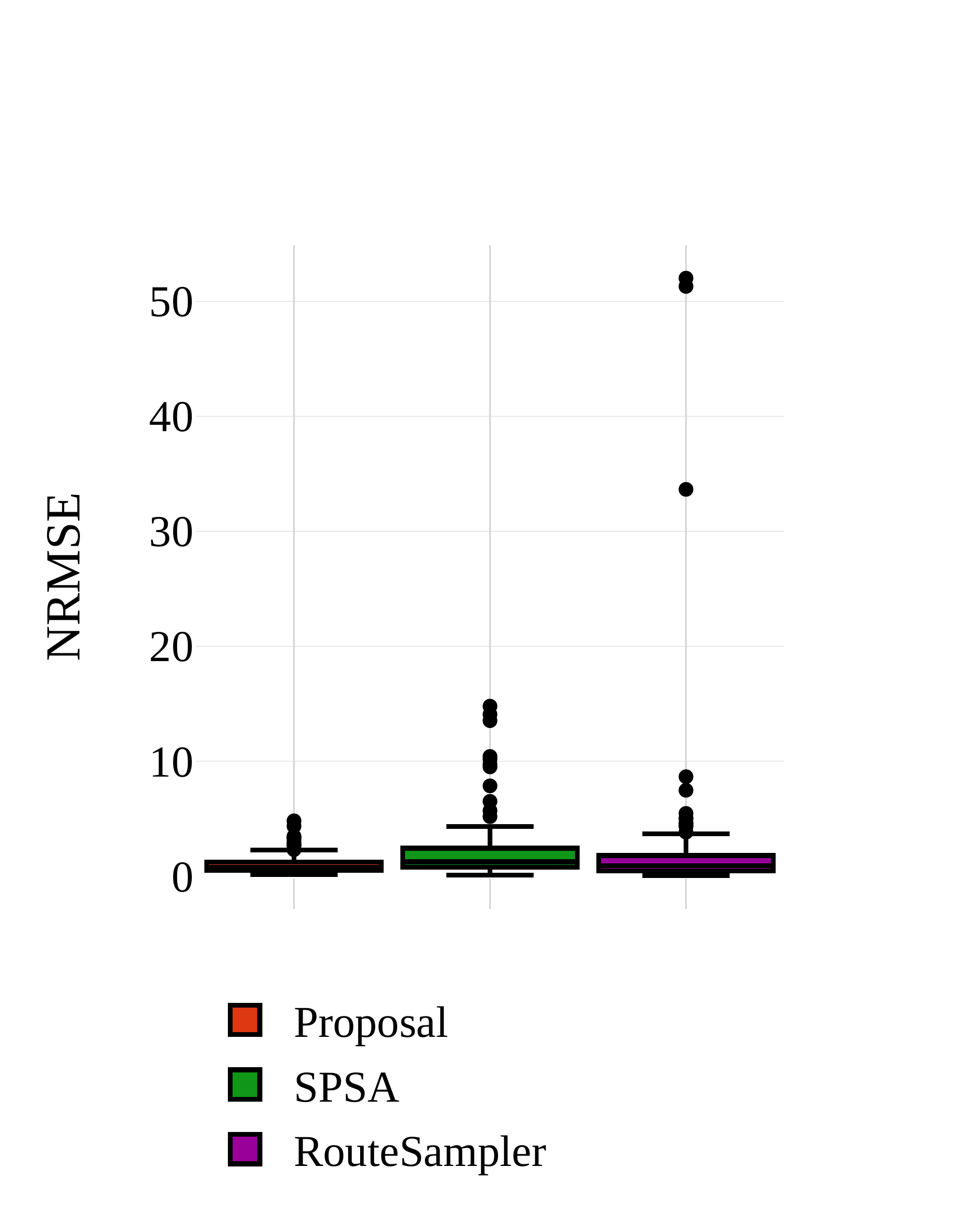}
        \caption{}
        \label{fig:rmse_sensor_test}
    \end{subfigure}
    \vskip\baselineskip
    \begin{subfigure}[b]{\boxplotsize\textwidth}
        \centering
        \includegraphics[width=\textwidth]{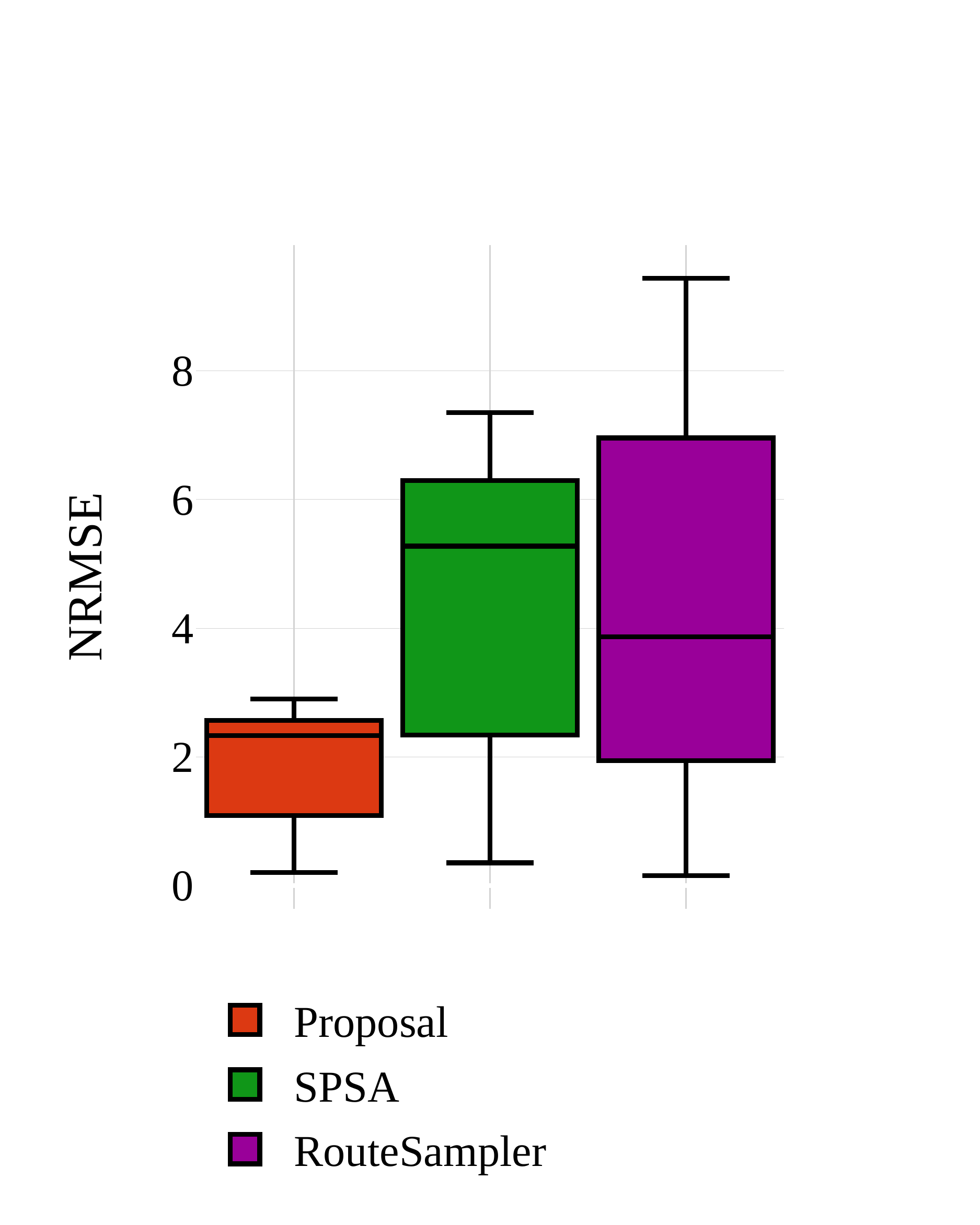}
        \caption{}
        \label{fig:rmse_time_train}
    \end{subfigure}
    \begin{subfigure}[b]{\boxplotsize\textwidth}
        \centering
        \includegraphics[width=\textwidth]{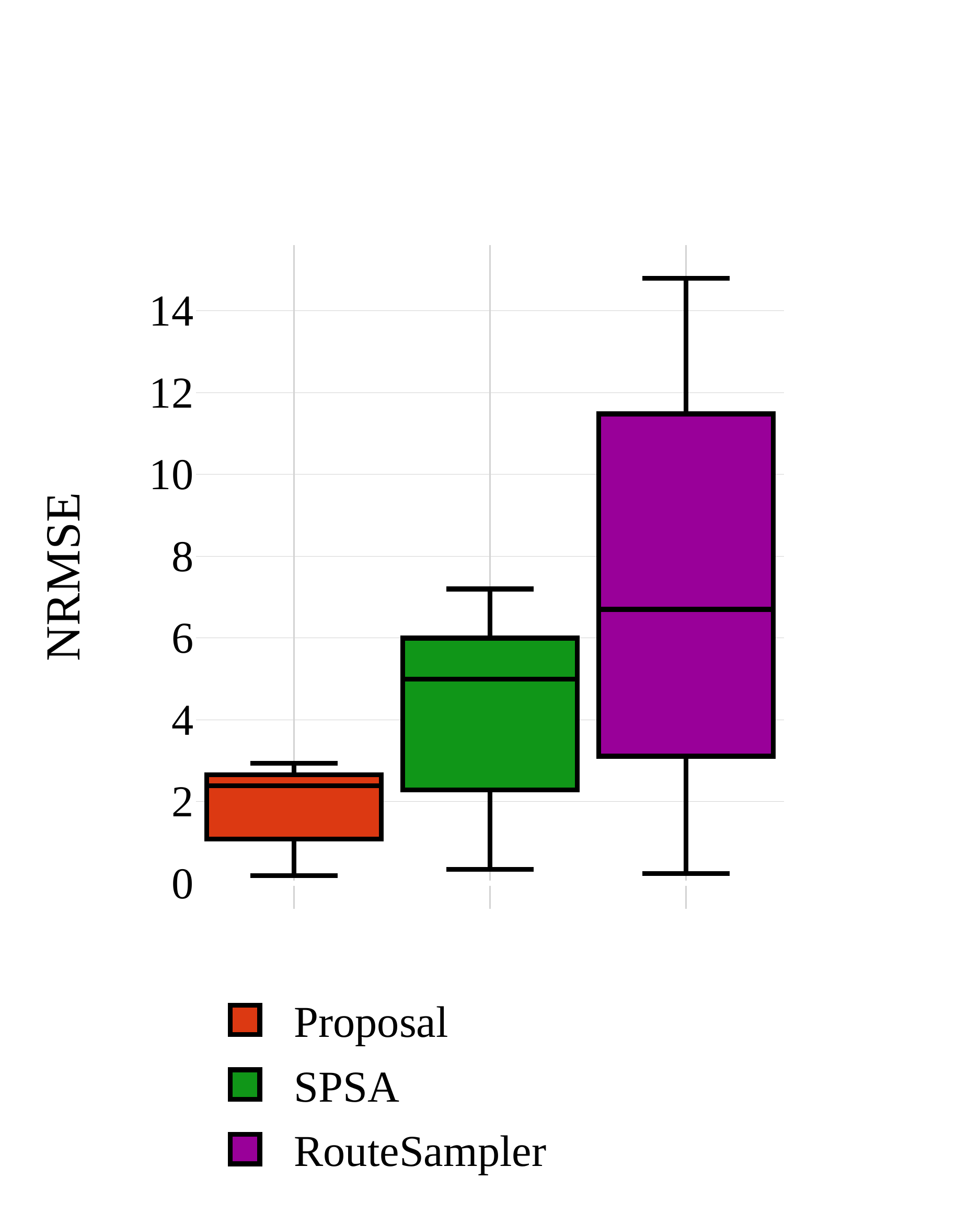}
        \caption{}
        \label{fig:rmse_time_test}
    \end{subfigure}
    \caption{Average root mean squared traffic count error (RMSE) calculated in all the regions and every time interval. Figure~\ref{fig:rmse_sensor_train} (training set) and~\ref{fig:rmse_sensor_test} (test set) are calculated by considering each sensor as a variable, Figure~\ref{fig:rmse_time_train} (training set) and~\ref{fig:rmse_time_test} (test set) considering each time interval as a variable.}
    \label{fig:rmse}
\end{figure}

\begin{figure}[!ht]
    \centering
    \begin{subfigure}[]{.45\textwidth}
        \centering
        \includegraphics[width=\textwidth]{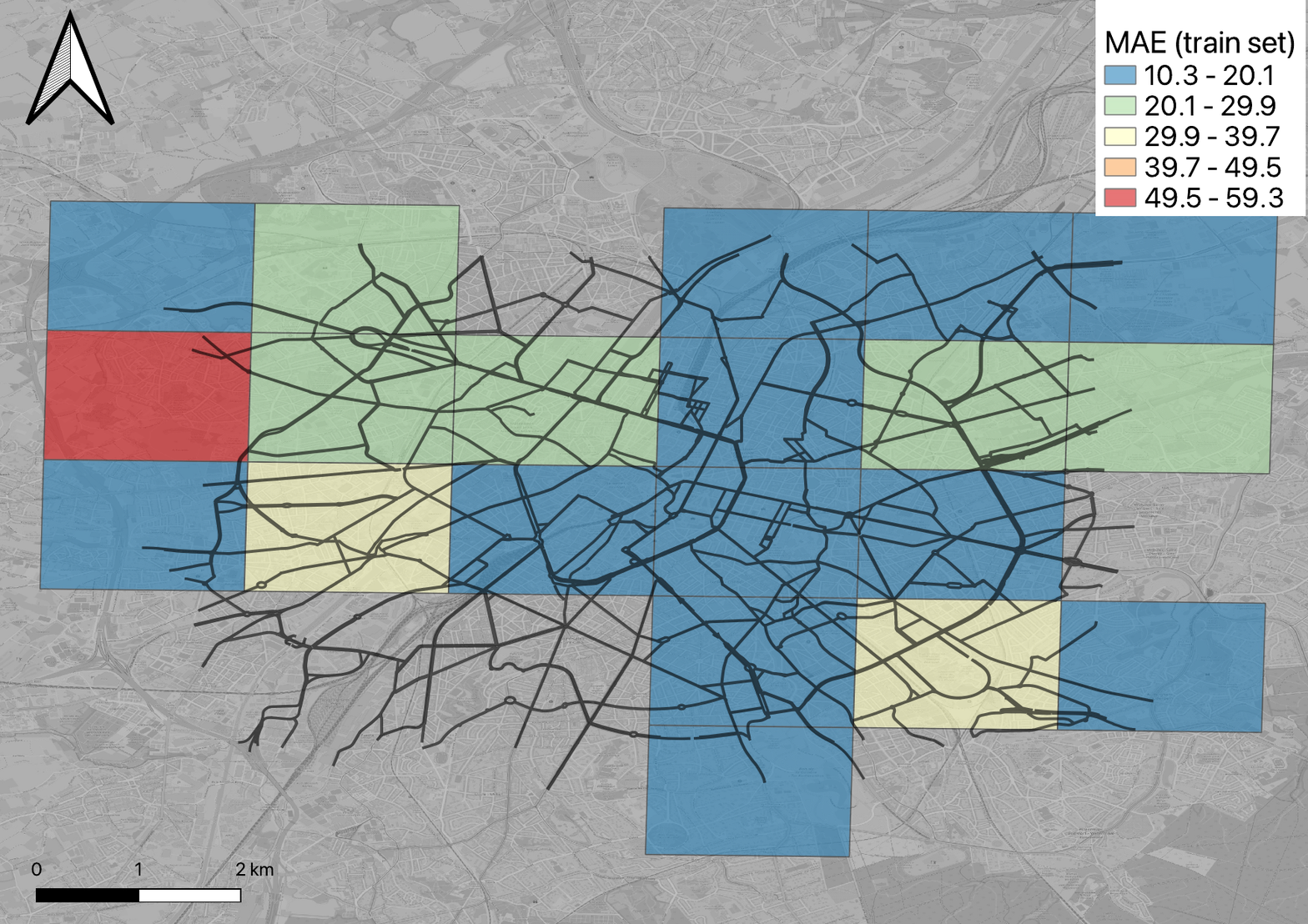}
        \caption{}
        \label{fig:mae_region_train_2000}
    \end{subfigure}
    \hfill
    \begin{subfigure}[]{.45 \textwidth}  
        \centering 
        \includegraphics[width=\textwidth]{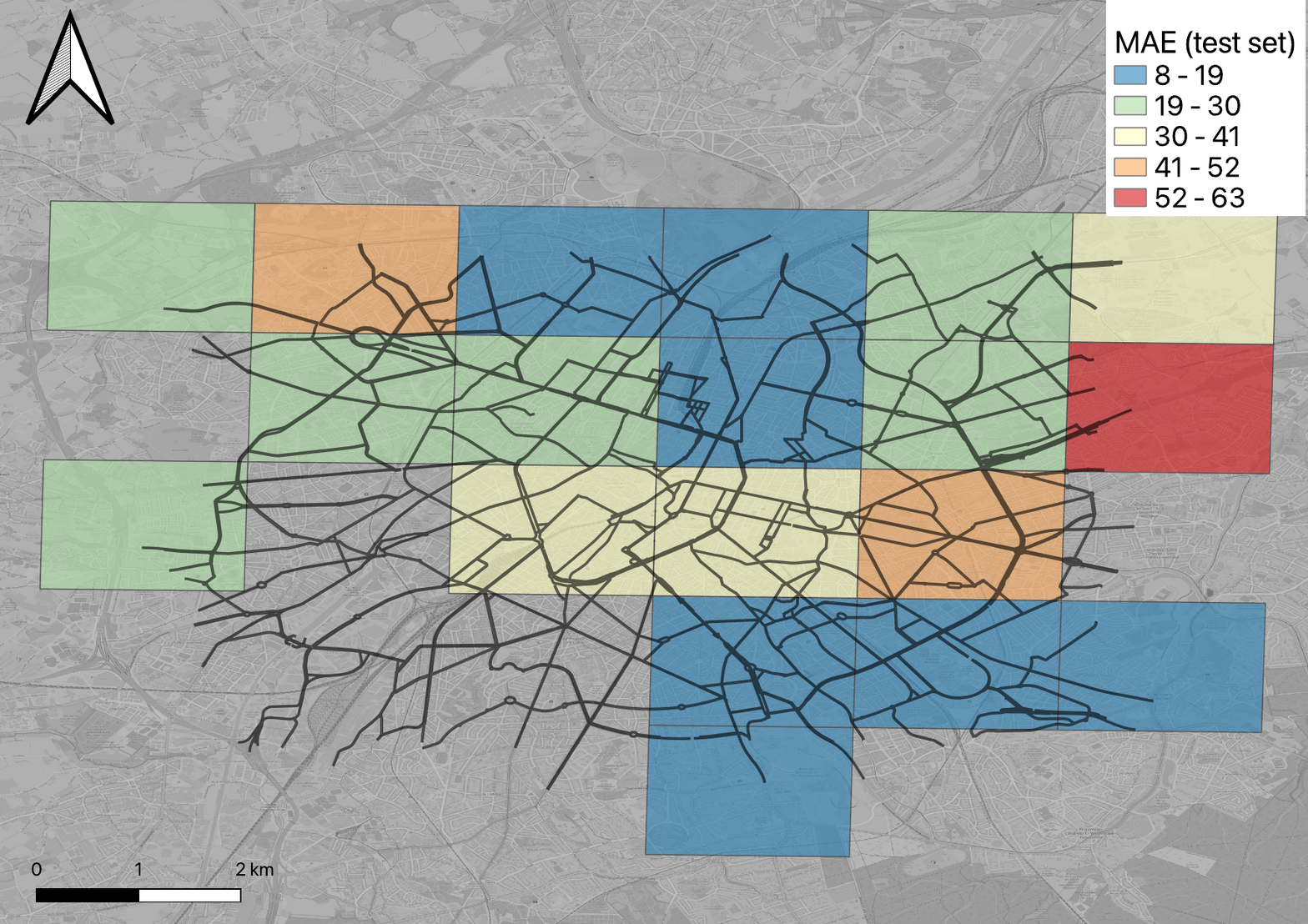}
        \caption{}
        \label{fig:mae_region_test_2000}
    \end{subfigure}
    \vskip\baselineskip
    \begin{subfigure}[]{.45\textwidth}
        \centering
        \includegraphics[width=\textwidth]{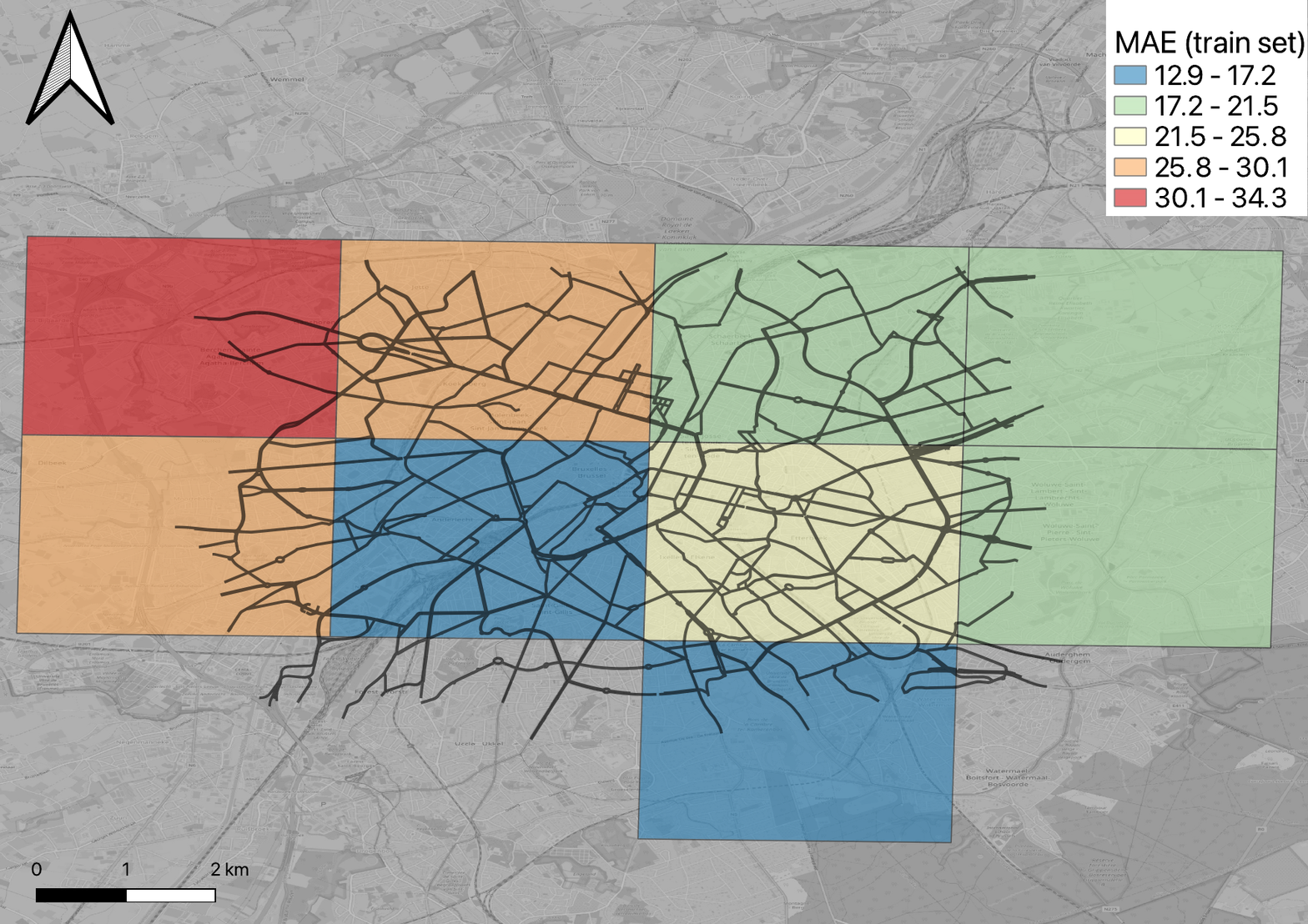}
        \caption{}
        \label{fig:mae_region_train_3500}
    \end{subfigure}
    \hfill
    \begin{subfigure}[]{.45 \textwidth}  
        \centering 
        \includegraphics[width=\textwidth]{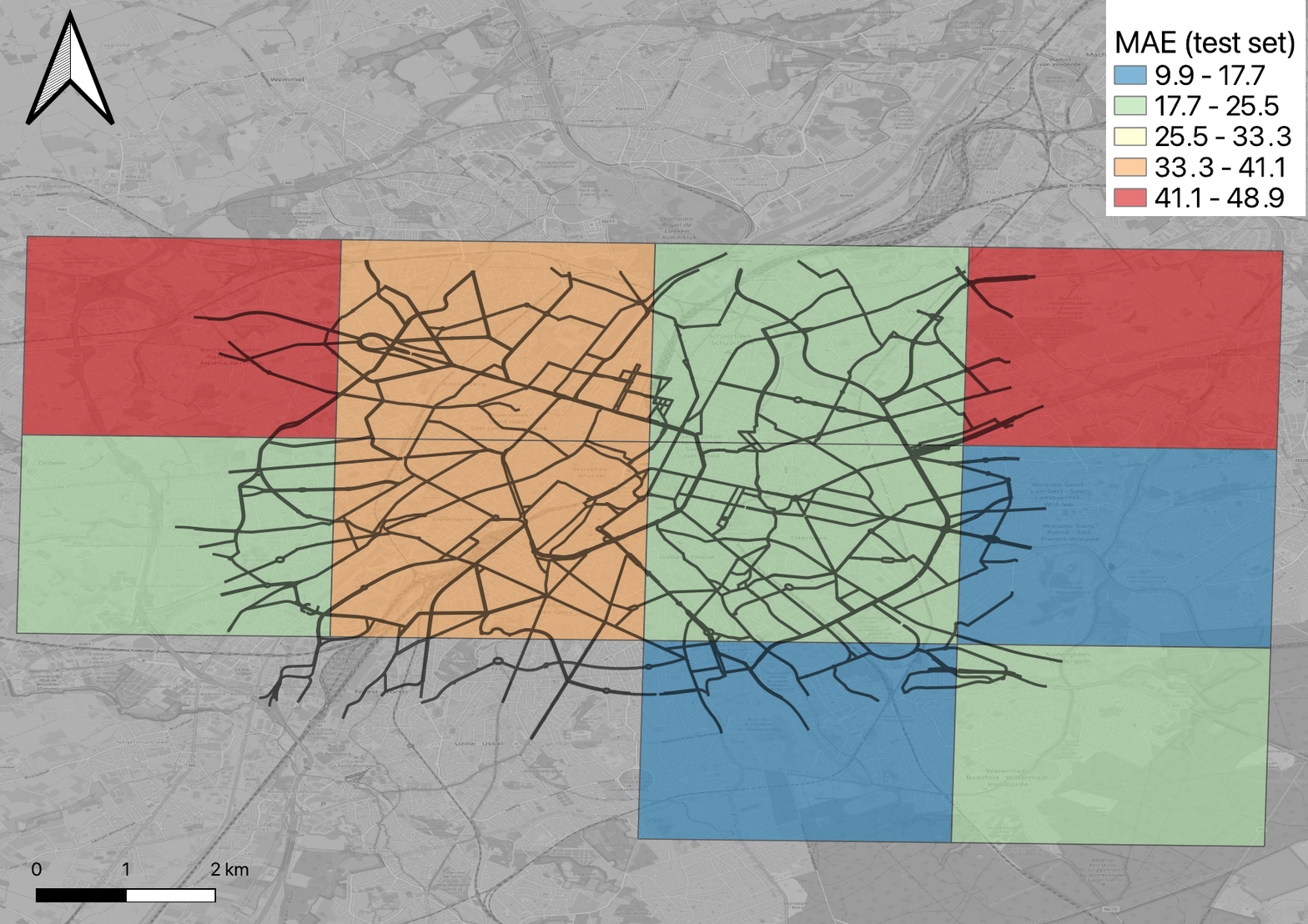}
        \caption{}
        \label{fig:mae_region_test_3500}
    \end{subfigure}
    \vskip\baselineskip
    \begin{subfigure}[]{.45\textwidth}
        \centering
        \includegraphics[width=\textwidth]{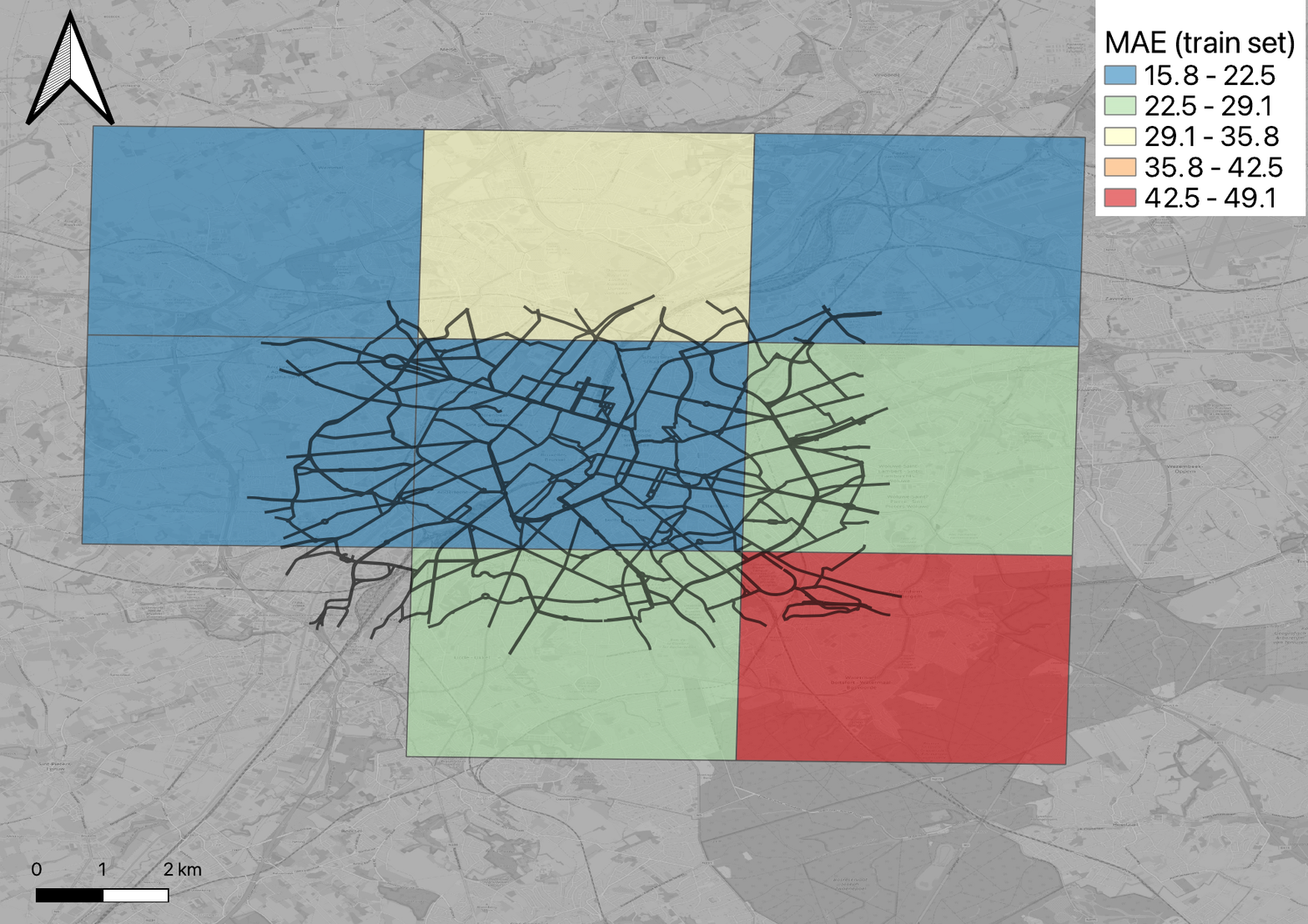}
        \caption{}
        \label{fig:mae_region_train_5000}
    \end{subfigure}
    \hfill
    \begin{subfigure}[]{.45 \textwidth}  
        \centering 
        \includegraphics[width=\textwidth]{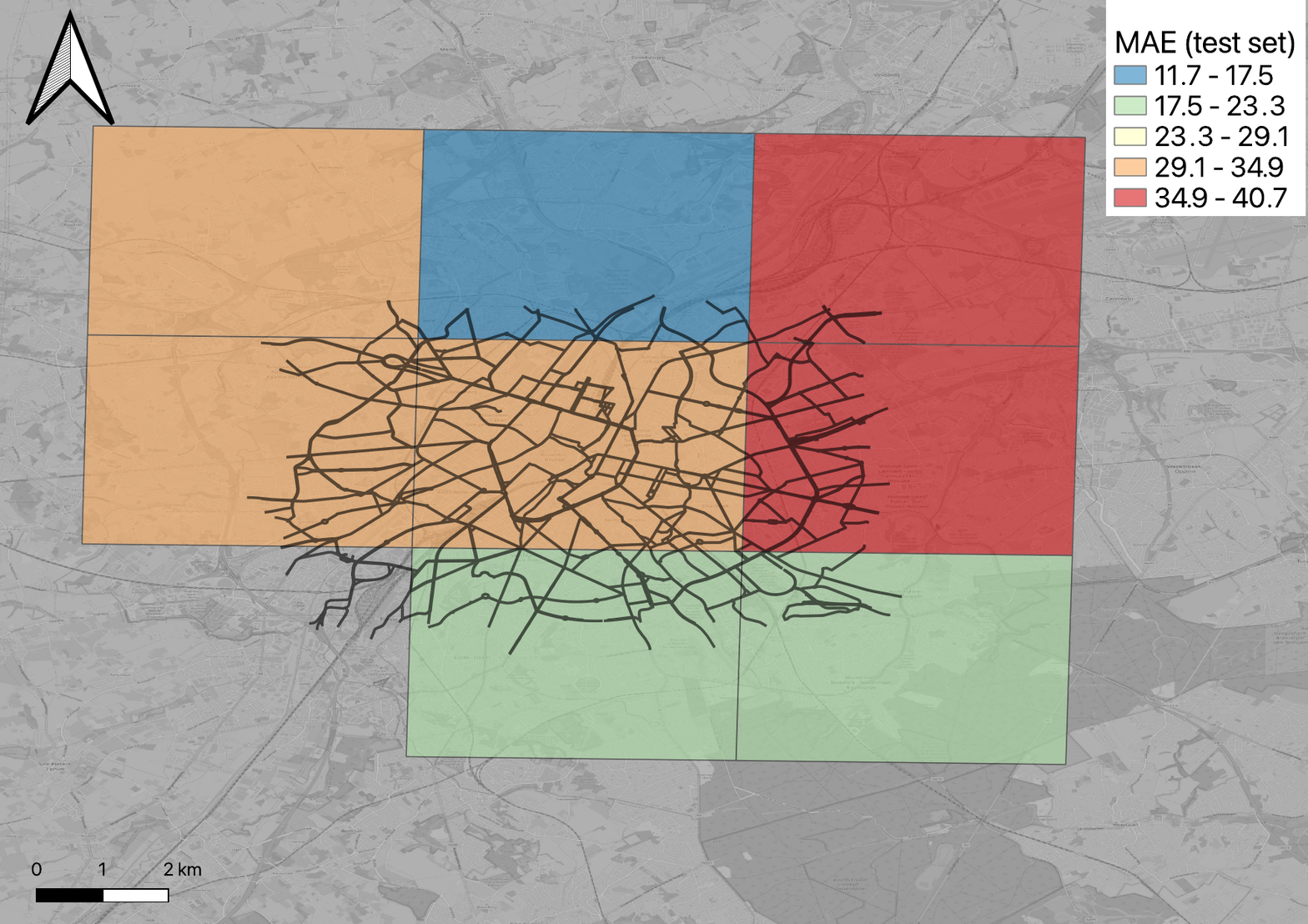}
        \caption{}
        \label{fig:mae_region_test_5000}
    \end{subfigure}
    
    \caption{Spatial distribution of the traffic count error (MAE), calculated as the average absolute difference between real and simulated traffic counts over the modeled time intervals. Figure~\ref{fig:mae_region_train_2000},~\ref{fig:mae_region_train_3500} and~\ref{fig:mae_region_train_5000} show the average absolute error obtained using the training set and regions of size 2000m$^2$, 3500m$^2$, and 5000m$^2$ respectively. Figure~\ref{fig:mae_region_test_2000},~\ref{fig:mae_region_test_3500} and~\ref{fig:mae_region_test_5000} show the average absolute error obtained using the test set and regions of size 2000m$^2$, 3500m$^2$, and 5000m$^2$. The regions do not cover the entire road network because some sensors are situated in edges that are assigned to nearby regions, therefore reducing the number of regions. Best seen in color.}
    \label{fig:mae_by_regions}
\end{figure}

Figure~\ref{fig:mae_by_regions} shows the spatial distribution of the MAE using regions of size 2000m$^2$, 3500m$^2$ and 5000m$^2$. The error reported for each region is calculated as the average of the MAE calculated in all the time intervals. The set of regions is not the same in both figures. This is because in some regions the available traffic monitoring devices are associated only with the training set or test set. The first outcome is the ability to identify the regions where the error is most concentrated, and these regions remain consistent across different region sizes. This enables a more precise localization of the error and a better assessment of the quality of the data used for calibration. On average, the majority of the region has an MAE of less than 20 in the training and testing set. The accuracy of the estimated traffic is influenced by the size of the regions. When using larger regions, the calibration process considers the average of the sensors within each region to adjust the amount of traffic passing through, resulting in coarser outcomes compared to using smaller regions. The average difference between the total number of simulated and actual vehicles, across all time intervals, is approximately 277 when using regions of 2000m$^2$, 238 with regions of 3500m$^2$, and 211 with regions of 5000m$^2$. However, when considering the average objective function value (calculated as the average objective function obtained from all the regions), we obtain 21 for regions of 2000m$^2$, 22 for regions of 3500m$^2$, and 25 for regions of 5000m$^2$. This demonstrates that, although the overall results may appear better with larger regions, the objective function value is lower with smaller regions, ultimately leading to more accurate results.

Further investigation could improve the local accuracy of the traffic model, such as evaluating the correctness of the road network, analyzing the inclusion of traffic lights, or verifying the presence of noise in the input traffic data. Different regions' sizes could be used to delimit the local part of the environment where the error is concentrated and identify errors in the data or the road network.

\begin{figure}[!ht]
    \centering
    \begin{subfigure}[b]{.48\textwidth}
        \centering
        \includegraphics[width=\textwidth]{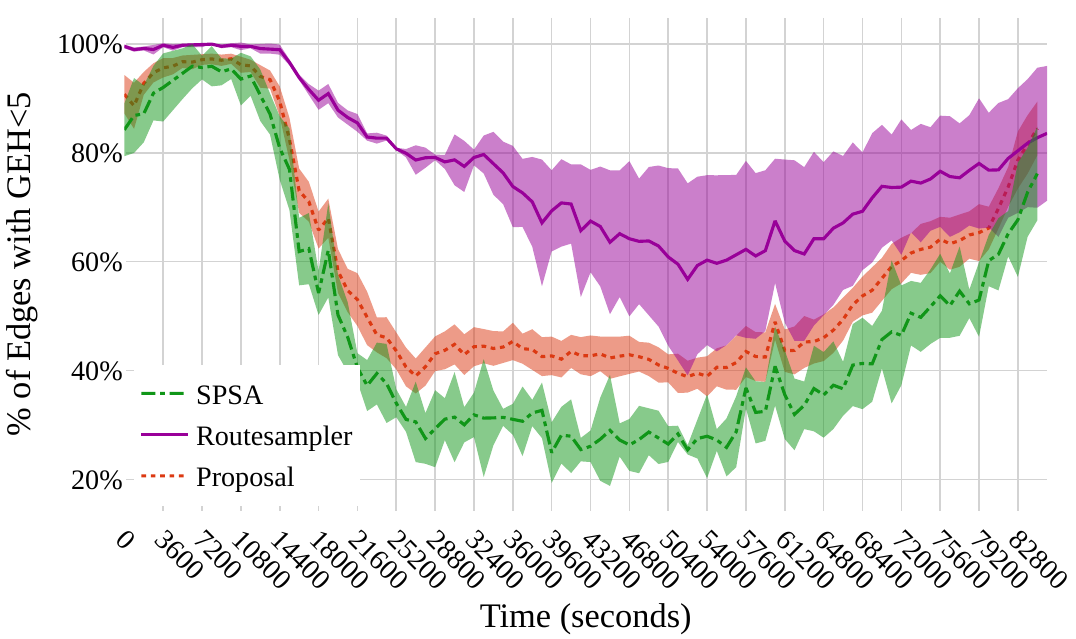}
        \caption{}
        \label{fig:sqv_15min}
    \end{subfigure}
    \begin{subfigure}[b]{.48\textwidth}
        \centering
        \includegraphics[width=\textwidth]{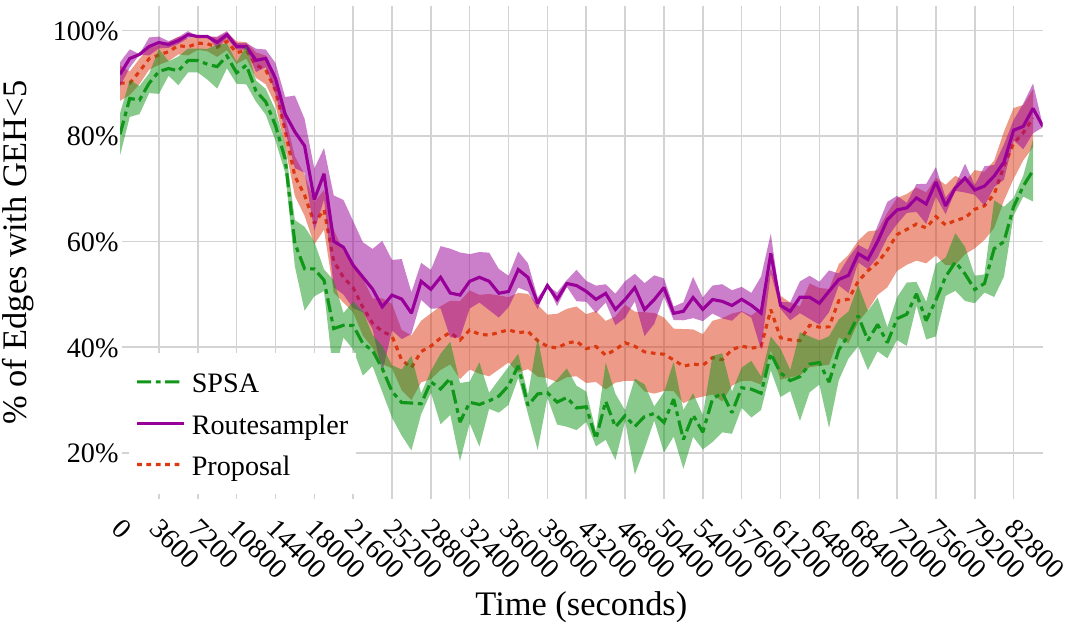}
        \caption{}
        \label{fig:sqv_1h}
    \end{subfigure}
    \caption{Percentage of edges whose GEH value is less than 5. The values are obtained using the training sensors (Figure~\ref{fig:sqv_15min}) and the test sensors (Figure~\ref{fig:sqv_1h}).}
    \label{fig:sqv}
\end{figure}

The GEH indicator is a standard measure to evaluate the goodness-of-fit of a traffic model compared to reality. Figure~\ref{fig:sqv} reports for each time interval, the percentage of edges whose value of GEH is less than 5. Generally, a model is considered as realistic if at least 85\% of the edges have a GEH$<$5~\cite{wei_evaluating_2023}. Despite the results indicating that less than 85\% of edges have a GEH$<$5, those obtained by the proposed method are as good as those obtained by RouteSampler in the test set, with the exception that this method shows a significant variability over the conducted experiments. This suggests that the RouteSampler method is unable to maintain a stable GEH value in the considered experimental setting.

\section*{Discussion}\label{sec:limitation}

An important aspect of traffic calibration is having access to data from the real world. Different studies~\cite{chao2020survey,rong2023interdisciplinary} agree that data quality and availability are fundamental requirements in calibrating traffic simulation models. The use of CCTV-based monitoring infrastructures can be an obstacle: cameras cannot be installed in all areas of an urban environment, and data is not always available to the community~\cite{10360829}. The limitations of the proposed method relate mainly to the available data. The traffic data may be incomplete or subject to errors due to inaccuracies in traffic monitoring devices, which can affect the model's accuracy. Furthermore, there may be gaps in data coverage in areas where these devices are absent.

Traffic calibration is an under-determined problem: for a given set of traffic counts, an infinite number of traffic models can be defined to fit the input data. This leads to a high degree of uncertainty in the generated models, making the decision-making process non-trivial. In the proposed method, we cannot guarantee that a trajectory derived from traffic counts corresponds to actual real-world conditions.

\section*{Conclusion and Future Works}\label{sec:conclusion}

This work presents a simulation-based method optimization technique for estimating traffic models from real measurements. Our proposal iteratively calibrates traffic models by removing or adding vehicles so that the difference between the traffic observed in the simulation and the reality is minimized. We use the open-source simulator SUMO for simulating traffic. Compared to the state-of-the-art methods, our proposal does not require pre-calibrated OD matrices, allows calibrating traffic locally in space and time, and ensures temporal coherence of traffic conditions across consecutive time intervals. By performing local optimization, it is possible to identify parts of the environment where the calibration technique outputs less accurate traffic. This information can be used to conduct further analysis in the input data or vary the parameters of the calibration technique to produce an accurate model without interfering with the other regions.

We compare the results obtained by our proposal to those obtained by standard calibration methods: one available with SUMO, and one based on the SPSA optimization technique. The results prove that the proposed technique outperforms the two baseline methods.  

In future works, we will investigate methods to generate simulated traffic based on the attractiveness of places over time and define new mechanisms for calibrating multi-modal traffic, thus taking into account different types of mobility. Also, more data will be provided and released to the community. This will be beneficial to build a collection of scenarios and assess the accuracy of the calibrated model under different traffic days (for instance, comparing all the weekends, of week days).

\bigskip

\textbf{Acknowledgement}\quad This research work is being developed in the context of TORRES (Traffic prOcessing foR uRban EnvironmentS), a Joint R\&D Project (2022-RDIR-59b) funded by ``R\'{e}gion de Bruxelles-Capitale - Innoviris''. G. Bontempi is also supported by the Service Public de Wallonie Recherche under grant nr 2010235–ARIAC by DigitalWallonia4.ai. 

Many thanks to Mohamed Aarab and Nguyen Huu Chuong from Brussels Mobility for generously providing the traffic data used for the experimentations.

The work of D. A. Guastella was carried out during his stay at ULB.

\bigskip

\textbf{Author Contribution}\quad \textbf{Davide A. Guastella}: conception, implementation, experimental design, paper writing; \textbf{Alejandro Morales-Hernández}: problem statement formulation, paper writing; \textbf{Gianluca Bontempi}: conception, problem statement formulation, proofreading; \textbf{Bruno Cornelis}: problem statement formulation, proofreading.

\bigskip

\textbf{Competing interest}\quad The author(s) declare no competing interests.

\bigskip

\textbf{Availability of materials and data}\quad The online version contains supplementary material available.

\bibliographystyle{unsrt} 
\bibliography{biblio}

\onecolumn
\newpage
\clearpage
\section*{Appendices}


\begin{appendices}

\section{Computational Time}
\label{appendix:computational_time}


\begin{figure}[!ht]
    \centering
    \begin{subfigure}[b]{.52\textwidth}
        \centering
        \includegraphics[width=\textwidth]{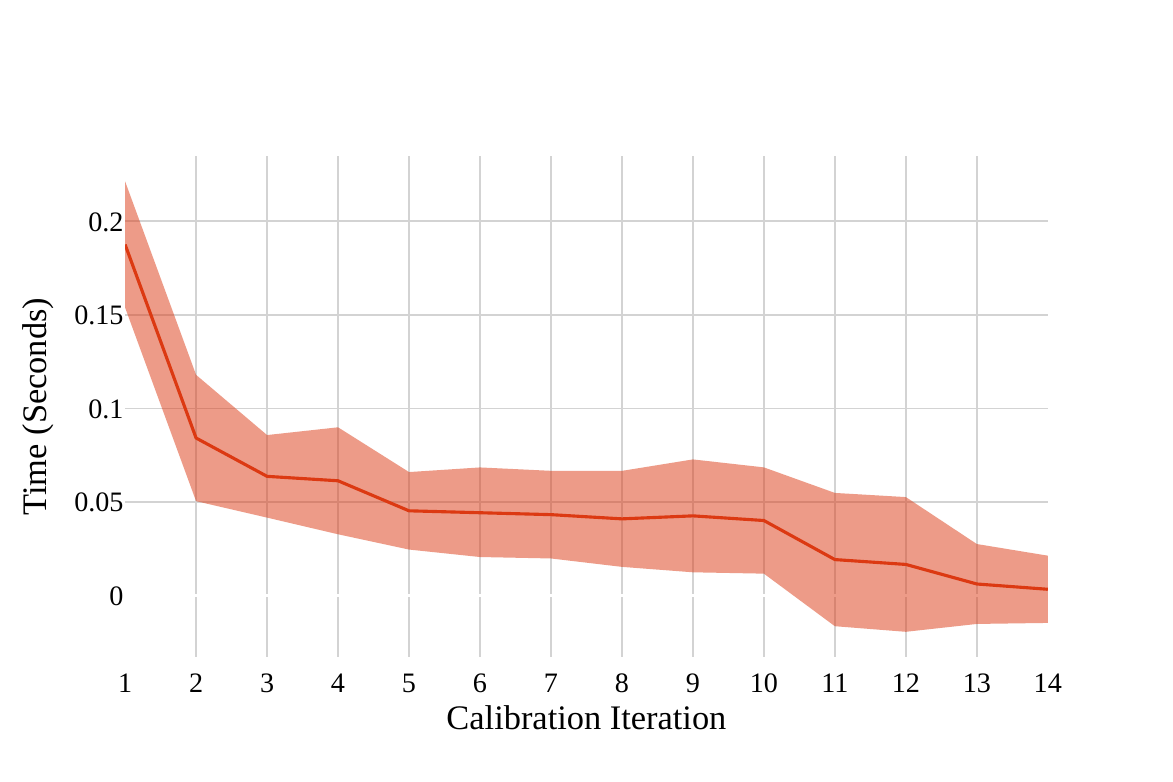}
        \caption{}
        \label{fig:calib_time_1201}
    \end{subfigure}
    
    \begin{subfigure}[b]{.52\textwidth}
        \centering
        \includegraphics[width=\textwidth]{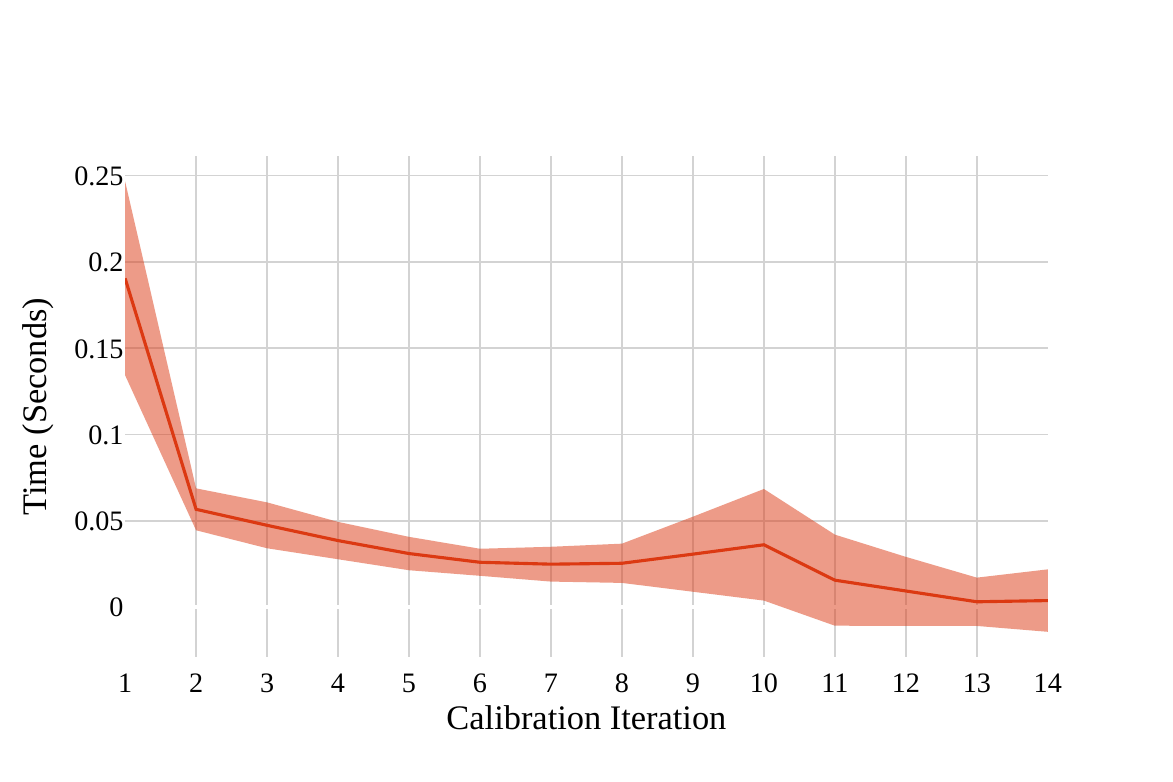}
        \caption{}
        \label{fig:calib_time_1202}
    \end{subfigure}
    
    \begin{subfigure}[b]{.52\textwidth}
        \centering
        \includegraphics[width=\textwidth]{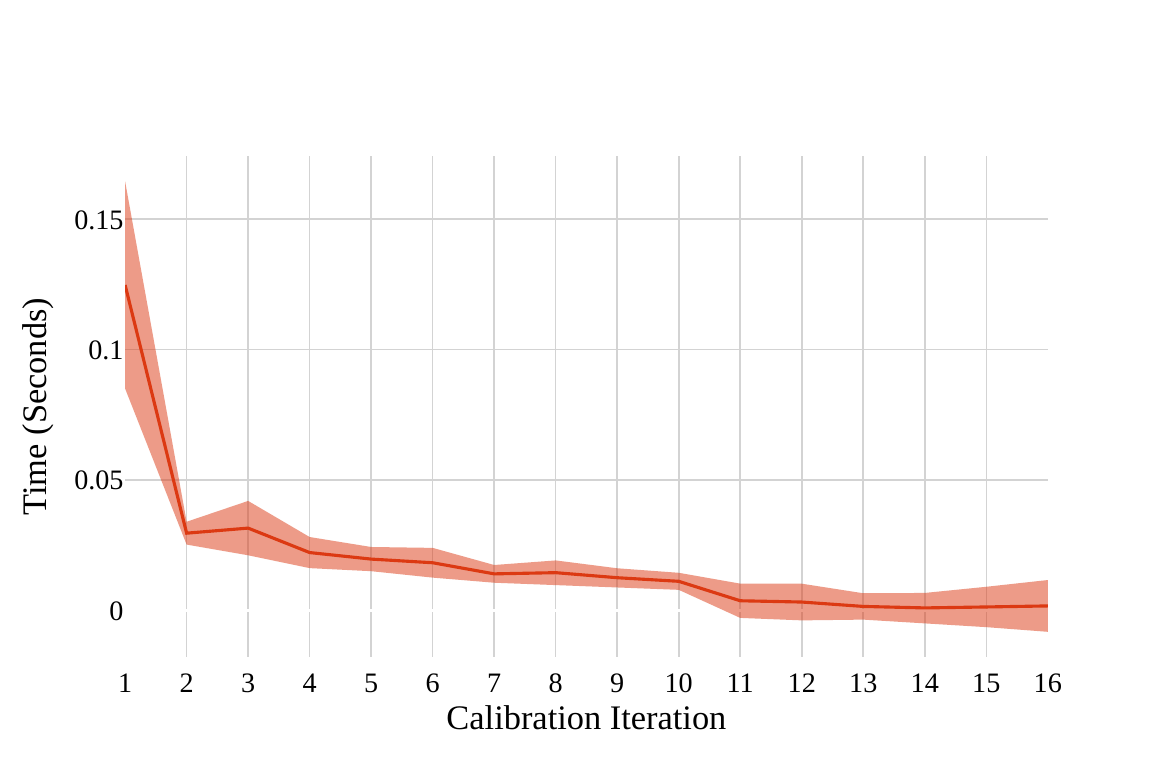}
        \caption{}
        \label{fig:calib_time_1203}
    \end{subfigure}
    \caption{Computational time (seconds) required to calibrate traffic using the proposed method. The results relate to December 1, 2023 (Figure~\ref{fig:calib_time_1201}), December 2, 2023 (Figure~\ref{fig:calib_time_1202}), and December 3, 2023 (Figure~\ref{fig:calib_time_1203}).}
    \label{fig:calib_time}
\end{figure}

\newpage
\section{Experimental Results - December 1, 2023}
\label{appendix:results_dec1}


\begin{figure}[!h]
    \centering
    \begin{subfigure}[b]{.48\textwidth}
        \centering
        \includegraphics[width=\textwidth]{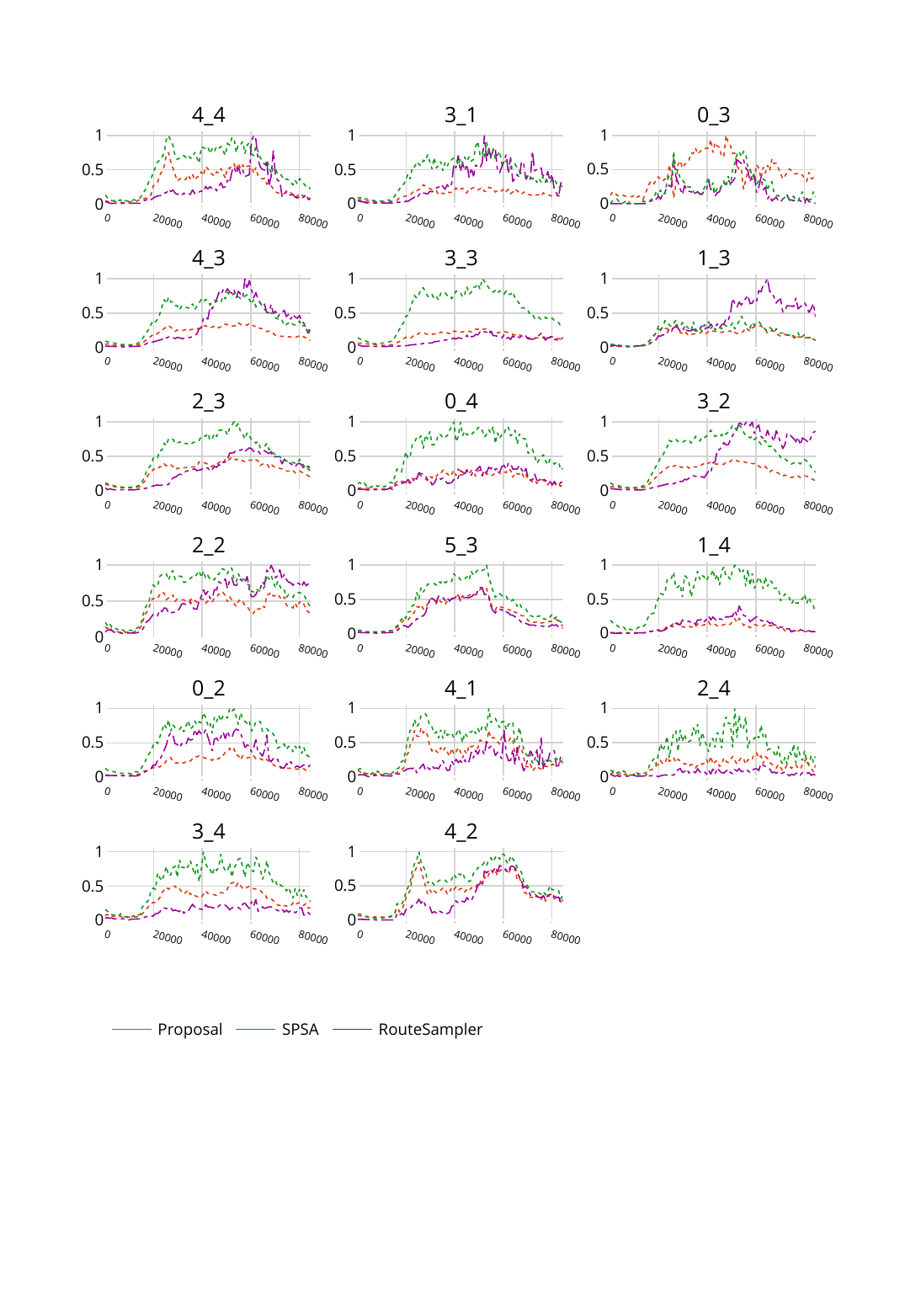}
        \caption{December 1, 2023 (training set)}
        \label{fig:calib_error_1201_train}
    \end{subfigure}
    \hfill
    \begin{subfigure}[b]{.48\textwidth}
        \centering
        \includegraphics[width=\textwidth]{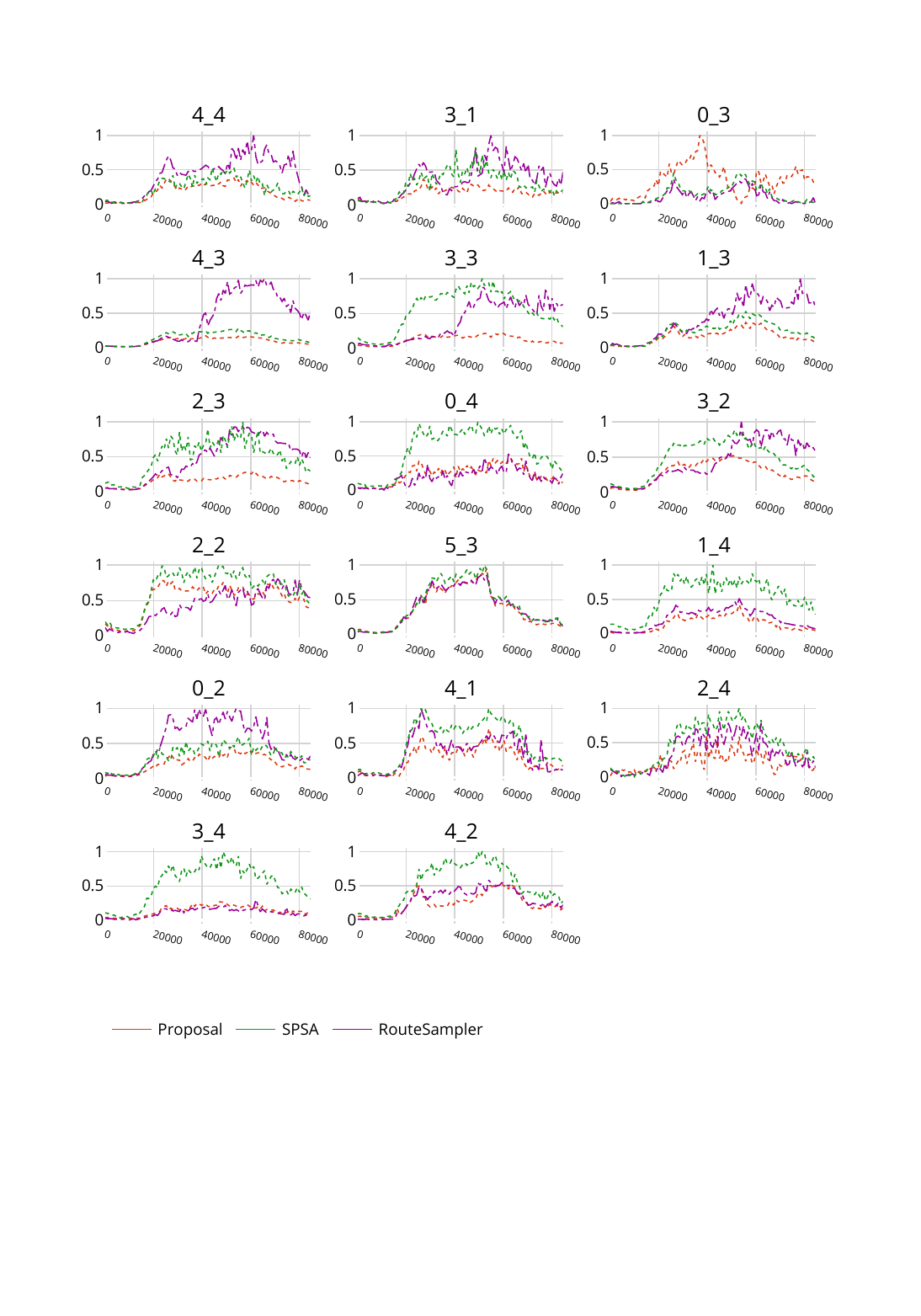}
        \caption{December 1, 2023 (test set)}
        \label{fig:calib_error_1201_test}
    \end{subfigure}
    \caption{Local average absolute traffic counts difference (MAE), calculated on the training set (Figure~\ref{fig:calib_error_1201_train}) and test set (Figure~\ref{fig:calib_error_1201_test}), using the data from December 1, 2023.}
    \label{fig:calib_error_1201}
\end{figure}


\begin{figure}[!h]
    \centering
    \begin{subfigure}[b]{.49\textwidth}
        \centering
        \includegraphics[width=\textwidth]{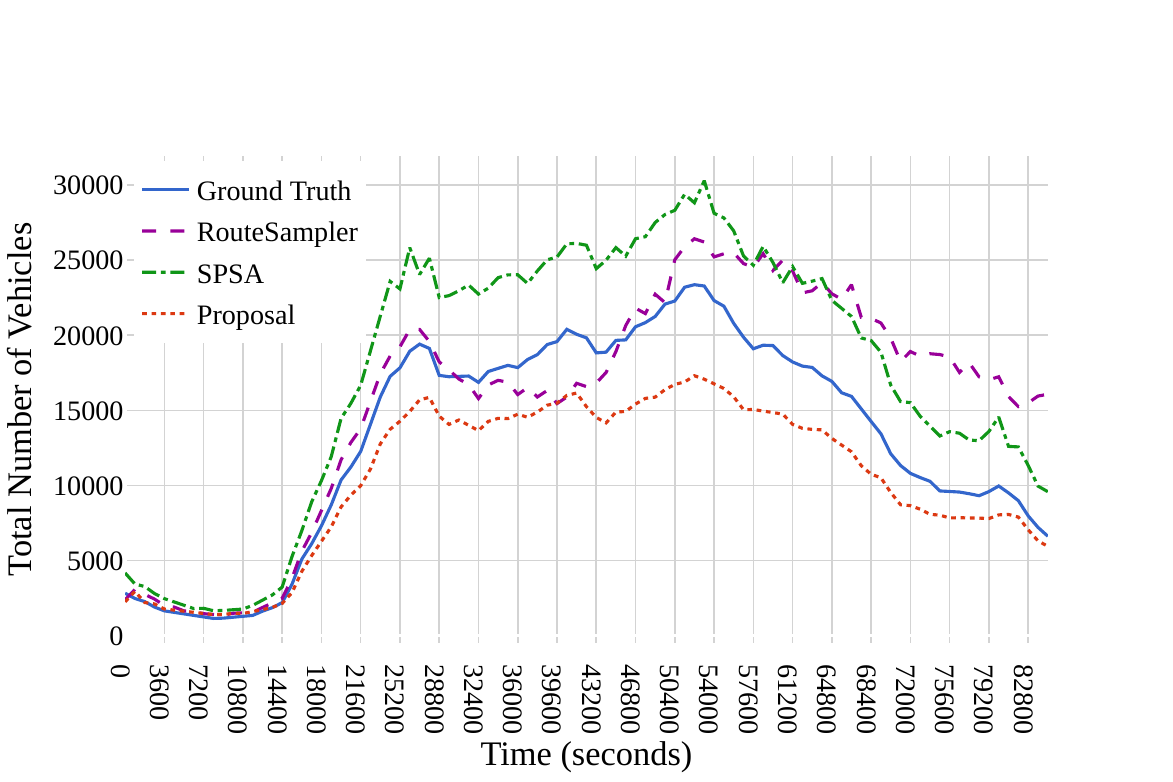}
        \caption{December 1, 2023 (training set)}
        \label{fig:traffic_vol_1201_train}
    \end{subfigure}
    \begin{subfigure}[b]{.49\textwidth}
        \centering
        \includegraphics[width=\textwidth]{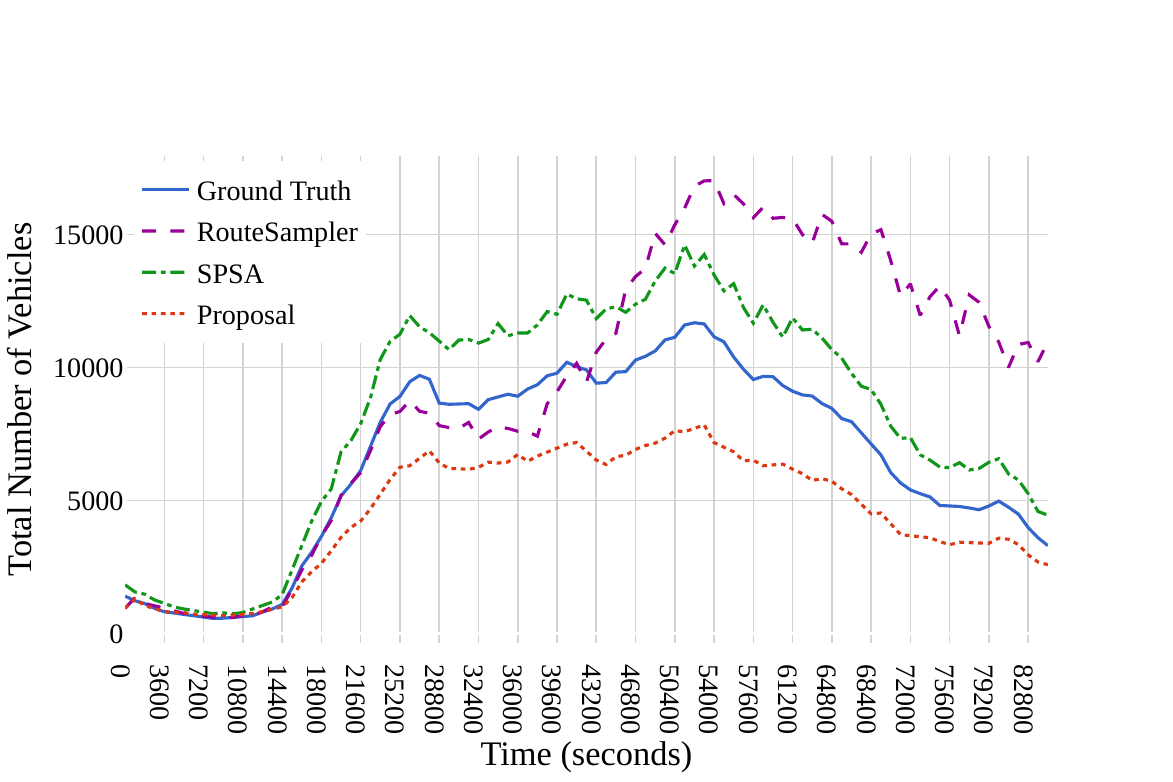}
        \caption{December 1, 2023 (test set)}
        \label{fig:traffic_vol_1201_test}
    \end{subfigure}
    \caption{Total number of vehicles observed in all the regions and every time interval, in the ground truth and the simulation. Figure~\ref{fig:traffic_vol_1201_train} shows the total number of vehicles obtained by the training sensors, Figure~\ref{fig:traffic_vol_1201_test} by the test sensors.}
    \label{fig:traffic_vol_1201}
\end{figure}


\begin{figure}[!h]
    \centering
    \begin{subfigure}[b]{.49\textwidth}
        \centering
        \includegraphics[width=\textwidth]{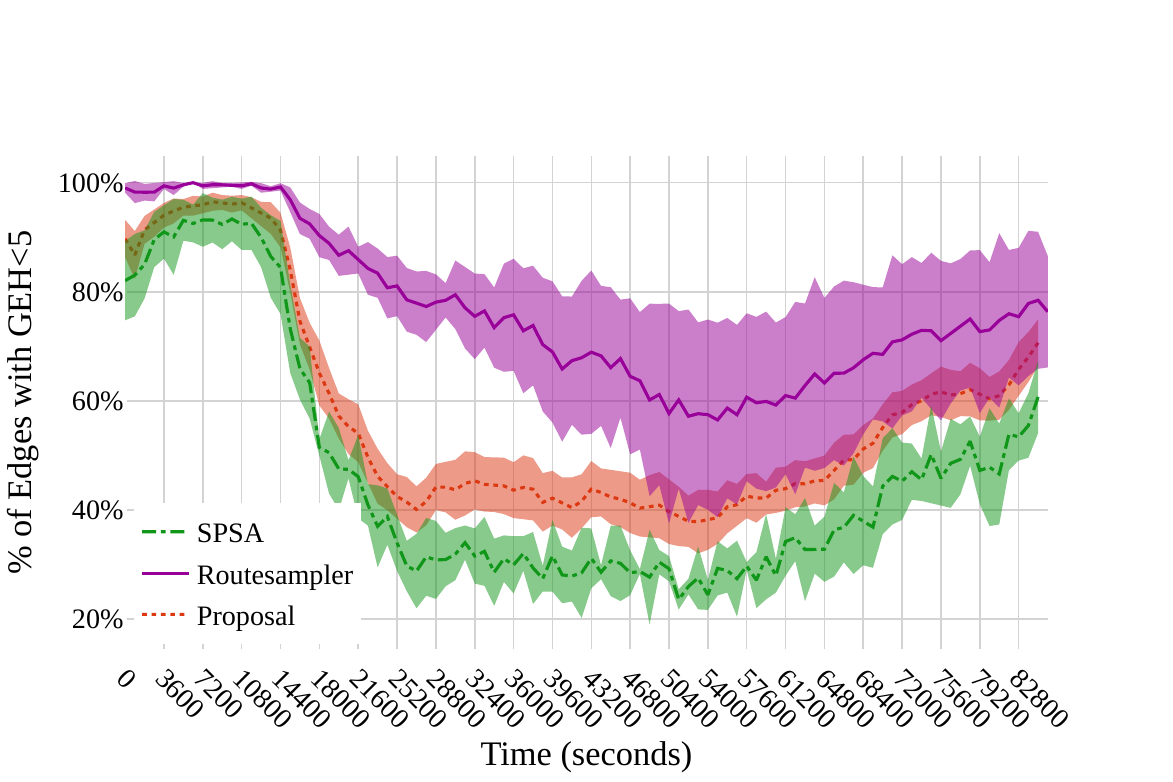}
        \caption{December 1, 2023 (training set)}
        \label{fig:geh_1201_train}
    \end{subfigure}
    \begin{subfigure}[b]{.49\textwidth}
        \centering
        \includegraphics[width=\textwidth]{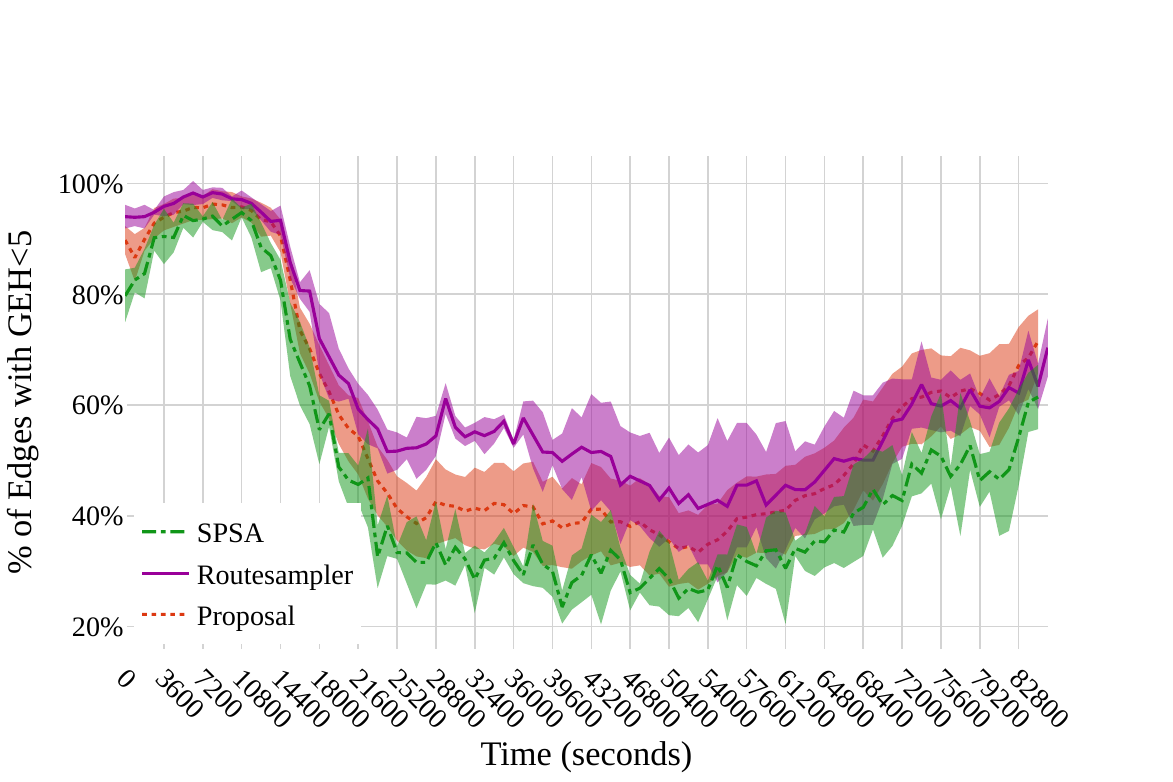}
        \caption{December 1, 2023 (test set)}
        \label{fig:geh_1201_test}
    \end{subfigure}
    \caption{Percentage of edges whose GEH value is less than 5. This is obtained by the proposed and baseline techniques. The values are obtained using the training sensors (Figure~\ref{fig:geh_1201_train}) and the test sensors (Figure~\ref{fig:geh_1201_test}).}
    \label{fig:geh_1201}
\end{figure}


\begin{figure}[!h]
    \centering
    \begin{subfigure}[b]{.49\textwidth}
        \centering
        \includegraphics[width=\textwidth]{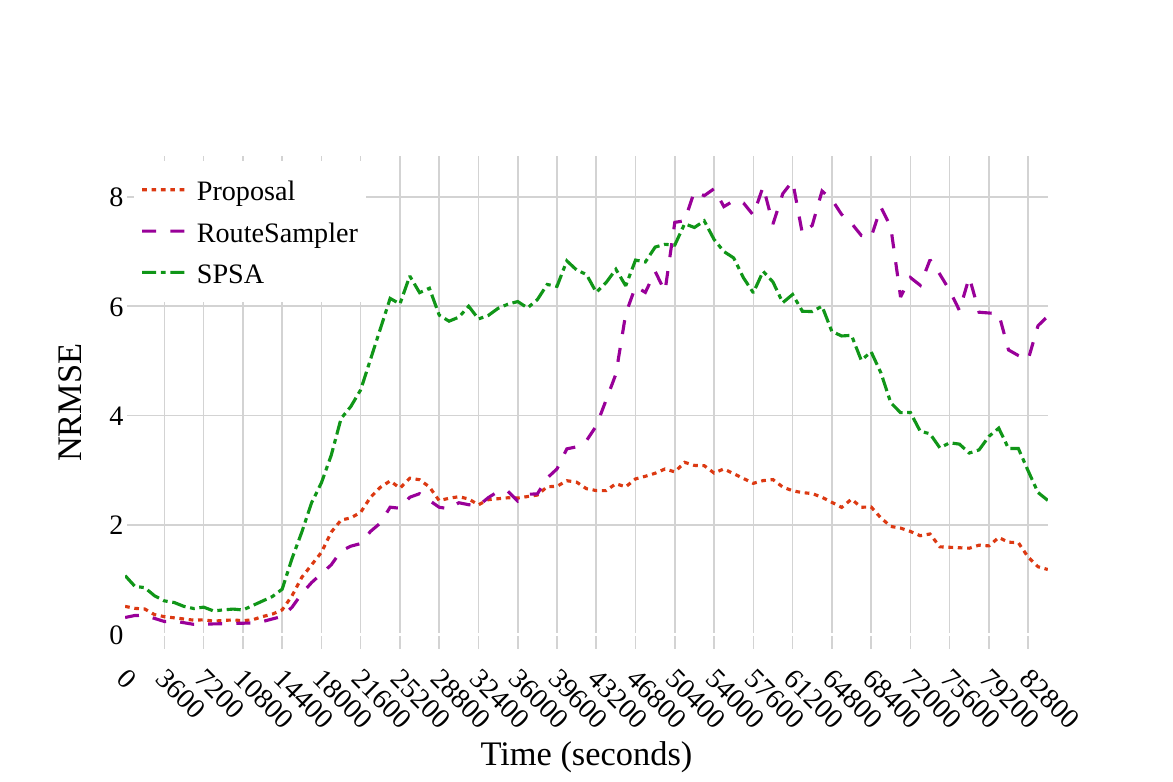}
        \caption{December 1, 2023 (training set)}
        \label{fig:nrmse_1201_train}
    \end{subfigure}
    \begin{subfigure}[b]{.49\textwidth}
        \centering
        \includegraphics[width=\textwidth]{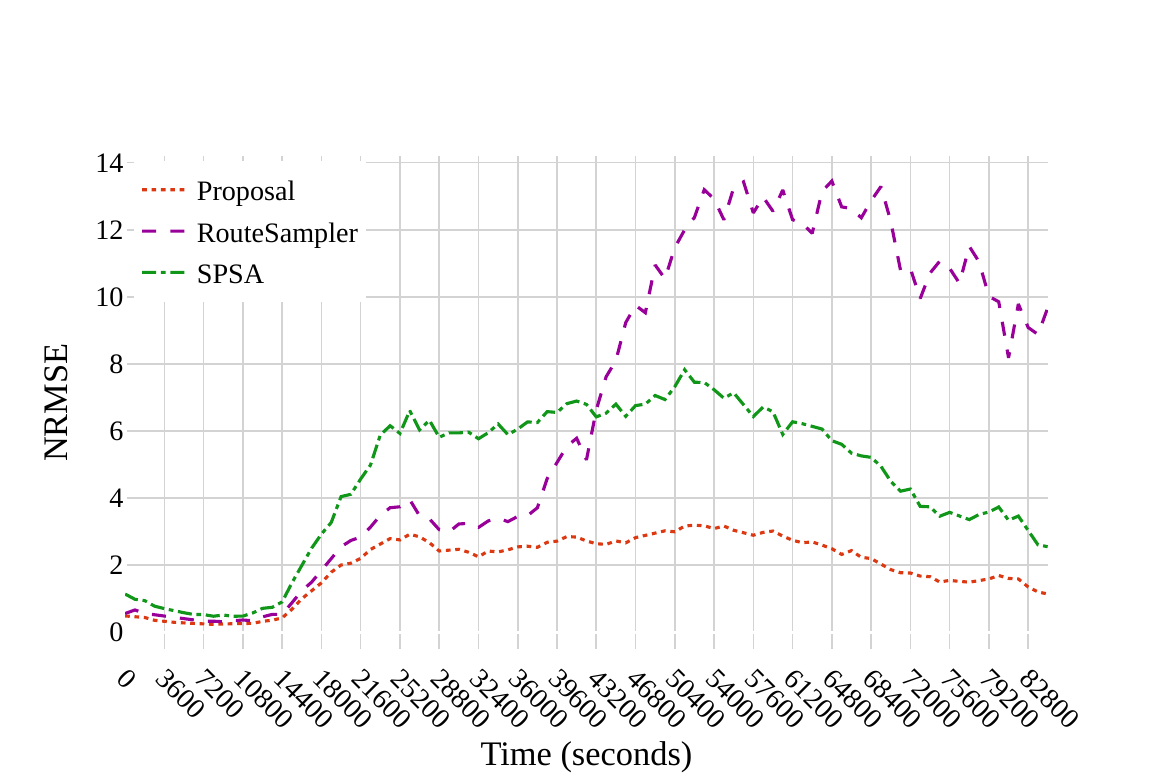}
        \caption{December 1, 2023 (test set)}
        \label{fig:nrmse_1201_test}
    \end{subfigure}
    \caption{Normalized RMSE, calculated using the training set (Figure~\ref{fig:nrmse_1201_train}) and the test set (Figure~\ref{fig:nrmse_1201_test}).}
    \label{fig:nrmse_1201}
\end{figure}


\clearpage
\newpage

\section{Experimental Results - December 2, 2023}
\label{appendix:results_dec2}


\begin{figure}[!h]
    \centering
    \begin{subfigure}[b]{.48\textwidth}
        \centering
        \includegraphics[width=\textwidth]{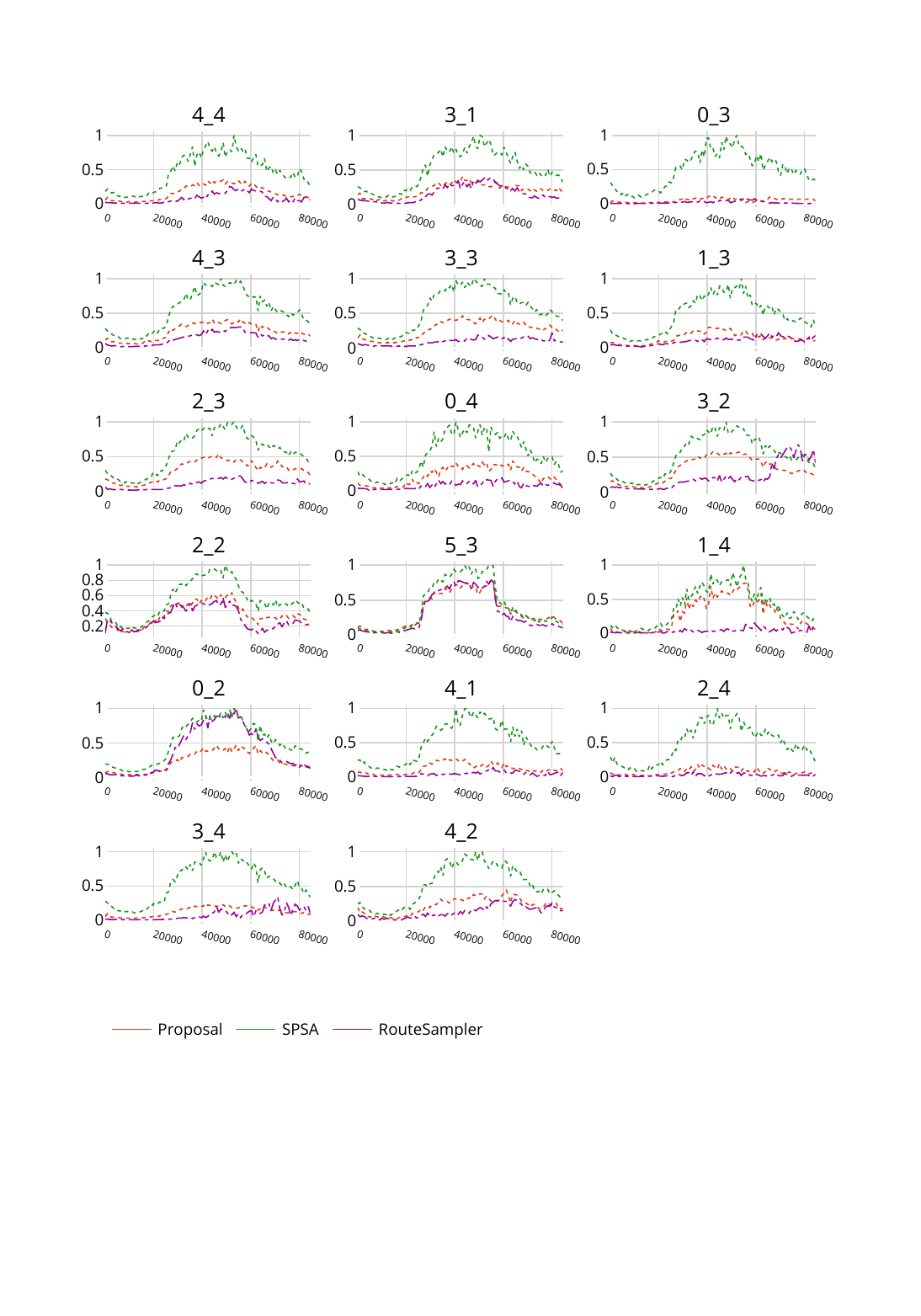}
        \caption{December 2, 2023 (training set)}
        \label{fig:calib_error_1202_train}
    \end{subfigure}
    \begin{subfigure}[b]{.48\textwidth}
        \centering
        \includegraphics[width=\textwidth]{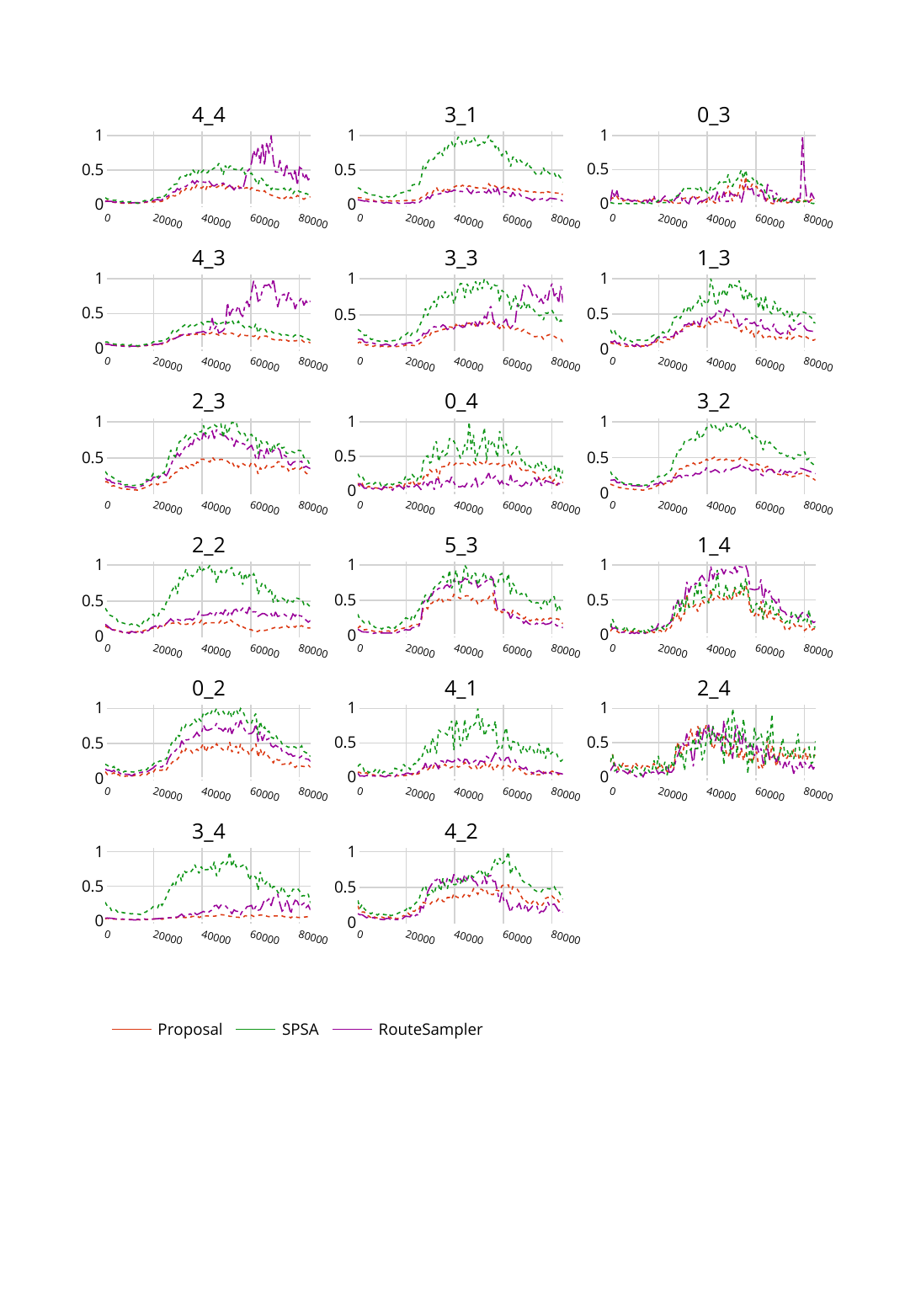}
        \caption{December 2, 2023 (test set)}
        \label{fig:calib_error_1202_test}
    \end{subfigure}
    \caption{Local average absolute traffic counts difference (MAE), calculated on the training set (Figure~\ref{fig:calib_error_1202_train}) and test set (Figure~\ref{fig:calib_error_1202_test}), using the data from December 2, 2023.}
    \label{fig:calib_error_1202}
\end{figure}


\begin{figure}[!h]
    \centering
    \begin{subfigure}[b]{.49\textwidth}
        \centering
        \includegraphics[width=\textwidth]{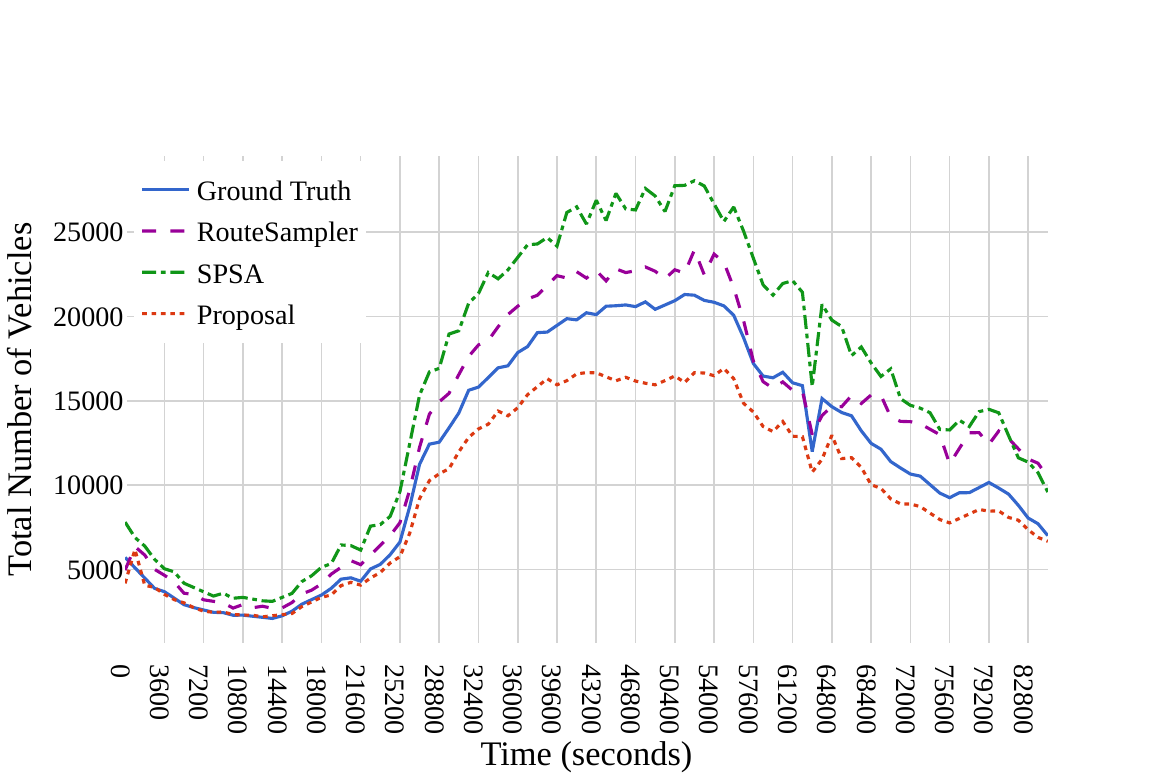}
        \caption{December 2, 2023 (training set)}
        \label{fig:traffic_vol_1202_train}
    \end{subfigure}
    \begin{subfigure}[b]{.49\textwidth}
        \centering
        \includegraphics[width=\textwidth]{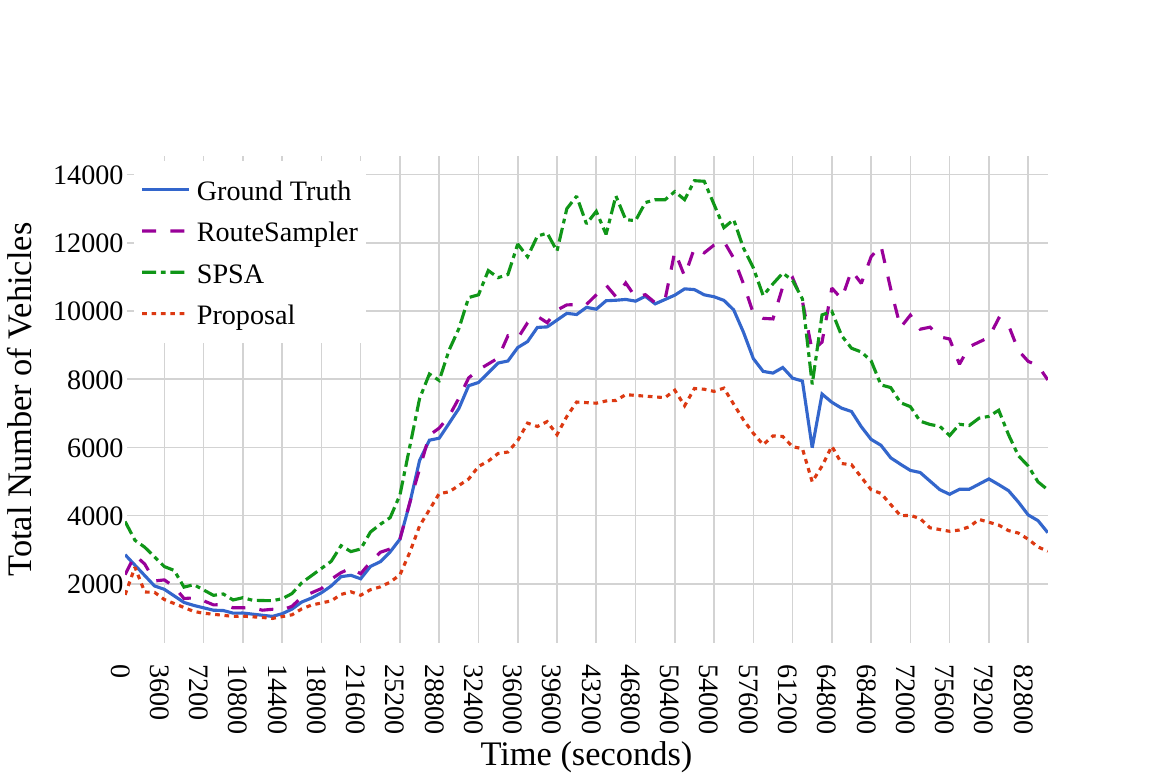}
        \caption{December 2, 2023 (test set)}
        \label{fig:traffic_vol_1202_test}
    \end{subfigure}
    \caption{Total number of vehicles observed in all the regions and every time interval, in the ground truth and the simulation. Figure~\ref{fig:traffic_vol_1202_train} shows the total number of vehicles obtained by the training sensors, Figure~\ref{fig:traffic_vol_1202_test} by the test sensors.}
    \label{fig:traffic_vol_1202}
\end{figure}


\begin{figure}[!h]
    \centering
    \begin{subfigure}[b]{.49\textwidth}
        \centering
        \includegraphics[width=\textwidth]{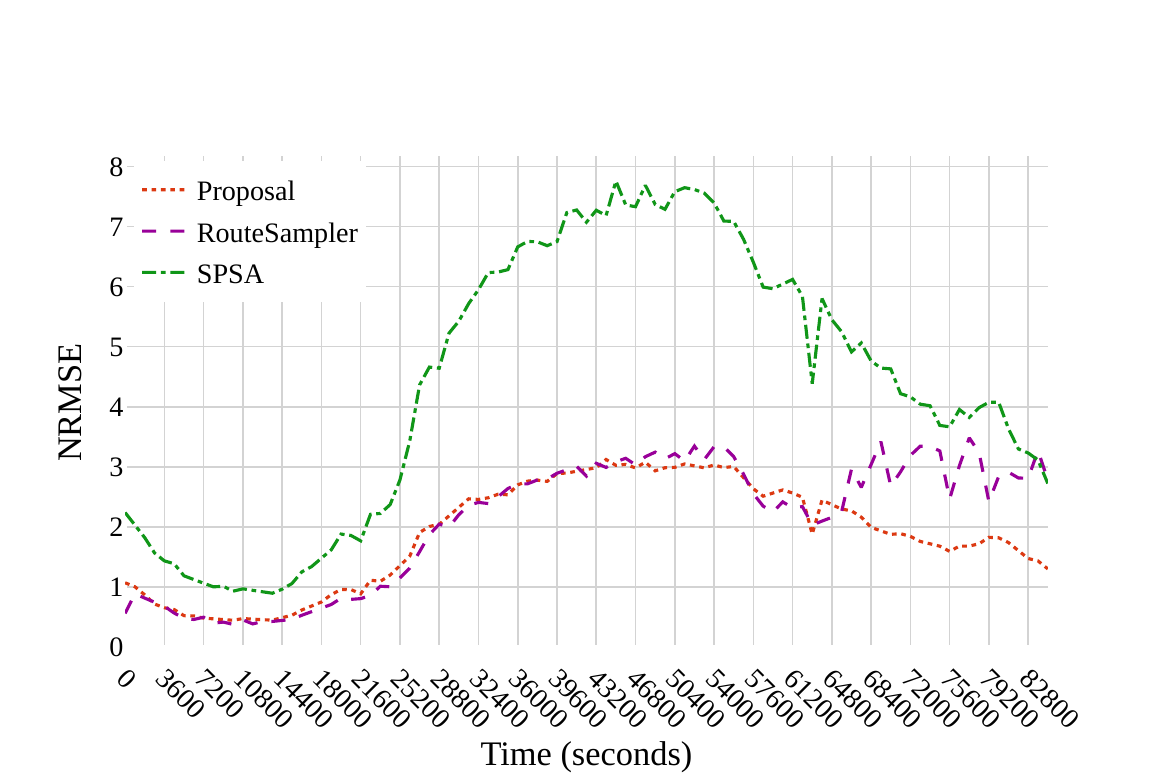}
        \caption{December 2, 2023 (training set)}
        \label{fig:nrmse_1202_train}
    \end{subfigure}
    \begin{subfigure}[b]{.49\textwidth}
        \centering
        \includegraphics[width=\textwidth]{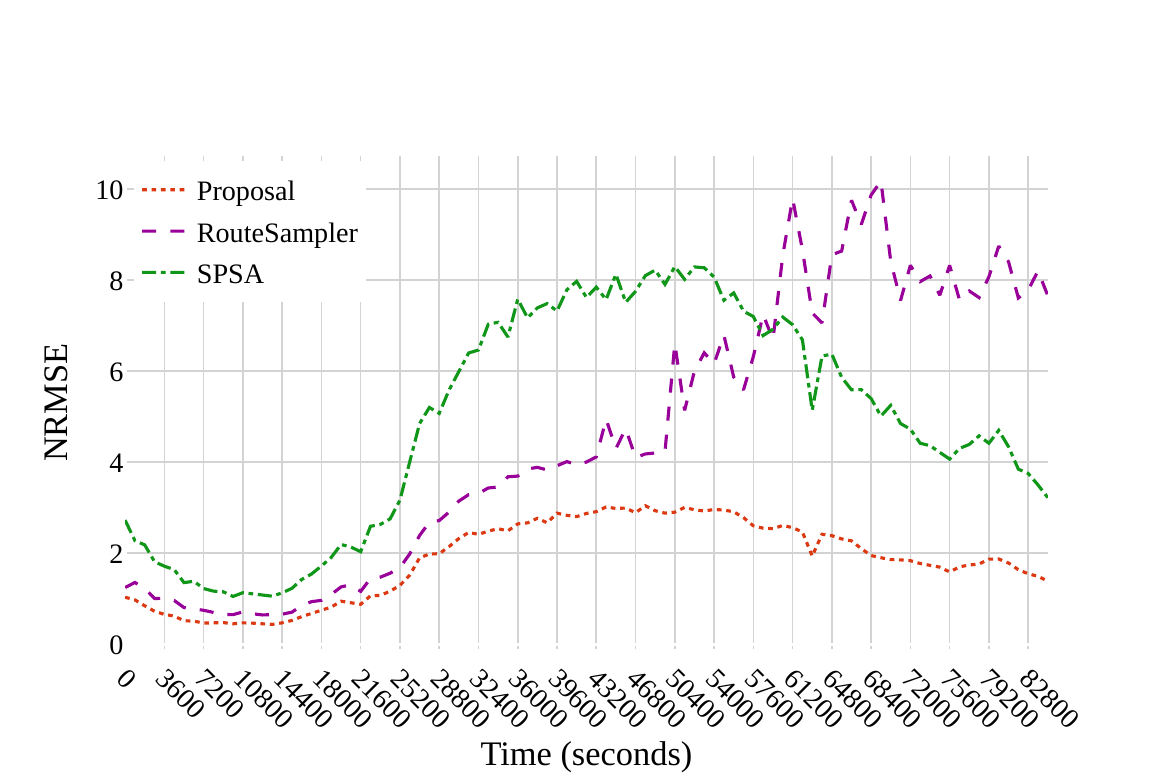}
        \caption{December 2, 2023 (test set)}
        \label{fig:nrmse_1202_test}
    \end{subfigure}
    \caption{Normalized RMSE, calculated using the training set (Figure~\ref{fig:nrmse_1202_train}) and the test set (Figure~\ref{fig:nrmse_1202_test}).}
    \label{fig:nrmse_1202}
\end{figure}


\begin{figure}[!h]
    \centering
    \begin{subfigure}[b]{.49\textwidth}
        \centering
        \includegraphics[width=\textwidth]{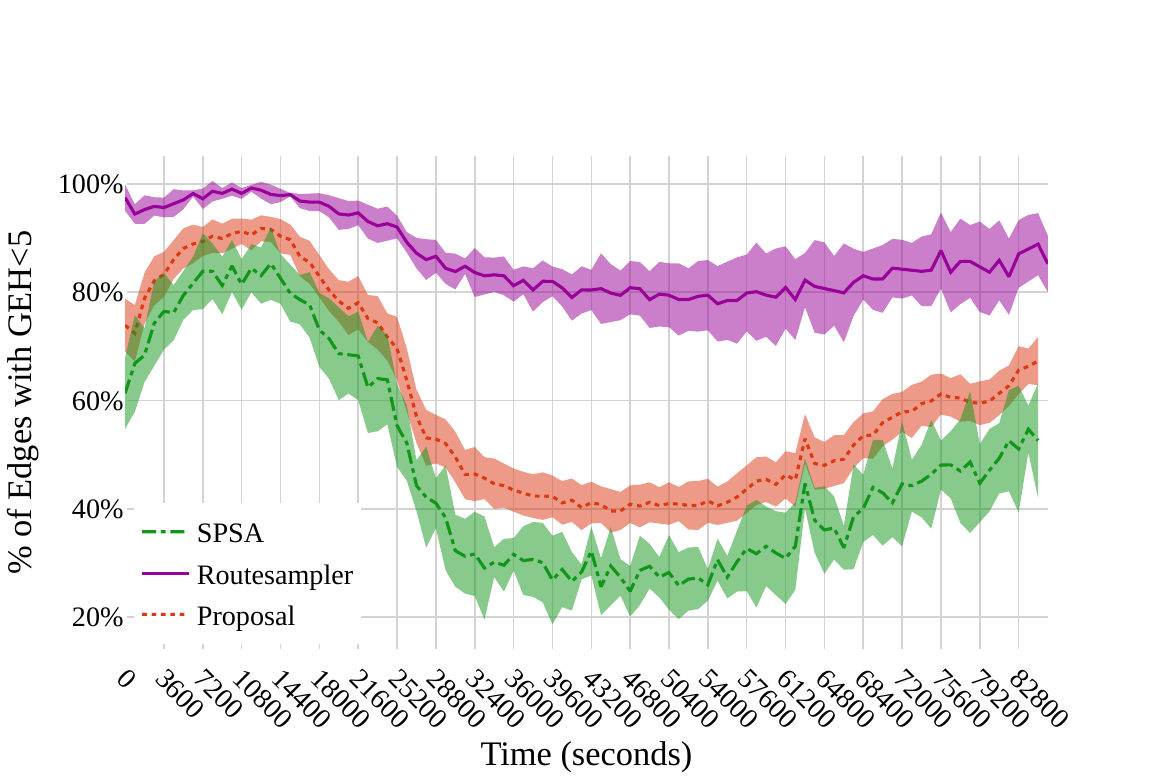}
        \caption{December 2, 2023 (training set)}
        \label{fig:geh_1202_train}
    \end{subfigure}
    \begin{subfigure}[b]{.49\textwidth}
        \centering
        \includegraphics[width=\textwidth]{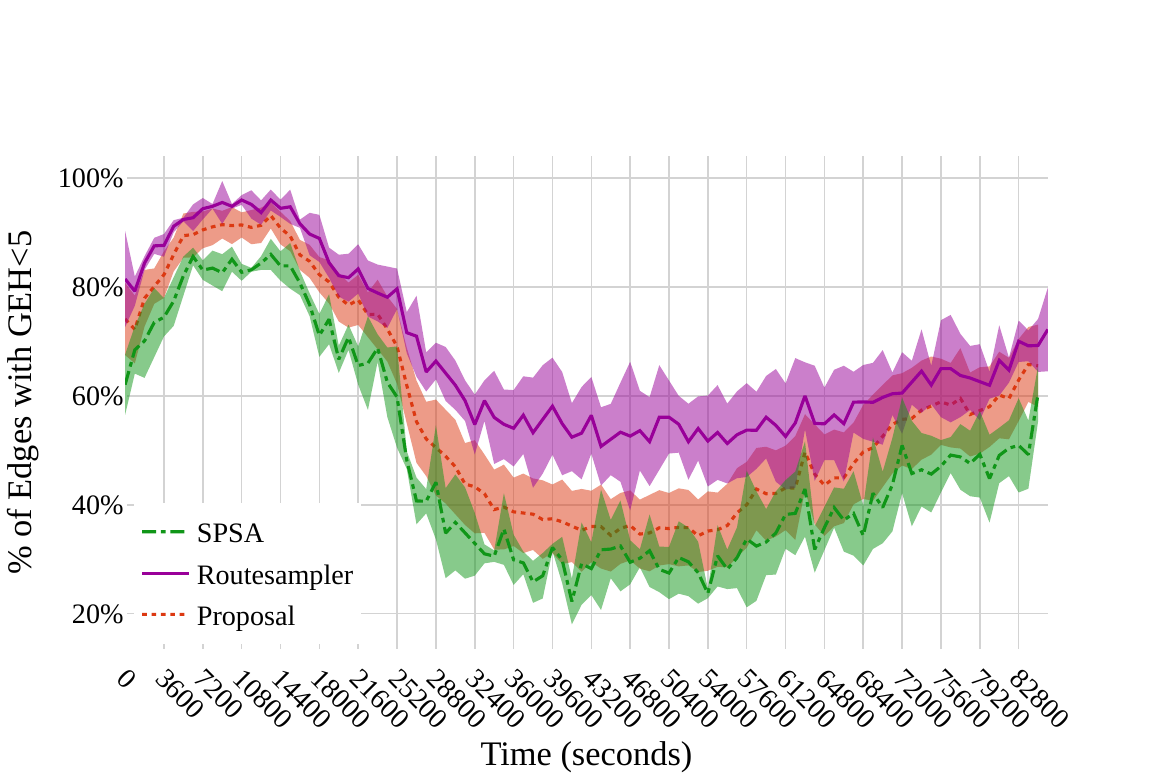}
        \caption{December 2, 2023 (test set)}
        \label{fig:geh_1202_test}
    \end{subfigure}
    \caption{Percentage of edges whose GEH value is less than 5. This is obtained by the proposed and baseline techniques. The values are obtained using the training sensors (Figure~\ref{fig:geh_1202_train}) and the test sensors (Figure~\ref{fig:geh_1202_test}).}
    \label{fig:geh_1202}
\end{figure}

\newpage
\clearpage
\section{Experimental Results - December 3, 2023}
\label{appendix:results_dec3}


\begin{figure}[!h]
    \centering
    \begin{subfigure}[b]{.48\textwidth}
        \centering
        \includegraphics[width=\textwidth]{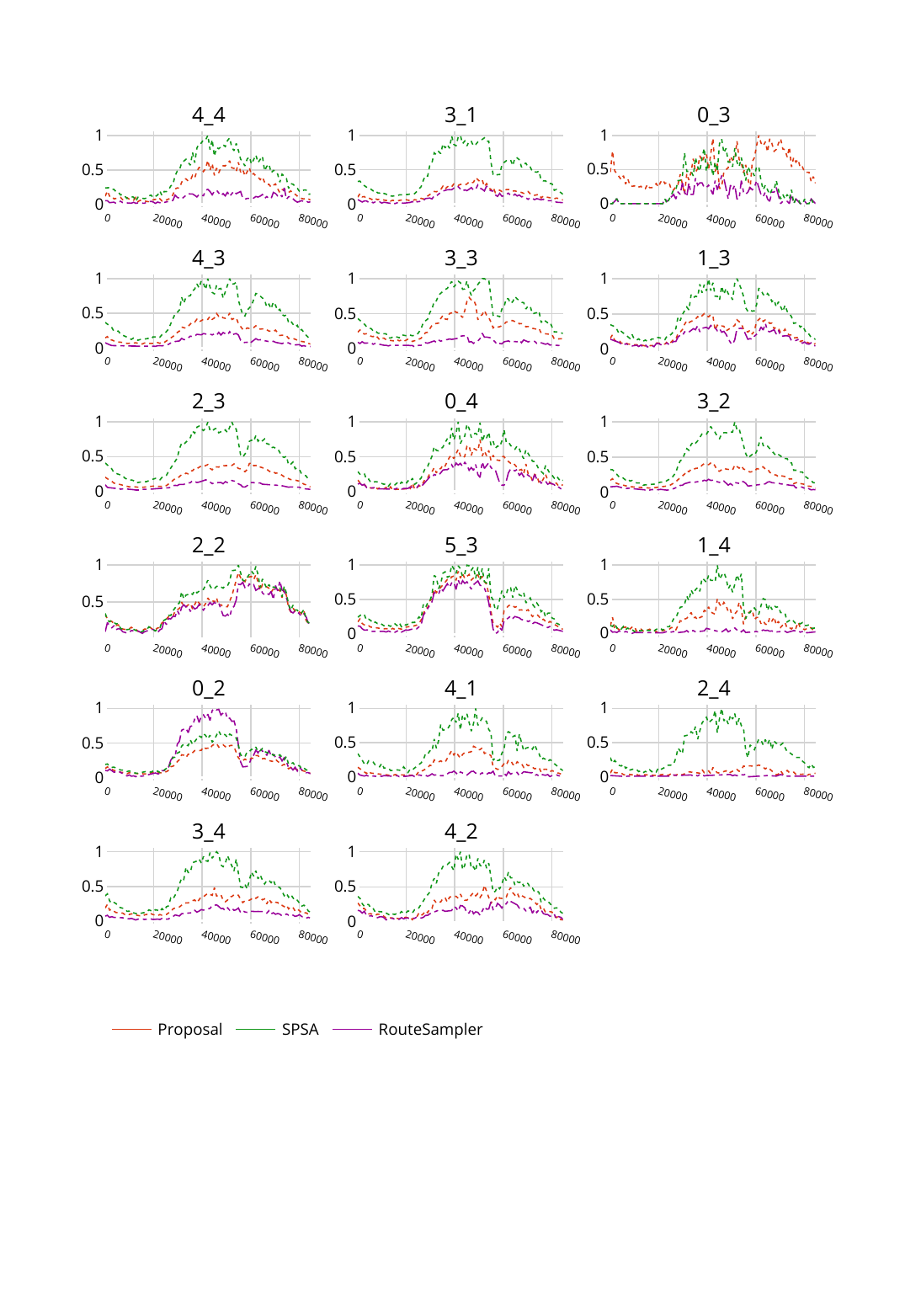}
        \caption{December 3, 2023 (training set)}
        \label{fig:calib_error_1203_train}
    \end{subfigure}
    \begin{subfigure}[b]{.48\textwidth}
        \centering
        \includegraphics[width=\textwidth]{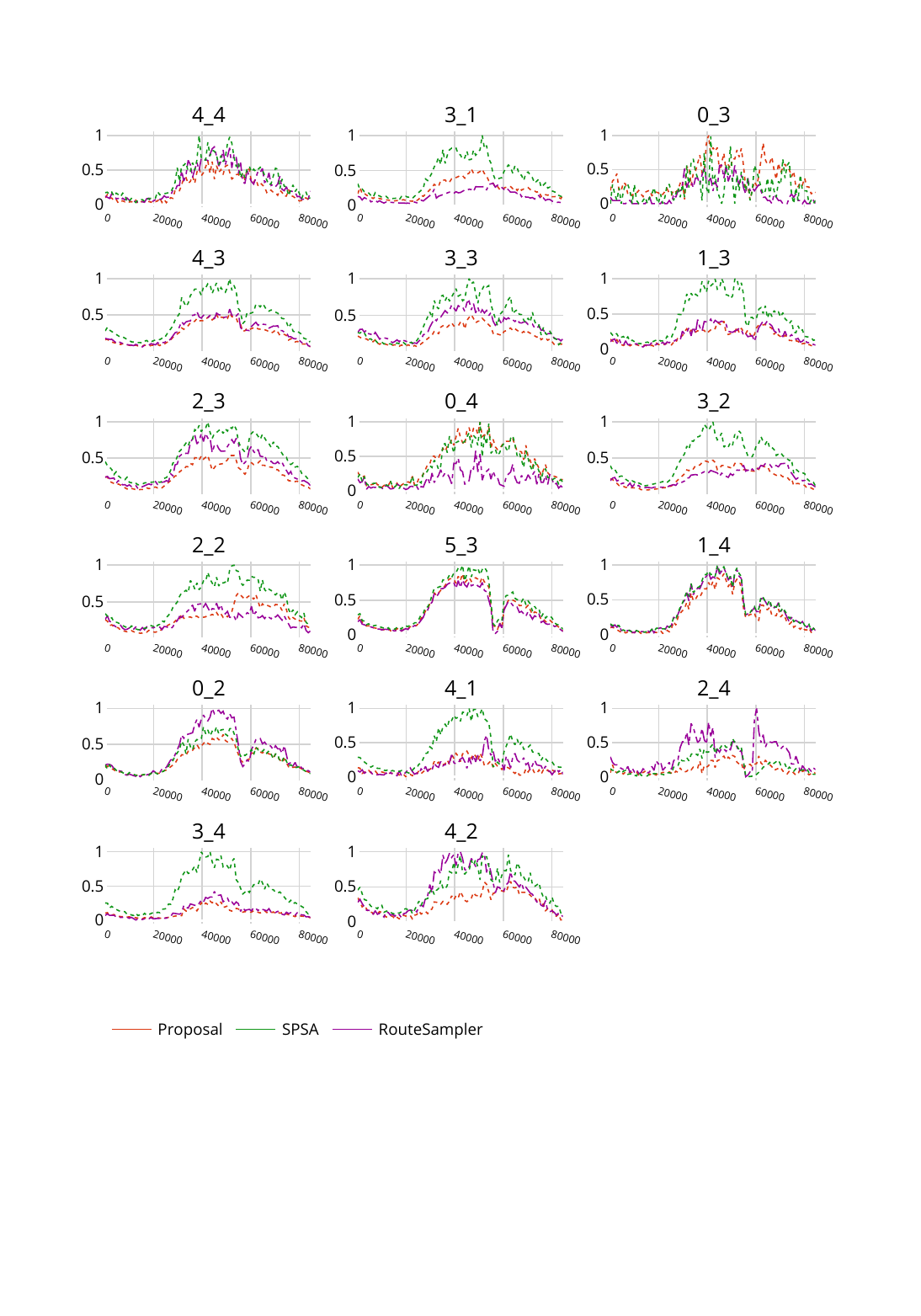}
        \caption{December 3, 2023 (test set)}
        \label{fig:calib_error_1203_test}
    \end{subfigure}
    \caption{Local average absolute traffic counts difference (MAE), calculated on the training set (Figure~\ref{fig:calib_error_1203_train}) and test set (Figure~\ref{fig:calib_error_1203_test}), using the data from December 3, 2023.}
    \label{fig:calib_error_1203}
\end{figure}


\begin{figure}[!h]
    \centering
    \begin{subfigure}[b]{.49\textwidth}
        \centering
        \includegraphics[width=\textwidth]{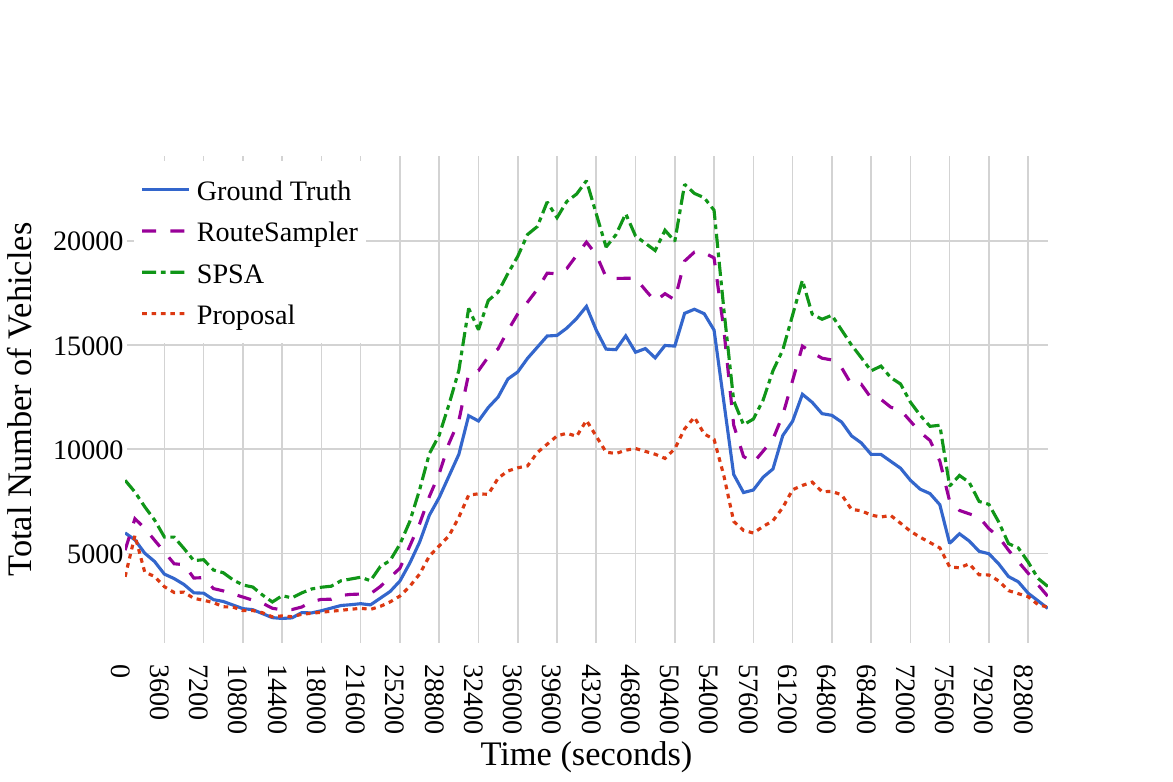}
        \caption{December 3, 2023 (training set)}
        \label{fig:traffic_vol_1203_train}
    \end{subfigure}
    \begin{subfigure}[b]{.49\textwidth}
        \centering
        \includegraphics[width=\textwidth]{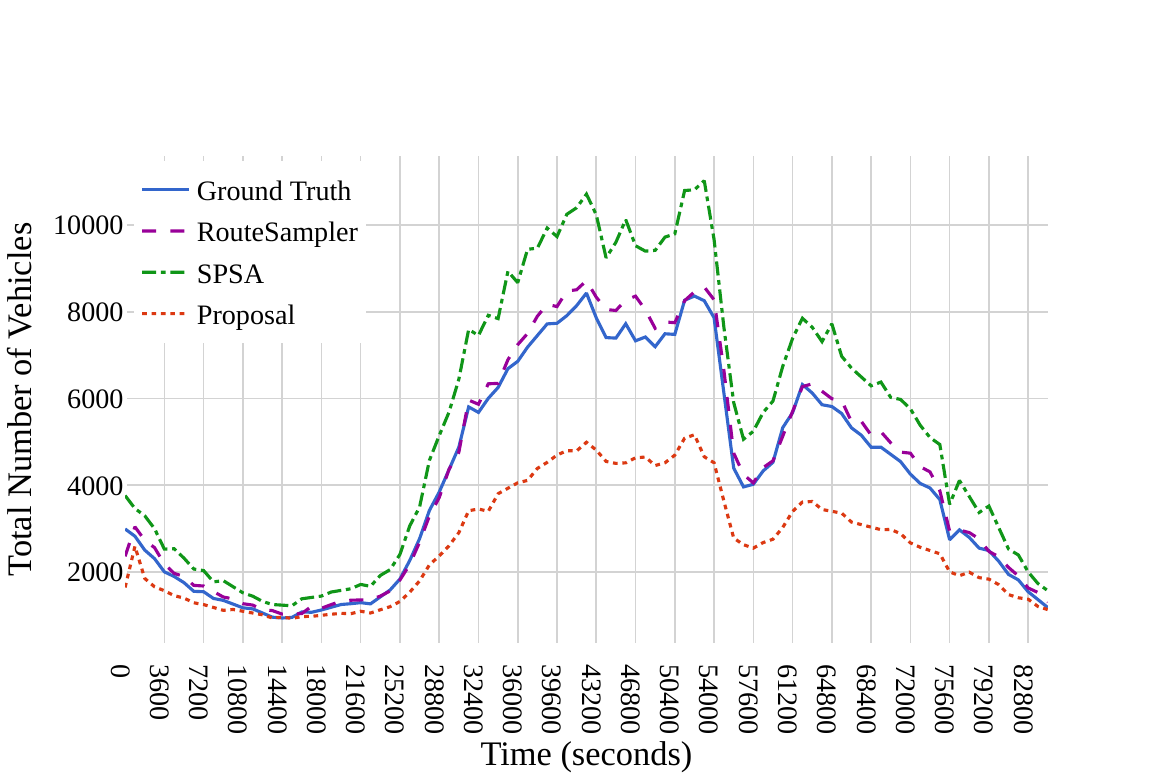}
        \caption{December 3, 2023 (test set)}
        \label{fig:traffic_vol_1203_test}
    \end{subfigure}
    \caption{Total number of vehicles observed in all the regions and every time interval, in the ground truth and the simulation. Figure~\ref{fig:traffic_vol_1203_train} shows the total number of vehicles obtained by the training sensors, Figure~\ref{fig:traffic_vol_1203_test} by the test sensors.}
    \label{fig:traffic_vol_1203}
\end{figure}


\begin{figure}[!h]
    \centering
    \begin{subfigure}[b]{.49\textwidth}
        \centering
        \includegraphics[width=\textwidth]{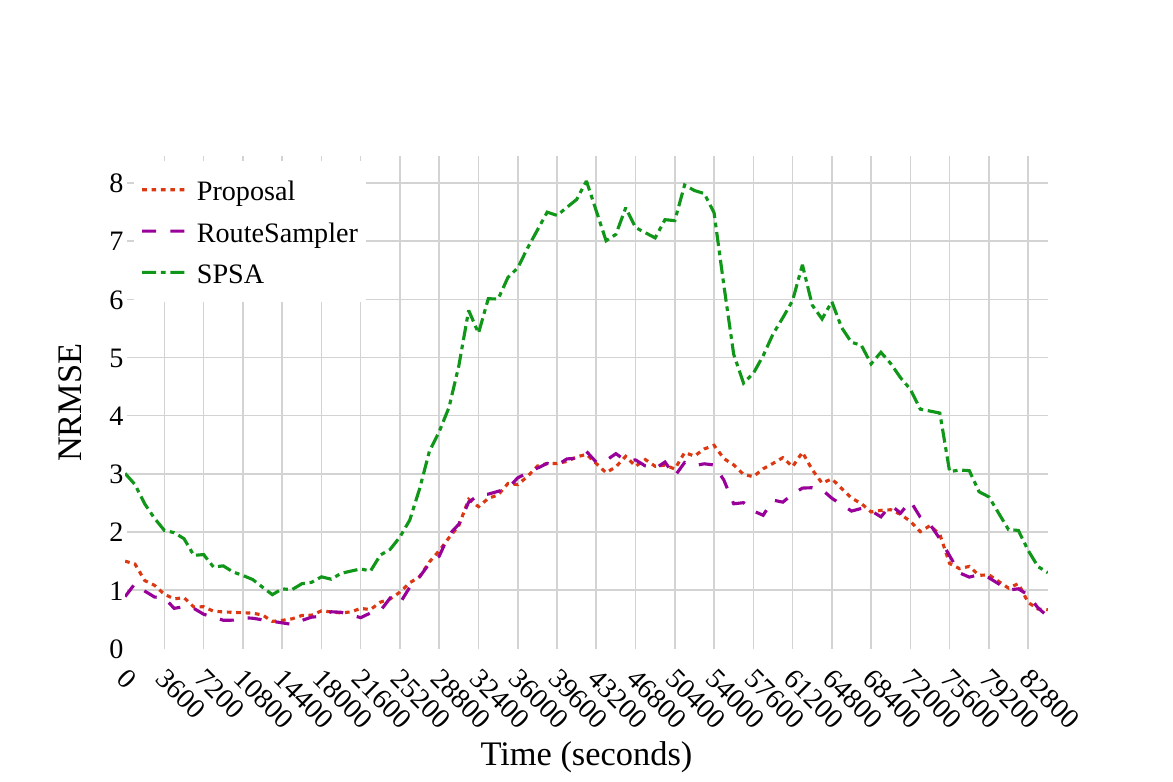}
        \caption{December 3, 2023 (training set)}
        \label{fig:nrmse_1203_train}
    \end{subfigure}
    \hfill
    \begin{subfigure}[b]{.49\textwidth}
        \centering
        \includegraphics[width=\textwidth]{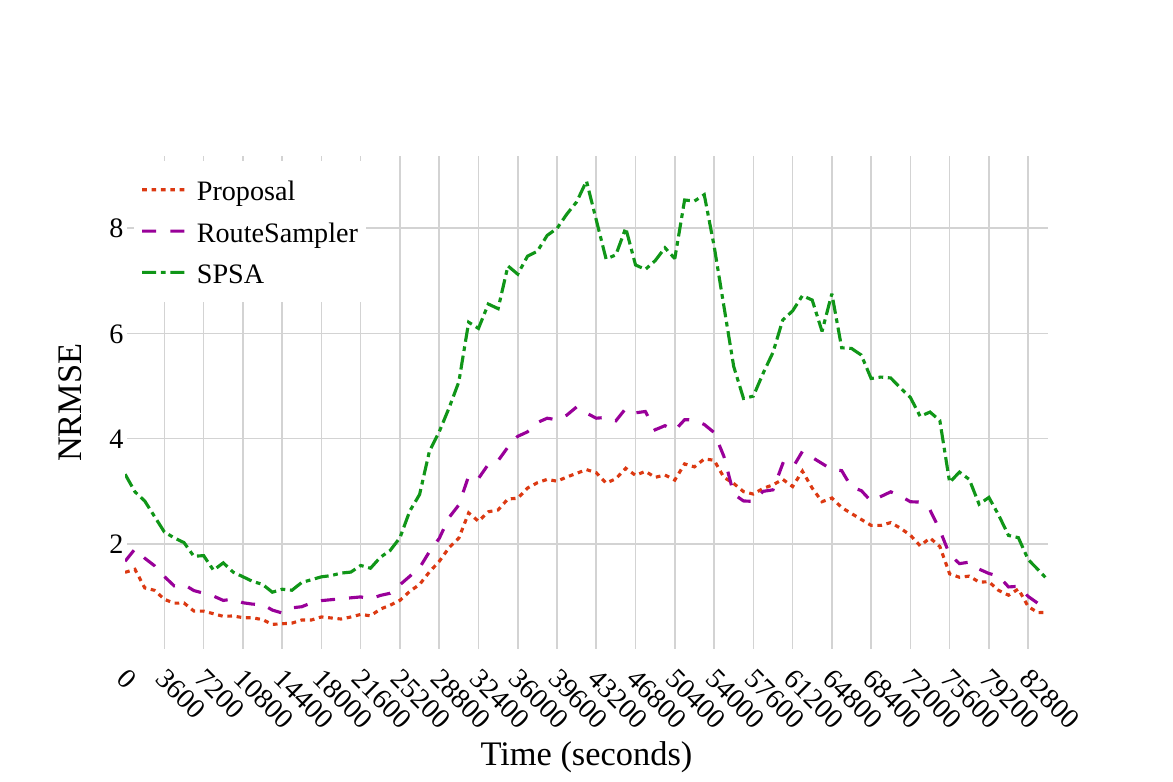}
        \caption{December 3, 2023 (test set)}
        \label{fig:nrmse_1203_test}
    \end{subfigure}
    \caption{Normalized RMSE, calculated using the training set (Figure~\ref{fig:nrmse_1203_train}) and the test set (Figure~\ref{fig:nrmse_1203_test}).}
    \label{fig:nrmse_1203}
\end{figure}


\begin{figure}[!h]
    \centering
    \begin{subfigure}[b]{.49\textwidth}
        \centering
        \includegraphics[width=\textwidth]{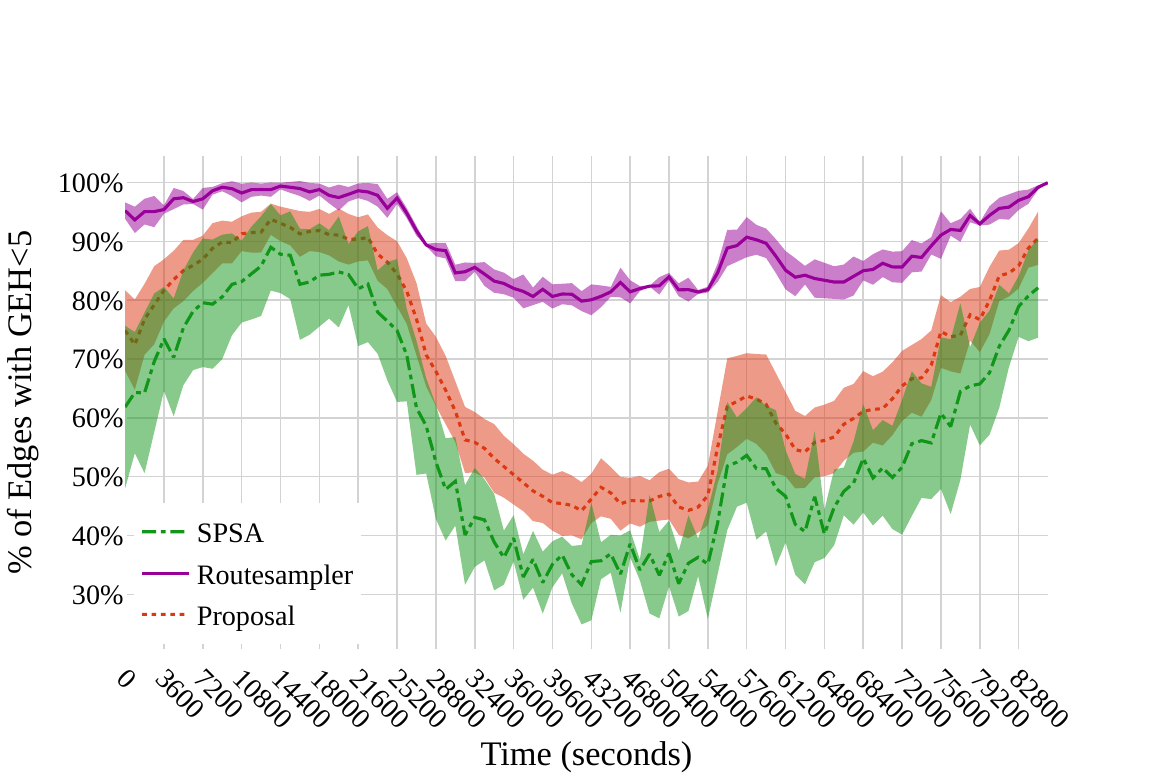}
        \caption{December 3, 2023 (training set)}
        \label{fig:geh_1203_train}
    \end{subfigure}
    \begin{subfigure}[b]{.49\textwidth}
        \centering
        \includegraphics[width=\textwidth]{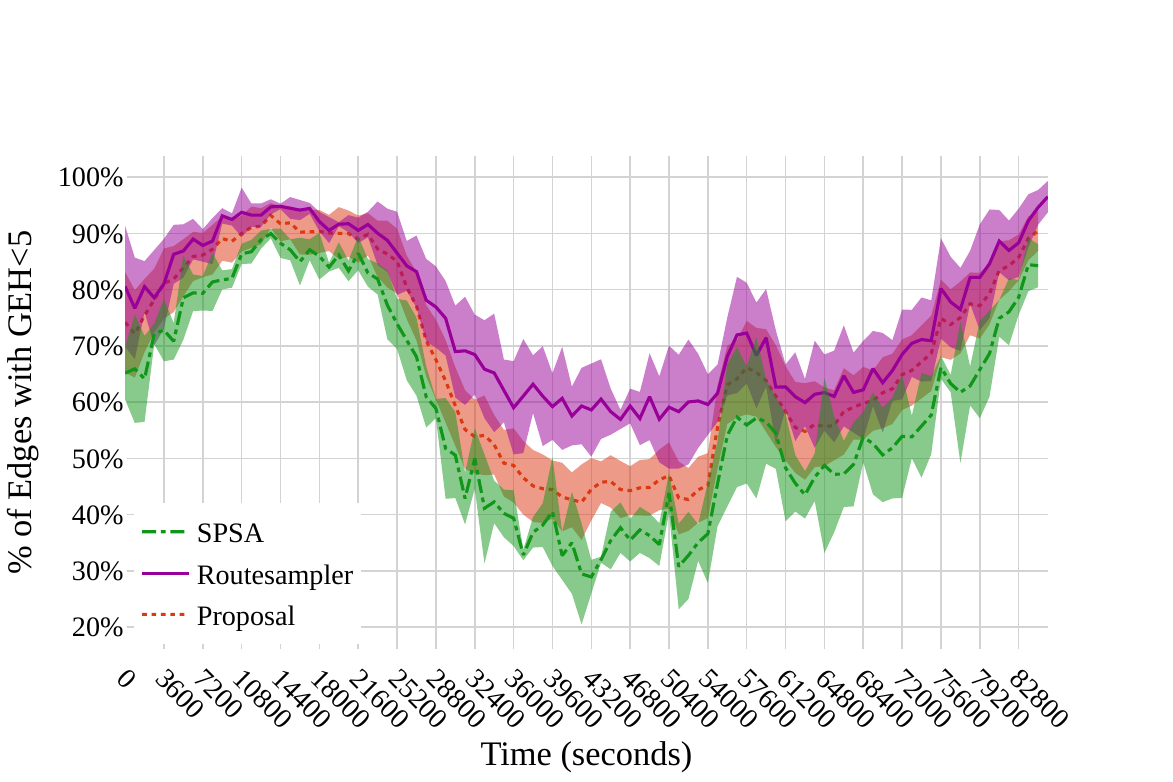}
        \caption{December 3, 2023 (test set)}
        \label{fig:geh_1203_test}
    \end{subfigure}
    \caption{Percentage of edges whose GEH value is less than 5. This is obtained by the proposed and baseline techniques. The values are obtained using the training sensors (Figure~\ref{fig:geh_1203_train}) and the test sensors (Figure~\ref{fig:geh_1203_test}).}
    \label{fig:geh_1203}
\end{figure}

\end{appendices}

\end{document}